%% file: main.tex
\documentclass{article} 
\usepackage{arxiv_neutral,times}

\input{math_commands.tex}

\usepackage{hyperref}
\usepackage{url}

\usepackage[utf8]{inputenc} 
\usepackage[T1]{fontenc}    
\usepackage{booktabs}       
\usepackage{amsfonts}       
\usepackage{nicefrac}       
\usepackage{microtype}      
\usepackage[dvipsnames]{xcolor} 

\usepackage{amsmath, amssymb}
\usepackage{graphicx}
\usepackage{xspace}
\usepackage{array}
\usepackage{makecell}
\usepackage{comment}
\usepackage{enumitem}
\usepackage{xcolor}
\usepackage[table]{xcolor}  
\usepackage{inconsolata} 
\usepackage{multirow}
\usepackage{wrapfig}   
\usepackage{adjustbox} 
\usepackage{float}     
\usepackage[most]{tcolorbox}
\usepackage[normalem]{ulem} 
\useunder{\uline}{\ul}{}   
\usepackage{arydshln}  
\usepackage[group-separator={,}]{siunitx}  
\usepackage[capitalise,noabbrev]{cleveref}
\usepackage{fontawesome5}  
\usepackage{caption}
\usepackage{longtable}   

\definecolor{TitleAccent}{HTML}{5A1822}

\definecolor{darkblue}{rgb}{0,0,0.5}
\hypersetup{
    colorlinks=true,
    citecolor=darkblue,
    linkcolor=darkblue,
    urlcolor=darkblue
}

\newcommand{\rqnum}[1]{%
  \tikz[baseline=(X.base)]{
    \node (X) [draw, circle, inner sep=1pt, fill=black!7] {\textbf{#1}};
  }%
}

\newcommand{\AmE}{\ensuremath{\mathrm{AmE}}}
\newcommand{\RefBrE}{\ensuremath{\mathrm{Ref}_{\mathrm{BrE}}}}

\newcommand{\bRefBrE}{\ensuremath{\boldsymbol{\mathrm{Ref}}_{\boldsymbol{\mathrm{BrE}}}}}

\newcommand{\sref}[1]{\S\ref{#1}}

\definecolor{revforest}{rgb}{0.00,0.40,0.00}

\usepackage[most]{tcolorbox}
\newtcolorbox{keybox}[1][]{
  enhanced,
  sharp corners,
  boxsep=0pt,
  left=3pt, right=3pt, top=3pt, bottom=3pt,
  before skip=0pt, after skip=0pt,
  colback=green!3,
  colframe=green!40!black,
  boxrule=.4pt,
  drop shadow={black!25},   
  #1
}

\newcommand{\usflag}[1][0.37cm]{\includegraphics[width=#1]{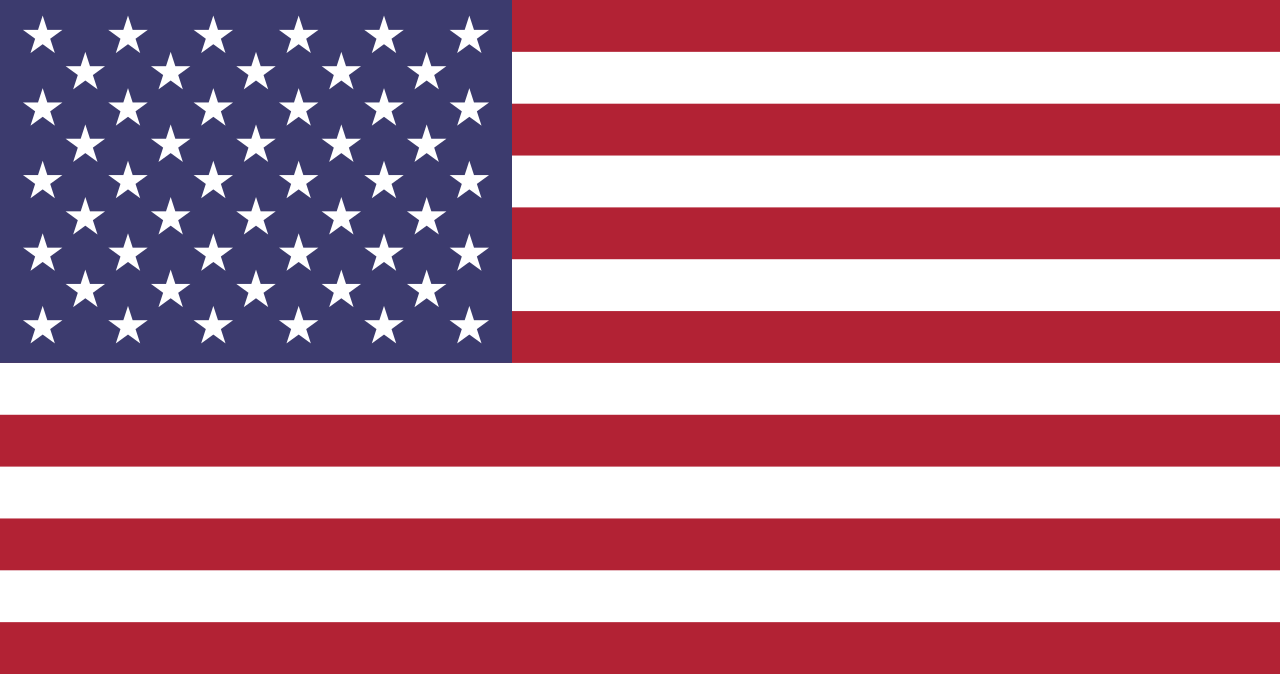}\xspace}
\newcommand{\ukflag}[1][0.37cm]{\includegraphics[width=#1]{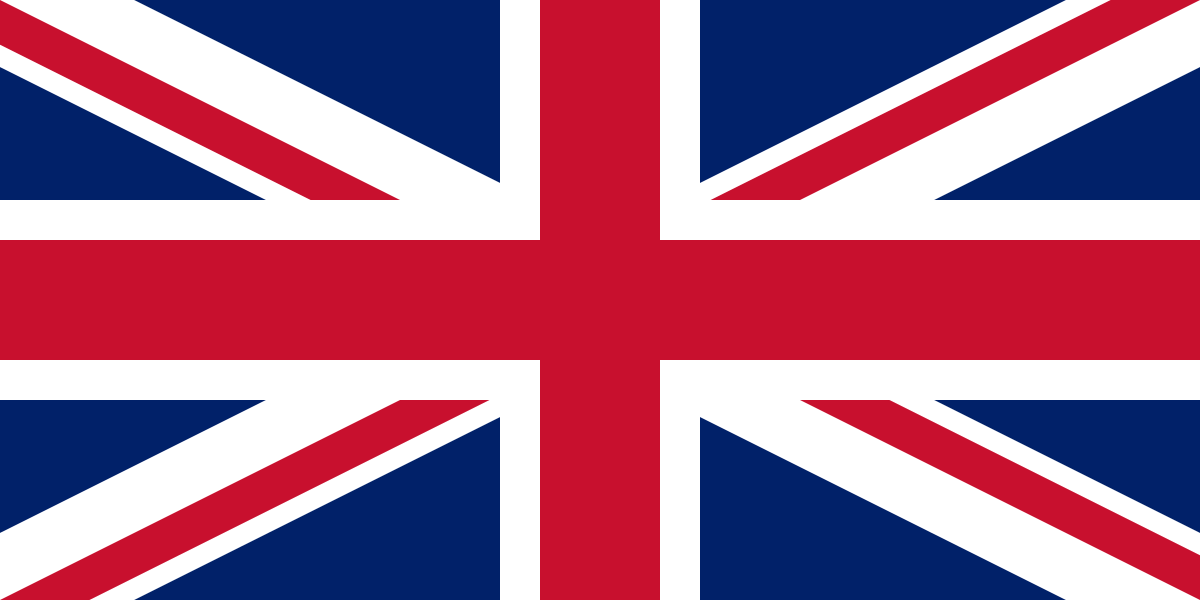}\xspace}
\newcommand{\chinaflag}[1][0.37cm]{\includegraphics[width=#1]{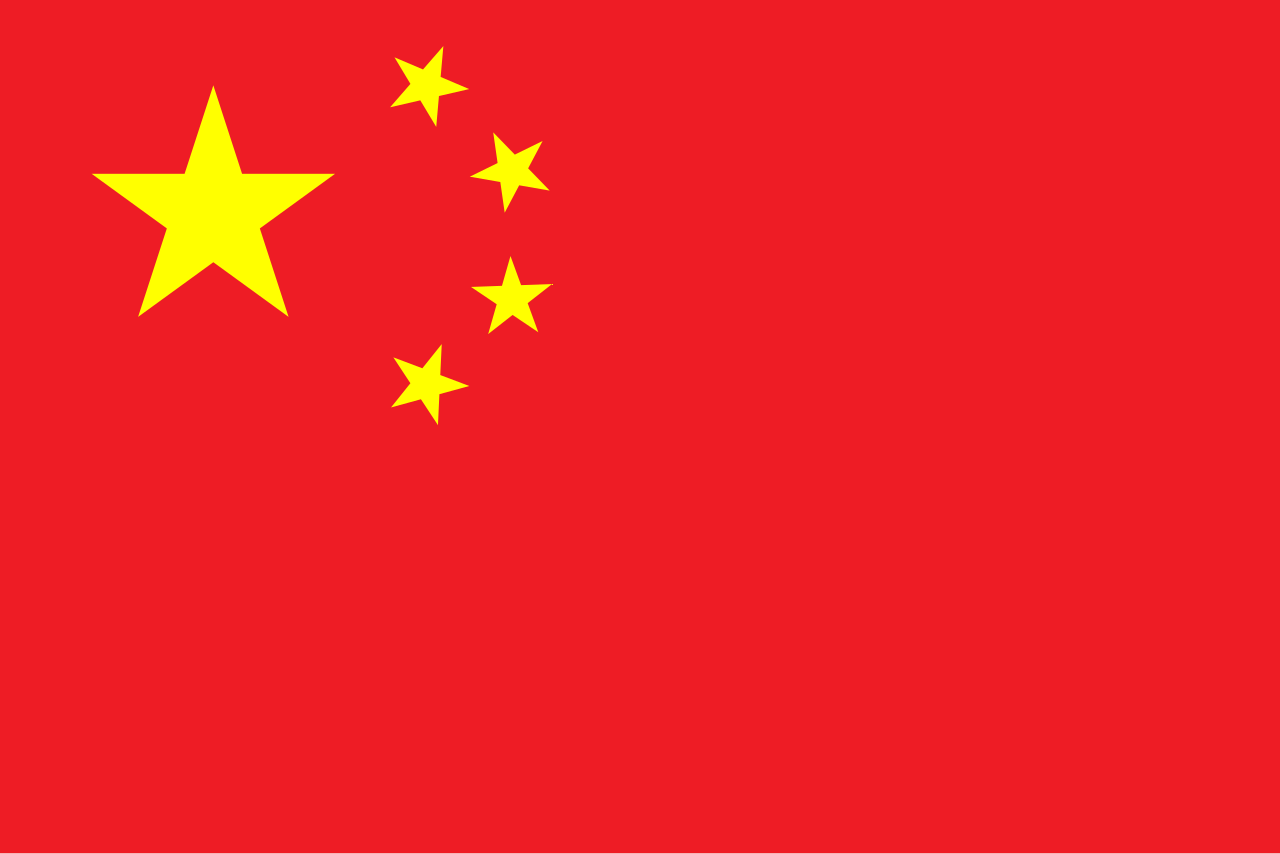}\xspace}
\newcommand{\franceflag}[1][0.37cm]{\includegraphics[width=#1]{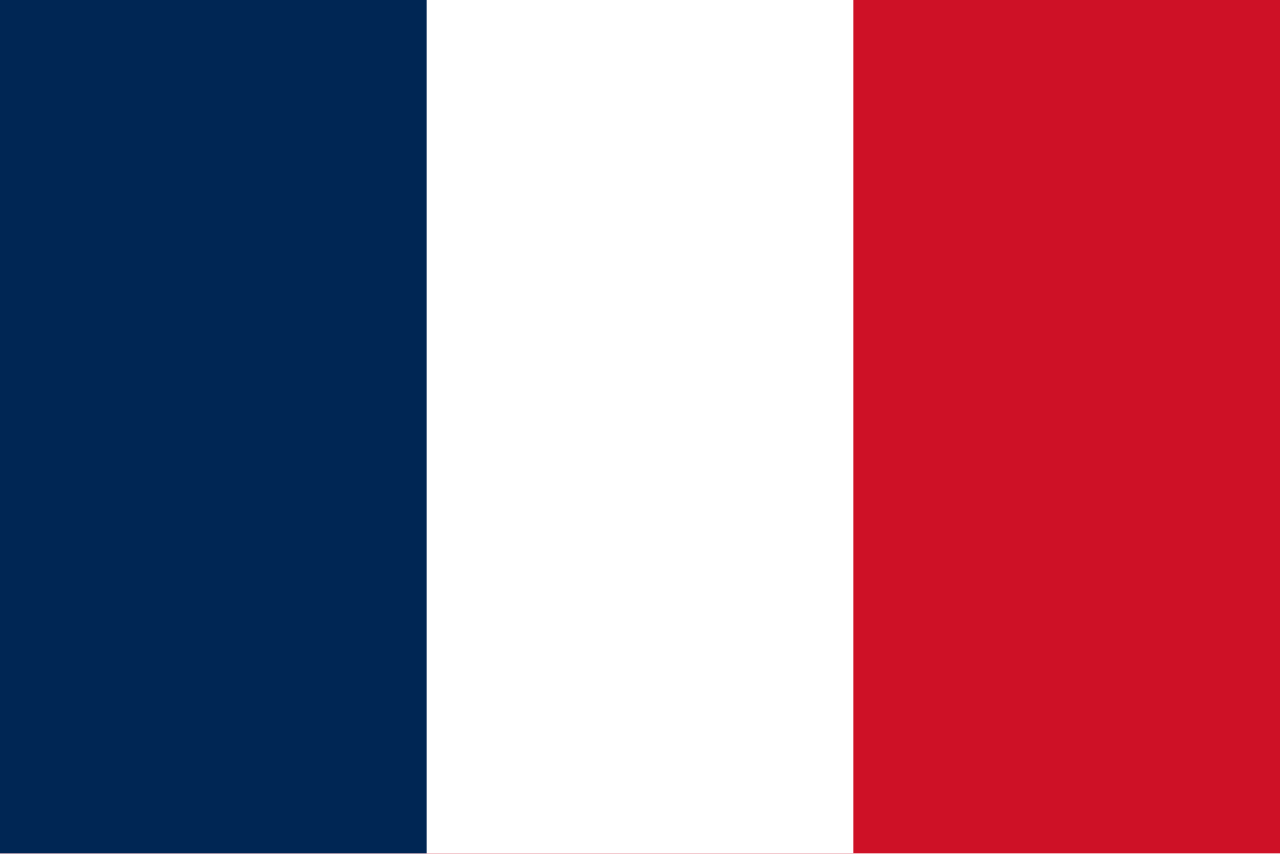}\xspace}
\newcommand{\italyflag}[1][0.37cm]{\includegraphics[width=#1]{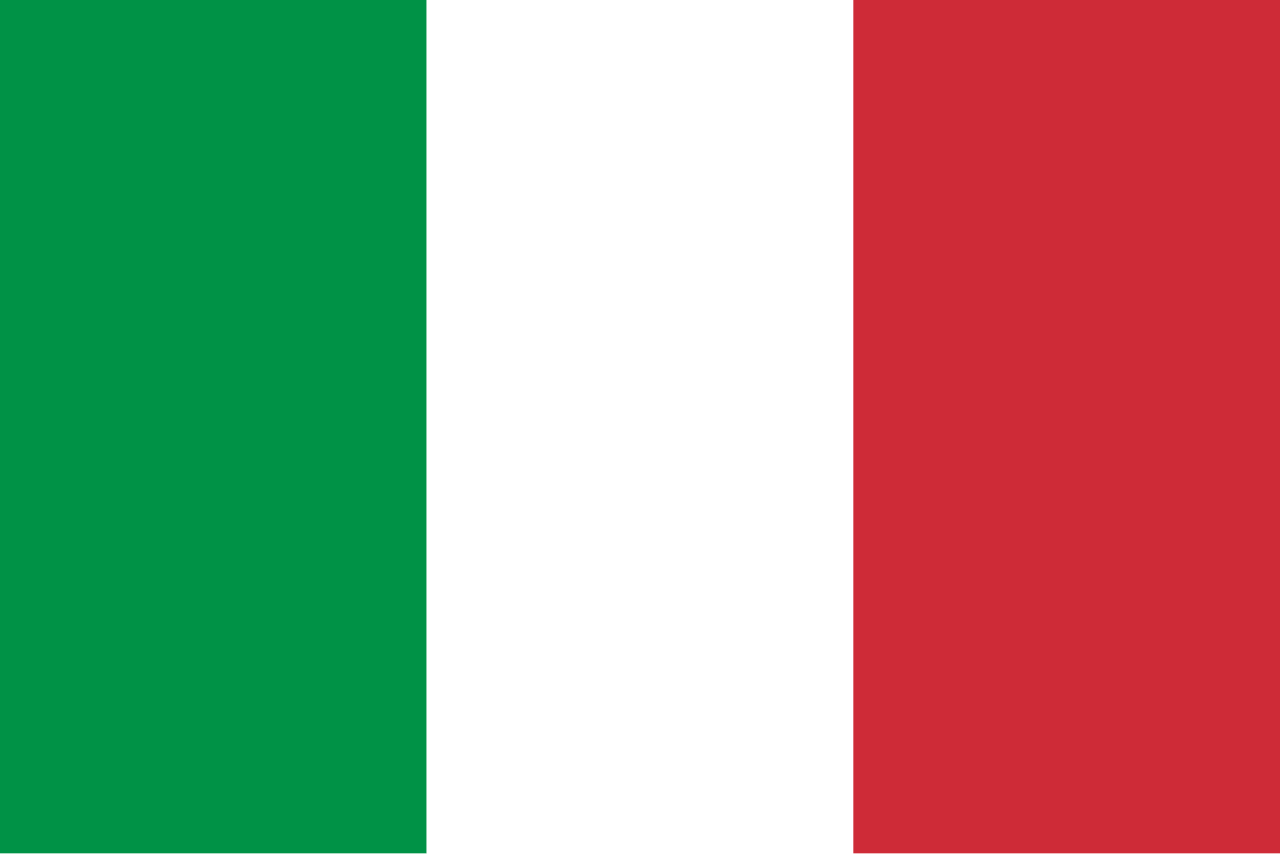}\xspace}
\newcommand{\uaeflag}[1][0.37cm]{\includegraphics[width=#1]{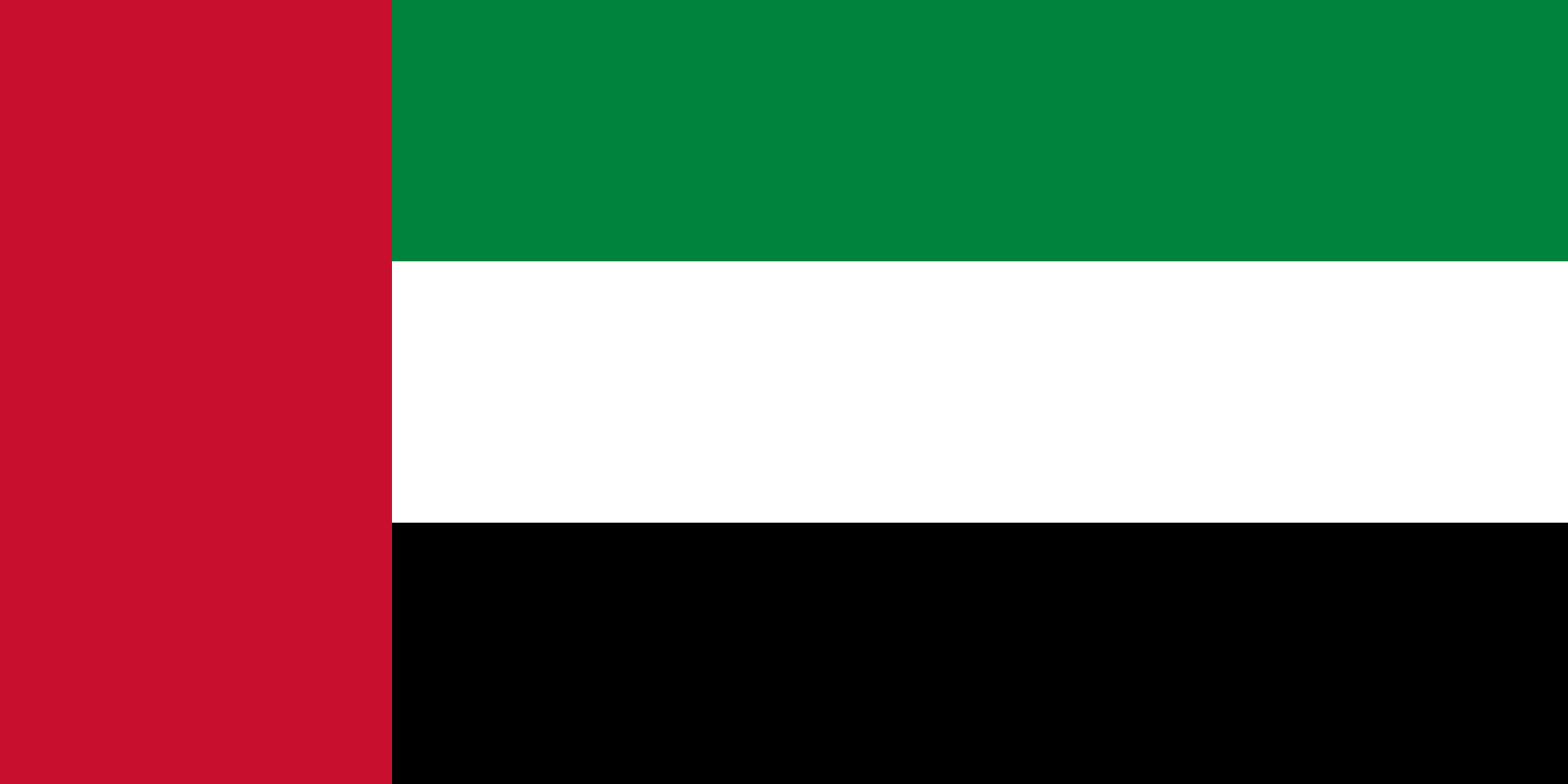}\xspace}

\definecolor{incCol}{named}{ForestGreen}
\definecolor{decCol}{named}{BrickRed}

\newcommand{\webIcon}{\raisebox{-2pt}{\includegraphics[height=1.15em]{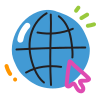}}\xspace}
\newcommand{\huggingface}{\raisebox{-1.5pt}{\includegraphics[height=1.05em]{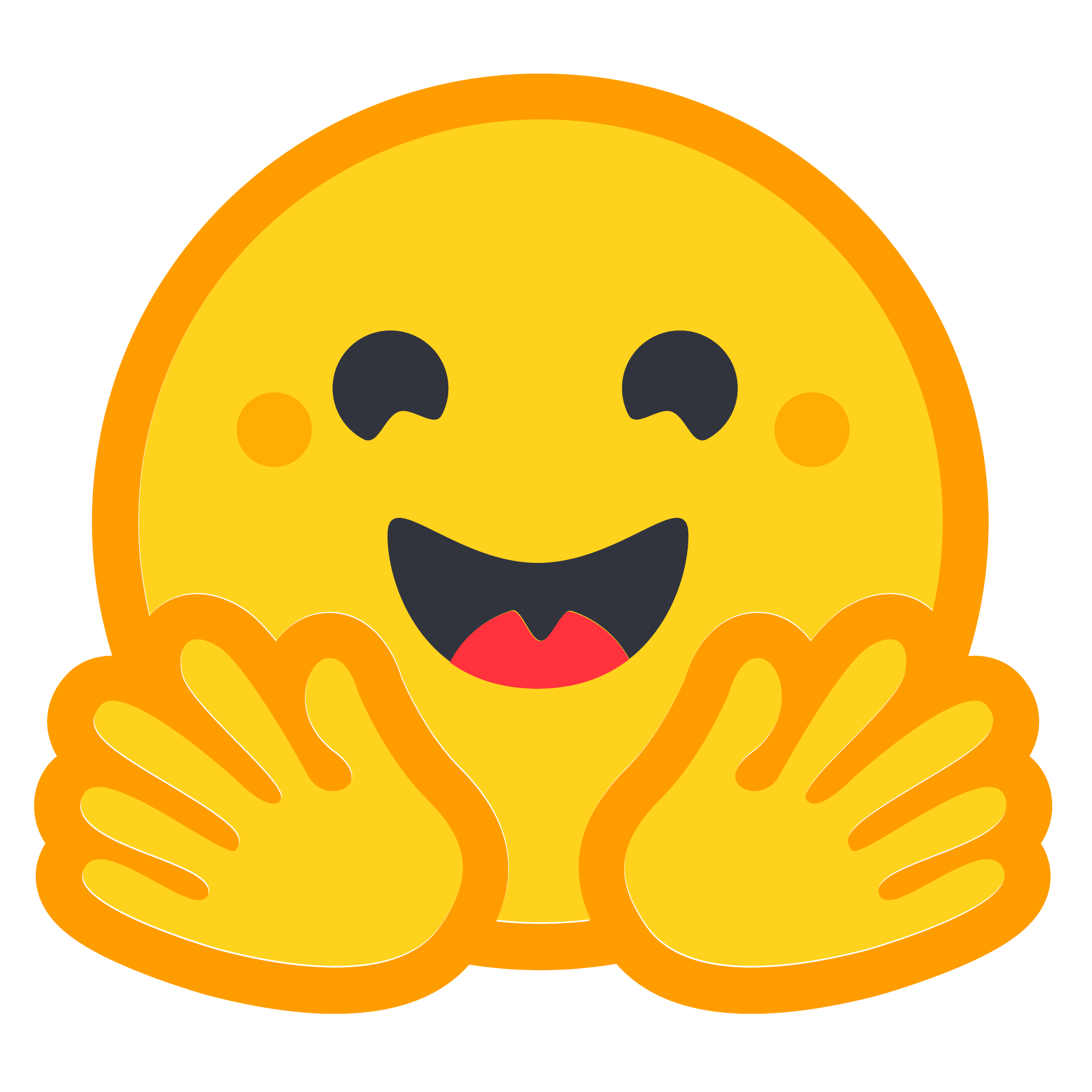}}\xspace}
\newcommand{\github}{\raisebox{-1.5pt}{\includegraphics[height=1.05em]{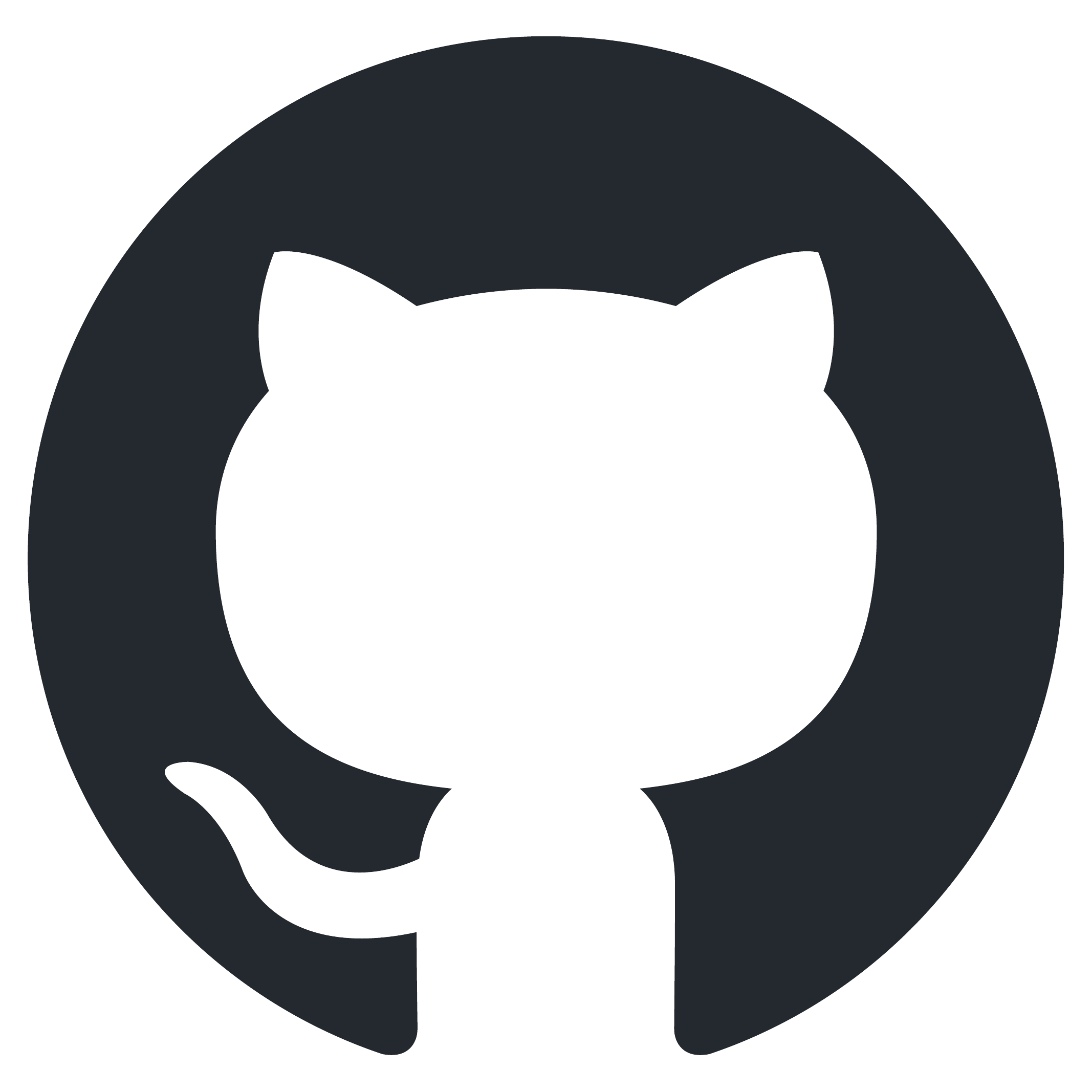}}\xspace}

\newcommand{\inc}[1]{\textcolor{BrickRed}{$\uparrow$\xspace#1\,\%}}
\newcommand{\dec}[1]{\textcolor{ForestGreen}{$\downarrow$\xspace#1\,\%}}

\definecolor{VarnishAlfaFour}{HTML}{01A2CA}
\definecolor{VarnishPapaThree}{HTML}{7446F2}
\definecolor{VarnishRomeoFour}{HTML}{D63F3F}
\definecolor{VarnishGolfThree}{HTML}{1EC28E}
\definecolor{VarnishOscarFour}{HTML}{FF9100}
\definecolor{VarnishFoxtrotThree}{HTML}{D864C9}
\definecolor{VarnishBravoThree}{HTML}{265ED4}
\definecolor{VarnishBravoFour}{HTML}{1B4596}
\definecolor{VarnishFoxtrotFour}{HTML}{A44397}

\newcommand{\iconWeb}{{\scriptsize\color{VarnishAlfaFour}\faGlobe}\xspace}
\newcommand{\iconRefs}{{\scriptsize\color{VarnishPapaThree}\faBookmark}\xspace}
\newcommand{\iconCode}{{\scriptsize\color{VarnishRomeoFour}\faCode}\xspace}
\newcommand{\iconBooks}{{\scriptsize\color{VarnishGolfThree}\faBook}\xspace}
\newcommand{\iconSocial}{{\scriptsize\color{VarnishOscarFour}\faComments}\xspace}
\newcommand{\iconPapers}{{\scriptsize\color{VarnishFoxtrotFour}\faGraduationCap}\xspace}
\newcommand{\iconStackExchange}{{\scriptsize\color{VarnishBravoThree}\faStackExchange}\xspace}

\definecolor{lockcolor}{HTML}{DAA520}  
\newcommand{\closedSource}{{\scriptsize\color{black}\faLock}\xspace}
\newcommand{\openSource}  {{\scriptsize\color{black}\faLockOpen}\xspace}

\title{
How Does \textcolor{TitleAccent}{``English (US)''} Become the Default?\\
Triangulating Structural Bias Towards American English\\ Across the LLM Pipeline
}

\author{Mir Tafseer Nayeem \quad Davood Rafiei\\
Department of Computing Science\\
University of Alberta\\
\texttt{\{mnayeem, drafiei\}@ualberta.ca}\\
}

\begin{document}

\maketitle

\begin{abstract}
Large language models (LLMs) are increasingly embedded in educational,
professional, and public infrastructure, yet widely used platforms expose
``English (US)'' as a primary English setting despite the global diversity
of English. We ask: \emph{How does ``English (US)'' become the default?}
We study this question as \emph{structural bias}, examining how geopolitical
histories of data curation, digital dominance, and linguistic standardization intersect with the LLM development pipeline. Using British English as a controlled reference, we construct a curated resource of 1,813 matched American English (AmE)--British English (BrE) variants and introduce
\textsc{DiAlign}, a dynamic, training-free method for estimating regional
alignment from distributional evidence. We triangulate the AmE preference
across \emph{data exposure $\rightarrow$ representation $\rightarrow$
generation}, jointly examining pretraining and post-training data, tokenizer
behavior and provenance, model prediction cost, and generated language
across developer countries, prompt conditions, domains and sources,
linguistic categories, and registers. AmE is consistently favored across
all six audited pretraining corpora and 21 post-training datasets, is
generally represented more compactly by tokenizers, and receives lower
prediction cost. It also remains the dominant generation default under
neutral English prompting; British-English prompting shifts this preference
toward BrE but does not consistently eliminate the AmE default. To our
knowledge, this is the first rigorous pipeline-wide study of
\emph{structural bias} across major phases of LLM development. Our findings
show that contemporary LLMs privilege AmE as the \emph{de facto} norm,
raising concerns about linguistic homogenization, epistemic injustice, and
inequity in global AI deployment, while providing a rigorous basis for
targeted component-level intervention.\footnote{\webIcon{} \github{} \huggingface{} \textbf{Project website:} \href{https://tafseer-nayeem.github.io/LLM-structural-bias/}{\texttt{\path{tafseer-nayeem.github.io/LLM-structural-bias}}}}
\end{abstract}


\section{Introduction}
\label{sec:introduction}

Large language models (LLMs) are rapidly expanding across technical,
professional, educational, legal, and public-sector applications~\citep{LAI2024181,KULAL2024100329,shahzad2025llmreview,StanfordAIIndex2026}. AI is now used in at least one
area of government in 35 of 36 OECD countries~\citep{OECDDigitalGovernment2026}, while national AI strategies increasingly emphasize sovereign control over compute, data, models, and infrastructure~\citep{reuel2025open,StanfordAIIndex2026}. As LLMs become part of institutional infrastructure, their defaults can operate at global scale. Language is one such default: English remains the most effective language for interacting with many LLMs~\citep{behzad-etal-2024-ask}, yet widely used platforms such as ChatGPT and Claude expose \texttt{``English (US)''} as a primary English setting. This motivates our central question: \emph{How does ``English (US)'' become the default?} If LLMs implicitly treat American English (AmE) as the default or normative form, the concern extends beyond stylistic preference to linguistic homogenization and uneven user experience worldwide~\citep{10.1145/3442188.3445922}. Answering this question requires rigorous pipeline-wide investigation that can localize the asymmetry
and identify where component-level improvements should be directed.

\begin{figure}[t]
    \centering
    \includegraphics[width=0.932\columnwidth]{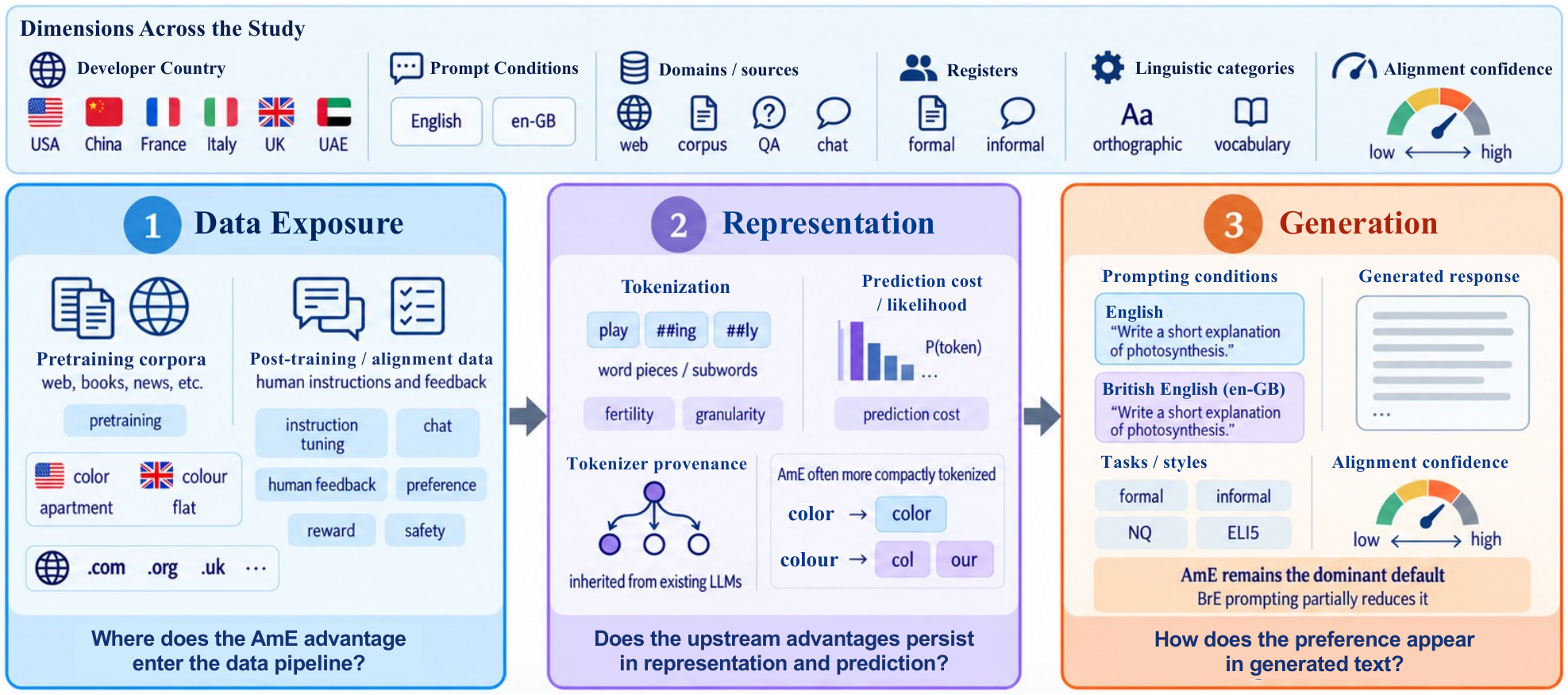}
    \vspace{-1.5mm}
    \caption{\small
    \textbf{Triangulating how ``English (US)'' becomes the default across the LLM pipeline.} We trace this asymmetry across three connected stages:
    \textbf{data exposure $\rightarrow$ representation $\rightarrow$ generation},
    using British English as the controlled reference, Ref(BrE). The study
    jointly examines pretraining and post-training data, tokenizer behavior
    and provenance, model prediction cost, and generated language. This
    triangulation reveals where the American English (AmE) advantage first becomes visible, whether it persists across components, and how it manifests in model outputs. Its strength and form are examined across pipeline stages, developer countries~(e.g., USA, UK, China, France, Italy, and UAE), prompt
    conditions~(e.g., default English vs.\ British English), domains and
    sources~(e.g., web, books, encyclopedic text, and mixed-domain corpora),
    linguistic categories (e.g., orthographic vs.\ vocabulary contrasts),
    registers (e.g., formal/encyclopedic vs.\ informal/conversational), and
    alignment confidence levels. This breadth provides a deeper,
    pipeline-level view of \emph{structural bias}.
    }
    \vspace{-2mm}
    \label{fig:ame-pipeline-triangulation}
\end{figure}

We study this problem as \emph{structural bias}: a systematic linguistic
advantage that appears across multiple components of model development,
not only in generated text. Sociolinguistic research shows that power and
prestige shape which linguistic forms are legitimized or marginalized
\citep{labov1972sociolinguistic,labov2006socialstratification}. In modern
LLMs, these processes intersect with geopolitical histories of data
curation, digital dominance, and linguistic standardization. We therefore
triangulate the AmE preference to demystify the ``English (US)'' default across
\emph{data exposure $\rightarrow$ representation $\rightarrow$ generation},
covering pretraining and post-training data, tokenization and tokenizer
provenance, model prediction cost, and generated language
(see \cref{fig:ame-pipeline-triangulation}). To our knowledge, this is the first
systematic and rigorous pipeline-wide experimental study of
\emph{structural bias} across the LLM
development pipeline.

This pipeline begins with data. Modern LLMs are trained on massive
internet-derived corpora~\citep{dodge-etal-2021-documenting}, and English
accounts for roughly \num{50}--\num{60}\% of global web content
\citep{petrosyan2025internetlanguages}. Its share is even higher in several
documented training mixtures: approximately \num{92.65}\% for
GPT-3,\footnote{\href{https://github.com/openai/gpt-3/blob/master/dataset_statistics/languages_by_word_count.csv}
{OpenAI GPT-3 Dataset Language Statistics (GitHub, accessed August 06,
2026)}} \num{89.7}\% for Llama~2
\citep{touvron2023llama2openfoundation}, and nearly \num{90}\% for
Claude~2~\citep{anthropic2023claude}. Yet the English-language Internet is
itself uneven. The global reach of U.S. media, technology companies,
commerce, and digital platforms has made AmE especially prominent online
\citep{10.1371/journal.pone.0197741}. Because web-scale pretraining draws from this ecosystem, inequalities in
which English forms are produced and circulated online can become
inequalities in model training. These imbalances may extend beyond
linguistic form, as dominant sources also encode cultural conventions,
institutional practices, and perspectives that receive greater visibility
in training data~\citep{10.1145/3442188.3445922}.

We examine a wider infrastructure question: how globally uneven data
ecosystems and historically shaped linguistic standards become reflected in
foundation-model infrastructure. To measure this asymmetry, we use
British English as a controlled reference, denoted \RefBrE{}. AmE and BrE
provide well-documented contrasts across orthography, vocabulary, grammar,
writing conventions, collocations, and multi-word usage.\footnote{Representative distinctions
between AmE and BrE are shown in \cref{tab:bre_ame_overview}, and more in Tables~\ref{tab:br-am-core} and~\ref{tab:br-am-phrases}.}
The comparison is also globally meaningful. Through histories of British
colonial expansion, 
BrE conventions became institutionally embedded in
education, government, journalism, and law across many English-using
contexts, including India, Nigeria, Australia, and New Zealand, among many
others~\citep{crystal2003english,0b15e89a04f846adb87cf3cb75ed6a2b}.
Later U.S. economic, cultural, technological, and digital influence expanded
the international reach of AmE~\citep{calabrese2015variation,10.1371/journal.pone.0197741}. These matched
regional forms therefore provide a measurable lens for evaluating
whether AmE receives a consistent advantage across the LLM pipeline.

Methodologically, we ground the analysis in \texttt{1,813} matched
\AmE{}--\RefBrE{} variants and introduce \textsc{DiAlign}, a dynamic,
training-free method for estimating regional alignment from lexical,
grammatical, structural, stylistic, and multi-word evidence
(\sref{sec:alignment-score}). We evaluate three stages:
\rqnum{1} \textbf{Data exposure}, across six pretraining corpora and
21 public post-training datasets
(\sref{sec:audit-pretraining-corpus});
\rqnum{2} \textbf{Representation}, through tokenization, tokenizer
provenance, semantic-equivalence controls, and model prediction
cost~(\sref{sec:regional-tokenizers}); and
\rqnum{3} \textbf{Generation}, under neutral English and explicit
British-English prompting~(\sref{sec:evaluate-generation}). The evidence
follows the same direction across all three stages. AmE is favored in all
six audited pretraining corpora and all 21 post-training datasets; its forms
are generally tokenized more compactly; matched \RefBrE{} counterfactuals
receive higher per-token prediction cost across all ten evaluated checkpoints
under the strict equal-token-count control; and AmE remains the dominant
generation default under neutral English prompting. \RefBrE{} prompting
shifts this preference, but does not consistently eliminate
the AmE default. The downstream preference is therefore accompanied by
upstream asymmetries in data exposure and representation, revealing a
pipeline-wide pattern. The framework also identifies where component-level
interventions can be directed, motivating practical recommendations for
corpus construction and filtering, tokenizer design and provenance auditing,
post-training data selection, and model evaluation
(\sref{sec:discussion-recommendations}). We organize the investigation
around three core research questions.

\vspace{1.5mm}
\begin{tcolorbox}[
    colback=gray!2,
    colframe=black,
    title=Research Questions,
    fonttitle=\bfseries
]
\small

\textbf{RQ1: Data Exposure.}
\emph{Where does an American English advantage appear in the data used to train and align LLMs?}
We audit pretraining and post-training data to measure AmE representation
(\sref{sec:audit-pretraining-corpus}).

\vspace{0.3em}

\textbf{RQ2: Representation.}
\emph{Does the AmE data advantage persist in tokenization and prediction cost?}
We examine tokenization, provenance, semantic-equivalence controls,
and paired model likelihood (\sref{sec:regional-tokenizers}).

\vspace{0.3em}

\textbf{RQ3: Generation.}
\emph{When LLMs generate English, does American English emerge as the default?}
We compare neutral English and British-English prompting across different datasets and
registers (\sref{sec:evaluate-generation}).

\end{tcolorbox}


\section{Methodology: Resources and Alignment}
\label{sec:methodology}

We introduce \textsc{DiAlign}, a dynamic, training-free method for estimating regional alignment from contrastive distributional evidence. The method compares a study variety against a reference distribution; here, American English (AmE) is the study variety and British English the controlled reference, Ref(BrE). Rather than scoring documents by topic or style, it decomposes text into overlapping local spans and aggregates
lexical, grammatical, structural, stylistic, and multi-word evidence. The
procedure generalizes to other variety pairs with comparable reference
distributions (Appendix~\ref{sec:dialign-generalization}).

\vspace{-1.5mm}
\paragraph{Regional Variant Corpus.}
We construct a curated corpus of \texttt{1,813} matched
AmE--Ref(BrE) variants for consistent analysis across the pipeline. The pairs were manually compiled from authentic linguistic references and web-based lexicons (Table~\ref{tab:lexicon-sources} of Appendix) and cover orthographic and lexical differences. Construction, filtering, and categorization details are provided in Appendix~\ref{sec:variant-corpus}.

\vspace{-1.25mm}
\subsection{\textbf{\textsc{DiAlign}}: Regional Alignment Score}
\label{sec:alignment-score}

Given a tokenized input $x=(t_1,\dots,t_N)$ of length $N$,
\textsc{DiAlign} returns $(P_{\AmE},P_{\RefBrE})$, representing relative
alignment with the study and reference varieties. The procedure has four
stages.

\vspace{-1.25mm}
\paragraph{$n$-gram Extraction.}
We extract all contiguous $n$-grams from $x$ for $2 \leq n \leq 5$:
\[
\begin{aligned}
\mathcal{G}(x)
&=
\bigcup_{n=2}^{5}
\left\{
g=(t_i,\ldots,t_{i+n-1})
\mid
1 \leq i \leq N-n+1
\right\},\\
|\mathcal{G}(x)|
&=
\sum_{n=2}^{5}(N-n+1)
=
4N-10,
\qquad N\geq5.
\end{aligned}
\]
Candidate spans are then filtered, providing dense local evidence even for
short inputs. To reduce topical and non-diagnostic signals, we discard spans
$g\in\mathcal{G}(x)$ containing named entities or consisting entirely of
stopwords. Additional filtering details are provided in
Appendix~\ref{sec:dialign-implementation-details}.

\vspace{-1.5mm}
\paragraph{Frequency Lookup.}
For each retained span $g$, we query separate American and British English
corpora in Google Books Ngram\footnote{\url{https://books.google.com/ngrams/}.}
and obtain normalized average yearly frequencies
$f_{\AmE}(g)$ and $f_{\RefBrE}(g)$ over $[y_{\min},y_{\max}]$.
Thus, the same local construction is compared across two variety-specific
frequency distributions under a common scale, rather than scored by raw
frequency in a single corpus. Spans with zero frequency in either reference
are discarded.

\paragraph{Signed Divergence.}
For each retained span, we compute the signed log-frequency ratio:
\[
\mathrm{LR}(g)
=
\log_2\!\left(
\frac{f_{\AmE}(g)}
     {f_{\RefBrE}(g)}
\right).
\]
Positive values favor AmE, negative values favor Ref(BrE), and
$\mathrm{LR}(g)=0$ indicates no directional signal. We down-weight weakly
differentiated spans using $\delta(g)$ and give additional weight to spans
containing an entry from the curated variant lexicon $\mathcal{D}$ using $w(g)$:
\[
\begin{aligned}
\delta(g)
&=
\frac{|f_{\AmE}(g)-f_{\RefBrE}(g)|}
     {f_{\AmE}(g)+f_{\RefBrE}(g)}
\in[0,1),
\qquad
w(g)
&=
\begin{cases}
\delta(g)\beta, & g\cap\mathcal{D}\neq\emptyset,\\
\delta(g), & \text{otherwise}.
\end{cases}
\end{aligned}
\]
Here, $\beta>1$ increases the contribution of explicitly
variety-diagnostic evidence.

\paragraph{Aggregation and Normalization.}
We aggregate positive and negative evidence separately:
\[
S_{\AmE}
=
\sum_{\substack{g\in\mathcal{G}(x)\\ \mathrm{LR}(g)>0}}
\mathrm{LR}(g)\,w(g),
\qquad
S_{\RefBrE}
=
\sum_{\substack{g\in\mathcal{G}(x)\\ \mathrm{LR}(g)<0}}
|\mathrm{LR}(g)|\,w(g).
\]
We then normalize:
\[
P_{\AmE}
=
\frac{S_{\AmE}}{S_{\AmE}+S_{\RefBrE}},
\qquad
P_{\RefBrE}
=
\frac{S_{\RefBrE}}{S_{\AmE}+S_{\RefBrE}}.
\]
The predicted alignment is
\[
\hat{y}(x)
=
\arg\max_{d\in\{\AmE,\RefBrE\}} P_d.
\]

\begin{wrapfigure}{r}{0.3455\columnwidth}
    \centering
    \vspace{-9.5mm}
    \includegraphics[
        width=\linewidth,
        keepaspectratio
    ]{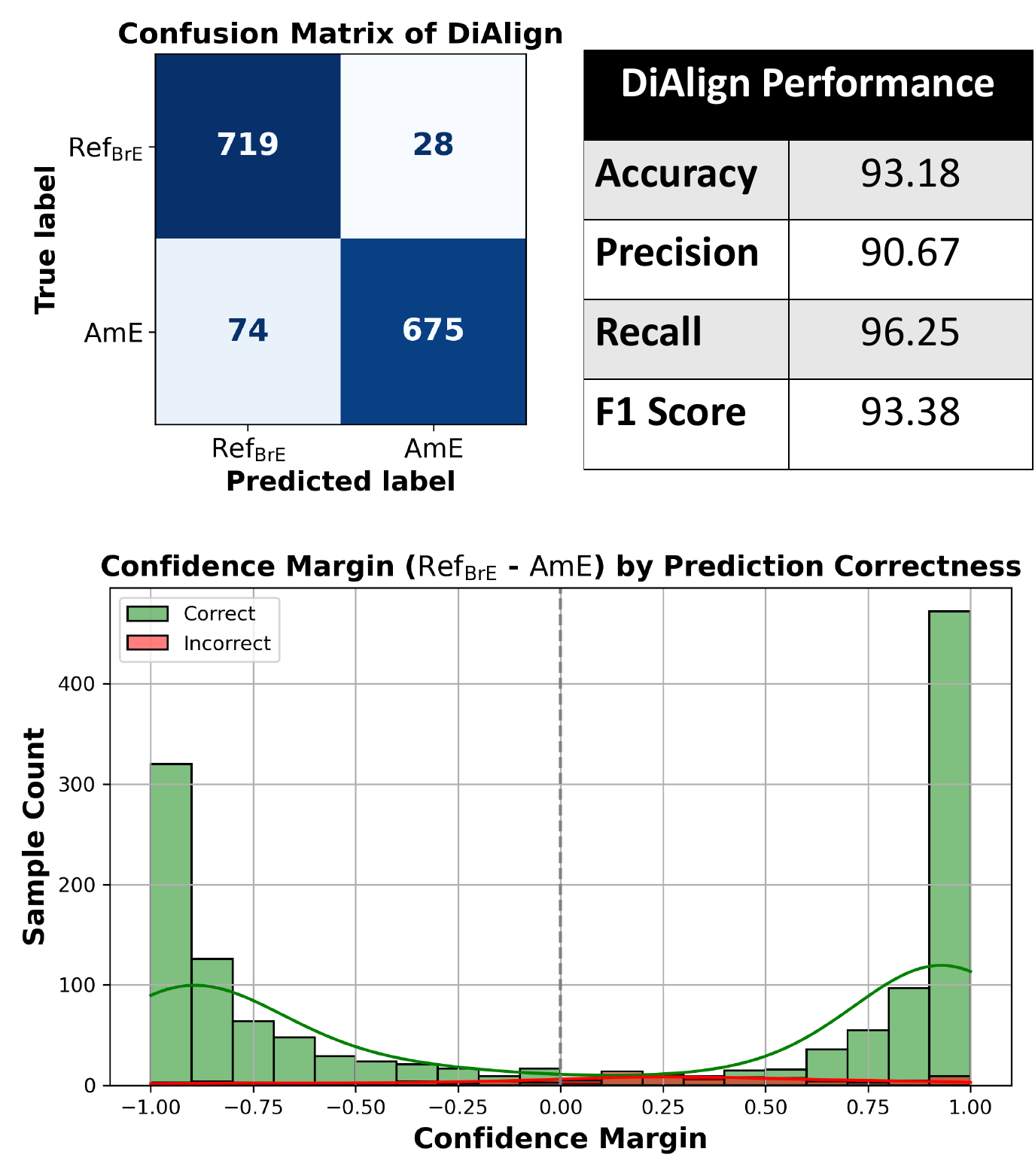}
    \vspace{-7mm}
    \caption{\small
    Meta-evaluation of \textsc{DiAlign} on a balanced
    AmE--Ref(BrE) benchmark, showing the confusion matrix,
    classification performance, and confidence-margin distributions.
    }
    \label{fig:DiAlign-performance}
    \vspace{-7mm}
\end{wrapfigure}

\paragraph{Meta-evaluation.}
We evaluate \textsc{DiAlign} on a balanced benchmark of \texttt{1,500} news
passages: 750 American English texts from HuffPost U.S. News and 750 British
English texts from BBC England, using Google Books reference frequencies
from \texttt{1950--2022}. \textsc{DiAlign} achieves
\textbf{93.18\% accuracy}, 90.67\% precision, 96.25\% recall, and 93.38 F1
(Figure~\ref{fig:DiAlign-performance}). Correct predictions concentrate at
larger confidence margins, while most errors occur near the decision
boundary. Restricting the reference window to \texttt{2000--2022} yields
similar performance, with \textbf{92.91\% accuracy} and 93.12 F1, indicating temporal robustness. Benchmark construction, ablations, confidence analysis, and full temporal-robustness results are reported in
Appendix~\ref{sec:dialign-meta-eval-details}.

Implementation details are provided in
Appendix~\ref{sec:dialign-implementation-details}, with a span-level
walkthrough in Appendix~\ref{sec:dialign-walkthrough}. Benchmark construction,
full validation results, ablations, analyses, and
temporal robustness are reported in
Appendix~\ref{sec:dialign-meta-eval-details}. Generalization beyond the
AmE--Ref(BrE) setting is discussed in Appendix~\ref{sec:dialign-generalization}.


\section{RQ1: Data Exposure Across Pretraining and Post-Training}
\label{sec:audit-pretraining-corpus}

Our goal is to determine whether American English is systematically
overrepresented in the data used to train and align LLMs. We examine two
stages of data exposure: \textbf{(1)} six open-access pretraining corpora and \textbf{(2)} 21 public post-training datasets. Using the matched lexical variants from
\cref{sec:methodology}, we measure orthographic and vocabulary differences
from corpus frequencies. For longer pretraining text, we also apply
\textsc{DiAlign} (\sref{sec:alignment-score}) to capture broader grammatical,
structural, stylistic, and multi-word variation. This audit establishes
whether an AmE advantage is already visible in the data.

\subsection{Pretraining Data Audit}

\paragraph{Methodology.}
For each lexical pair, we extract corpus frequencies
$f_{\AmE}$ and $f_{\RefBrE}$ and compute
\begin{equation}
\label{eq:variant-prob}
P_{\AmE}
=
\frac{f_{\AmE}}
     {f_{\AmE}+f_{\RefBrE}},
\qquad
P_{\RefBrE}
=
\frac{f_{\RefBrE}}
     {f_{\AmE}+f_{\RefBrE}}.
\end{equation}
These pair-level probabilities measure the relative prevalence of matched
forms within each corpus. We aggregate them by corpus and report orthographic
and vocabulary variants separately for direct comparison. To test directional
consistency, we apply the Wilcoxon signed-rank test to paired frequency
differences $(f_{\AmE}-f_{\RefBrE})$, following standard practice for skewed
corpus distributions~\citep{dror-etal-2018-hitchhikers}. All six corpora differ
significantly from parity ($p<0.01$). Dataset and preprocessing details are
provided in Appendices~\ref{sec:pretraining-datasets}
and~\ref{sec:preprocessing-pipeline}.

\paragraph{All six pretraining corpora favor AmE.}
Table~\ref{tab:pretraining-data-results} shows a consistent AmE skew across
all audited corpora. The strongest asymmetry appears in orthographic
variants, where AmE spellings often exceed 70\% of observed pair
frequencies. Vocabulary contrasts are more balanced but still lean toward
AmE. \textsc{DiAlign} shows the same direction for broader grammatical,
structural, stylistic, and multi-word evidence, indicating that the pattern
extends beyond isolated lexical choices.

\input{tables/pretraining-data-results}

\begin{figure*}[t]
    \centering
    \includegraphics[width=\linewidth]{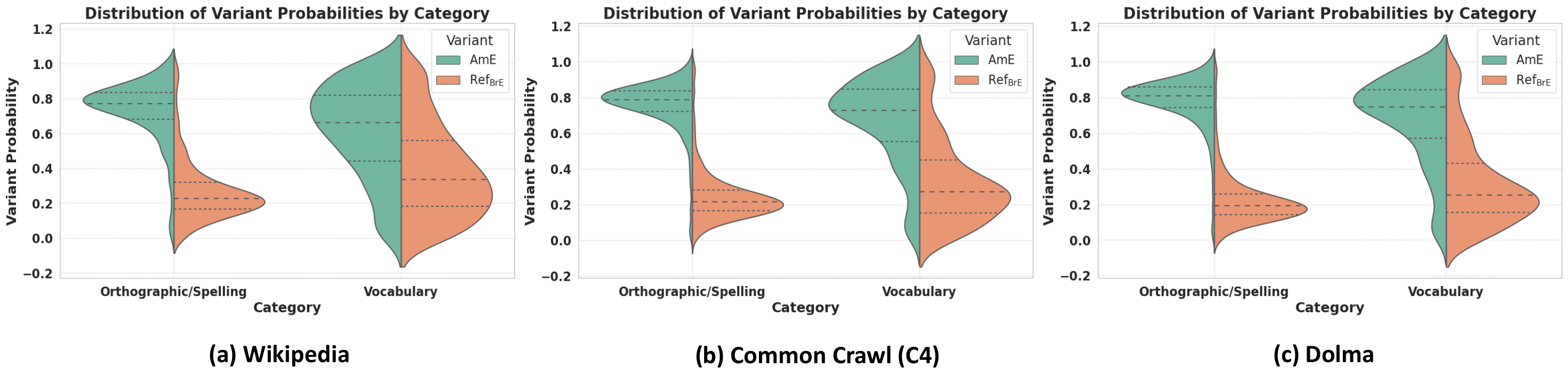}
    \vspace{-4.5mm}
    \caption{\small Distribution of AmE and Ref(BrE) variant probabilities across three pretraining corpora, shown separately for orthographic and vocabulary contrasts. Probabilities are computed from corpus frequencies over the matched variant pairs. Orthographic contrasts exhibit the strongest AmE skew, while vocabulary contrasts are more dispersed but lean toward AmE. Results for the remaining corpora are provided in Appendix Figure~\ref{fig:violin-others}.
    \vspace{-4mm}
    }
    \label{fig:violin-main}
\end{figure*}

\paragraph{Orthographic skew is stronger and more systematic than vocabulary skew.}
Figure~\ref{fig:violin-main} shows orthographic variants clustering more
strongly toward AmE across the corpora, while vocabulary variants are more
dispersed. The finer-grained breakdown in Appendix
Figure~\ref{fig:heatmap-main} shows the same pattern across ten linguistic
subcategories, including \texttt{-ize}/\texttt{-ise} and
\texttt{-og}/\texttt{-ogue}. Most favor AmE;
\texttt{-og}/\texttt{-ogue} is comparatively balanced, consistent with
continued use of forms such as \texttt{catalogue} and \texttt{dialogue} in
some American formal writing~\citep{neumann2023catalogue}.

\subsection{Post-Training Data Audit}
\label{sec:posttraining_audit}

Post-training introduces another major source of linguistic exposure after pretraining through instruction tuning, chat demonstrations, human feedback, preference data, reward modeling, and safety alignment. We audit 21 widely used public datasets spanning these stages. Because many post-training samples are short, we use the explicit matched-variant probabilities as defined in Eq.~\ref{eq:variant-prob}.

\paragraph{All 21 post-training datasets favor AmE.}
Across \texttt{11,574,254} samples, we identify
\texttt{28,860,119} explicit variant occurrences, of which
\textbf{76.17\%} are AmE. Every category in
Table~\ref{tab:posttraining_stage} favors AmE. Instruction-tuning and
task-mixture data are 75.95\% and 71.77\% AmE, respectively, while chat,
human-feedback, reward-model feedback, and safety-preference data reach
84.35\%, 83.08\%, 84.72\%, and 86.86\%. Preference, reward-modeling, and
safety resources contain \texttt{4,666,867} variant occurrences and are
\textbf{80.94\%} AmE overall.

\input{tables/posttraining_stage}

\paragraph{The AmE skew persists across datasets despite varying magnitude.}
All 21 datasets favor AmE, ranging from 65.29\% in Databricks Dolly 15k to
89.72\% in WizardLM Evol-Instruct v2
(Appendix~\ref{tab:posttraining_dataset_full}). As in pretraining,
orthographic contrasts show a stronger AmE preference than vocabulary
contrasts. The same directional pattern across heterogeneous resources
indicates that the skew is not confined to a single post-training dataset
or data type.

\vspace{-1.75mm}
\paragraph{AmE overrepresentation spans both data stages.}
AmE forms are favored in all six pretraining corpora and all 21 post-training datasets. Although the magnitude varies across resources, the direction remains stable, establishing data exposure as the first stage where the AmE advantage is systematically visible.

\vspace{2.95mm}
\begin{keybox}
\small
\textbf{Key Takeaway:}
American English is clearly overrepresented across pretraining and post-training
data, establishing an exposure asymmetry before downstream representation in RQ2 and
generation in RQ3.
\end{keybox}


\section{RQ2: Regional Representation and Prediction-Cost Asymmetry}
\label{sec:regional-tokenizers}

Our aim is to determine whether the AmE advantage observed in data exposure
is also visible in how LLMs \emph{represent} and \emph{predict} matched
regional forms. We examine two complementary levels: whether tokenizers encode
\AmE{} variants more compactly than \RefBrE{} counterparts, and whether
language models assign lower prediction cost to \AmE{} forms in otherwise
equivalent contexts. This separates tokenizer-level representational
asymmetry from preference expressed by the model itself.

\subsection{Tokenization: Representation Efficiency}

We examine how public tokenizers encode the matched
\AmE{}--\RefBrE{} variants. Our primary measure is \emph{fertility}, the
average number of subword tokens required to encode a word
~\citep{rust-etal-2021-good}. Lower fertility indicates more compact
representation as averages can obscure fragmented forms, we also
measure \emph{granularity}: how often variants are represented by 1, 2, or
3+ subwords. Fertility captures average encoding cost, while granularity
reveals its distribution across regional forms.

\input{tables/tokenizer-fertility}

\begin{figure*}[t]
    \centering
    \includegraphics[width=0.985\linewidth]{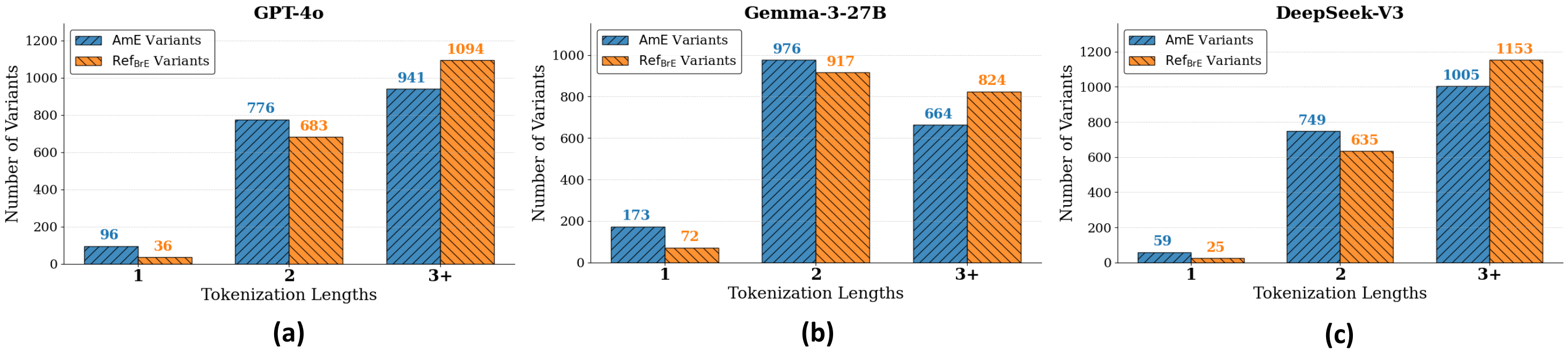}
    \vspace{-3.5mm}
    \caption{\small Tokenization granularity for \AmE{} and \RefBrE{} variants across three tokenizers. Each panel shows the number of variants represented by 1, 2, or 3+ subwords. \RefBrE{} forms appear more frequently in the 3+ subword category, indicating greater long-tail fragmentation. Remaining tokenizers appear in Appendix~Fig. \ref{fig:tokenizer-granularity-app}.
    \vspace{-3.5mm}
    }
    \label{fig:tokenizer-granularity}
\end{figure*}

\paragraph{AmE forms are generally represented more compactly.}
Across the evaluated tokenizers,
\cref{tab:tokenizer-fertility,fig:tokenizer-granularity} show that
\RefBrE{} variants require more subword tokens than matched
\AmE{} forms. The gap is largest for vocabulary contrasts, reaching
\(\Delta_v=18.72\%\), while orthographic gaps among AmE-favoring tokenizers
range from \(\Delta_o=2.85\%\) to \(5.42\%\). Granularity shows the same
pattern: \RefBrE{} forms occur more often in the 3+ token category,
indicating greater long-tail fragmentation. The magnitude varies across
tokenizers. Velvet slightly favors \RefBrE{} orthographic forms
(\(\Delta_o=-0.69\%\)); DeepSeek has the smallest vocabulary gap
(\(\Delta_v=12.66\%\)); and Gemma has the lowest fertility for both forms,
consistent with its larger 262K-token vocabulary. The pattern is
therefore more compact representation of \AmE{}, not uniform behavior across
tokenizer families.

\begin{wrapfigure}{r}{0.55\columnwidth}
    \centering
    \vspace{-5mm}
    \includegraphics[
        width=\linewidth,
        keepaspectratio
    ]{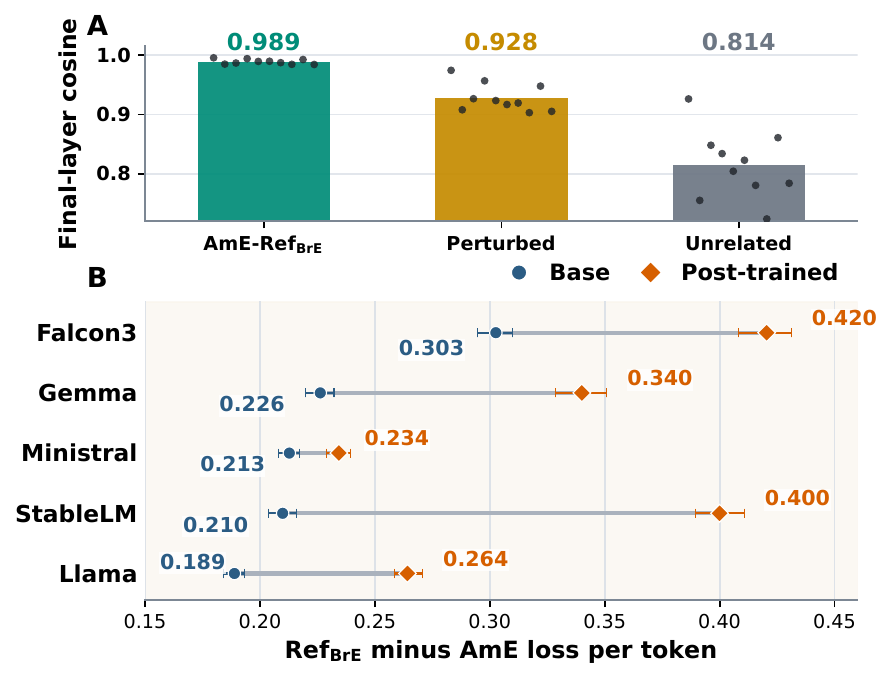}
    \vspace{-6mm}
    \caption{\small \textbf{Semantic equivalence, unequal prediction cost.} Top: matched \AmE{}--\RefBrE{} counterfactuals are more similar than character-perturbed and unrelated controls. Bottom: \RefBrE{} counterfactuals receive higher per-token loss across all five base/post-trained model families under equal sentence token counts. Error bars show paired-bootstrap 95\% CIs.
    }
    \label{fig:rq2-semantic-loss}
    \vspace{-4mm}
\end{wrapfigure}

\paragraph{Tokenizer reuse can transmit representational behavior across model families.}
Tokenizer provenance provides context for these differences. Of the nine
tokenizers evaluated, seven use independent or source vocabularies, while
Llama-3.3 and StableLM-2 inherit or extend U.S.-origin tokenizer artifacts.
StableLM-2 provides the clearest case: despite its association with a
UK-developed model, its tokenizer shows \textbf{100\% token-boundary identity}
with GPT-4 on both the \AmE{} and \RefBrE{} variant sets, with all
\texttt{100,256} shared base tokens retaining identical rank/id assignments.
We therefore separate tokenizer provenance from model developer location.
Tokenizer reuse provides a concrete mechanism through which representational
conventions can persist across distinct model families. Full vocabulary-overlap,
rank-identity, and token-boundary analyses are reported in
Appendix~\ref{app:tokenizer_provenance}.

\subsection{Prediction Cost: Model Preference Beyond Tokenization}

Compact segmentation does not by itself show which regional form a model
finds more probable. We therefore construct paired counterfactual sentences
$(x_i^{\AmE},x_i^{\RefBrE})$ that differ only in the target regional form and compare prediction costs across five base/post-trained model pairs (10 checkpoints).

\paragraph{Methodology.}
For model \(m\), we define per-token negative log-likelihood as
\[
\ell_m(x)
=
-\frac{1}{|\tau_m(x)|}
\sum_{t=1}^{|\tau_m(x)|}
\log p_m(w_t\mid w_{<t}),
\qquad
\Delta^{\mathrm{loss}}_{m,i}
=
\ell_m(x_i^{\RefBrE})-\ell_m(x_i^{\AmE}).
\]
Here, \(\tau_m(x)\) is the model token sequence. Positive
\(\Delta^{\mathrm{loss}}_{m,i}\) indicates higher prediction cost for the
\RefBrE{} counterfactual; for confirmatory analysis, we retain pairs with identical sentence token counts,
\[
|\tau_m(x_i^{\AmE})|
=
|\tau_m(x_i^{\RefBrE})|,
\]
so the remaining loss difference cannot be attributed to unequal token
counts.

We also verify semantic equivalence. Let \(h_{m,\ell}(x)\) denote the
mean-pooled contextual representation at layer \(\ell\). For each pair,
\[
\mathrm{sim}_{m,\ell,i}
=
\cos\!\left(
h_{m,\ell}(x_i^{\AmE}),
h_{m,\ell}(x_i^{\RefBrE})
\right),
\]
which we compare against character-perturbed and unrelated controls. This serves as a validity check for the paired prediction-cost experiment.

\paragraph{Semantically equivalent forms receive systematically different prediction costs.}
Matched \AmE{}--\RefBrE{} counterfactuals are substantially closer in
contextual representation space than character-perturbed or unrelated
controls (\cref{fig:rq2-semantic-loss}), supporting their semantic
equivalence. Yet all ten evaluated checkpoints assign higher per-token loss
to \RefBrE{} on the strict equal-token-count slice, with all paired-bootstrap
95\% confidence intervals excluding zero. Because sentence token count is
held fixed, this prediction-cost gap is distinct from the fragmentation
difference observed in tokenization. The AmE advantage therefore appears in
both segmentation and model-assigned probability.

\paragraph{The prediction-cost gap is larger in every evaluated post-trained checkpoint.}
Across the five model families with matched public base and post-trained
checkpoints, the \RefBrE{} loss gap is larger after post-training in every
family: Gemma, Ministral, Llama, StableLM, and Falcon3. This is a consistent
checkpoint-level pattern; isolating a causal post-training contribution
requires controlled intervention. Full results, bootstrap intervals, and
controlled slices are provided in Appendix~\ref{app:rq2-loss};
semantic-equivalence diagnostics in Appendix~\ref{app:rq2-semantic}; and
model-selection details in Appendix~\ref{app:rq2-model-scope}.

\vspace{2.95mm}
\begin{keybox}
\small
\textbf{Key Takeaway:}
The AmE advantage appears at two representation levels: \AmE{} forms are
encoded more compactly, while semantically matched \RefBrE{} forms
incur higher prediction cost even at equal token counts. Tokenizer lineage
shows how inherited vocabularies can carry representational behavior
across model families.
\end{keybox}


\section{RQ3: Regional Preference in Generation}
\label{sec:evaluate-generation}

Our goal is to determine whether the AmE advantage observed in data exposure
(RQ1) and representation (RQ2) is reflected in generated language. We use
\textsc{DiAlign} (\sref{sec:alignment-score}) to measure regional alignment
under neutral English prompting and explicit British-English prompting.

\input{tables/generation-alignment-results}

\paragraph{Experimental Setup.}
We evaluate open-domain question answering across two contrasting registers:
Natural Questions (NQ)~\citep{kwiatkowski-etal-2019-natural} as a more
formal, encyclopedic setting, and ELI5~\citep{fan-etal-2019-eli5} as a more
informal, conversational setting. To avoid lexical priming, we remove
questions containing variants from the AmE--Ref(BrE) lexicon. For each
question, we generate responses under two conditions: \texttt{English} and
\texttt{British English (en-GB)}, requesting 50 words to keep response length
comparable across models and conditions for controlled comparison.

Even at this length, overlapping 2--5-grams provide substantial local
evidence for \textsc{DiAlign} (\sref{sec:alignment-score}). Alignment is also
effectively uncorrelated with response length ($r=0.017$ overall;
Appendix~\ref{app:rq3-setup}). For each response, \textsc{DiAlign} returns
$(P_{\AmE},P_{\RefBrE})$. Table~\ref{tab:generation-alignment-results}
reports the percentage of generations classified as AmE and mean AmE
alignment $P_{\AmE}$ in brackets, denoted
$\mathrm{AmE}^{\mathrm{Default}}$ and $\mathrm{AmE}^{\mathrm{BrE}}$ for the
two prompting conditions. Dataset filtering, prompts
(Figure~\ref{fig:rq3-prompt}), model details, decoding parameters, and
length-robustness analyses are provided in Appendix~\ref{app:rq3-setup}.

\paragraph{AmE is the dominant generation default under neutral English prompting.}
Table~\ref{tab:generation-alignment-results} shows a consistent AmE preference when models are instructed to respond in English. Across most models and both datasets, a majority of outputs are classified as AmE, often with strong alignment scores. The upstream asymmetries identified in RQ1 and RQ2 therefore also appear in realized behavior: without regional steering, generated responses align more strongly with AmE than with \RefBrE{}.

\paragraph{British-English prompting shifts, but does not consistently eliminate, the AmE default.}
Requesting \texttt{British English (en-GB)} shifts outputs toward
\RefBrE{} across models, but substantial AmE alignment often remains,
particularly in the more formal NQ setting. The magnitude of this shift
varies across conditions, models, and registers, showing that downstream
prompting can alter realized language without consistently removing the AmE
default. Such output-level steering does not alter the upstream asymmetries
already empirically established in RQ1 and RQ2.

\paragraph{The strength of the AmE default varies across models and registers.}
The AmE preference is generally stronger in NQ than ELI5, while
British-English prompting produces larger shifts for several models in the
more conversational ELI5 setting. For example, Llama-3.3 falls to about
30\% AmE-classified outputs on ELI5 under British-English prompting. The
default is therefore not uniform: its magnitude depends on both the specific
model and local linguistic context.

\vspace{2.5mm}
\begin{keybox}
\small
\textbf{Key Takeaway:}
American English is the dominant generation default under neutral English
prompting. British-English prompting shifts outputs toward \RefBrE{} but
does not consistently eliminate the AmE default, with the magnitude of the
shift varying across models and registers. RQ3 provides clear downstream
behavioral evidence for the AmE advantage already established in data
exposure (RQ1) and representation (RQ2).
\end{keybox}


\section{Related Work}
\label{sec:related-work-main}

\paragraph{Training data audits and curation.}
Prior work examines how dataset composition, quality, provenance,
and curation shape model behavior. Studies show that data age,
coverage, and quality affect generalization
~\citep{longpre-etal-2024-pretrainers}, while audits identify
duplication, contamination, and provenance gaps in widely used corpora
~\citep{elazar2024whats,longpre2025bridging}. Other work studies corpus
construction~\citep{parmar-etal-2024-data}, register- and domain-aware
sampling~\citep{myntti2025register}, and reuse of filtered web data
~\citep{nguyen2025recycling}. We add within-English variation to this
auditing agenda by measuring AmE representation across pretraining and
post-training data.

\paragraph{Tokenization and representation.}
Tokenization can encode systematic differences before generation begins.
Prior work documents substantial variation in tokenization efficiency across
languages, with consequences for sequence length, context capacity, and
inference cost
~\citep{petrov2023language,rust-etal-2021-good,
ahia-etal-2023-languages,ali-etal-2024-tokenizer}. Related studies connect
tokenization to demographic bias~\citep{an-rudinger-2023-nichelle},
lexical fragility~\citep{gallifant-etal-2024-language}, and gender bias in
machine translation~\citep{iluz-etal-2023-exploring}. We study this problem
within English by combining tokenization efficiency, tokenizer provenance,
and paired model likelihood for matched AmE and Ref(BrE) forms.

\paragraph{English variation in NLP and LLMs.}
Research on African American English (AAE), South Asian Englishes (SAsE),
and other English varieties has shown systematic disparities in tagging,
classification, sentiment analysis, robustness, and model behavior
~\citep{jorgensen-etal-2016-learning,blodgett-etal-2016-demographic,
kiritchenko-mohammad-2018-examining,ziems-etal-2023-multi,Hofmann2024,
fleisig-etal-2024-linguistic,holt-etal-2024-perceptions}. We focus on a different question: how an
American-English default appears across the LLM pipeline. By linking
training data, tokenization and prediction cost, and generation, our analysis
treats regional preference as a structural property that can be observed
across multiple stages of model development. An extended discussion is provided in
Appendix~\ref{sec:extended-related-work}.


\vspace{-0.6mm}
\section{Conclusion}
\label{sec:conclusion}
\vspace{-0.6mm}
This paper presents the first rigorous pipeline-wide study of
\emph{structural bias} across phases of LLM development. By
triangulating pretraining and post-training data, tokenizer behavior and
provenance, model prediction cost, and generated language, we show that
American English receives an advantage across data exposure,
representation, and generation, emerging as the \emph{de facto} default.
These findings show how linguistic asymmetries persist across the LLM
pipeline, raising concerns about linguistic homogenization, epistemic
injustice, and inequity in global AI deployment, while providing a basis for
targeted, component-level improvements toward more inclusive language
technologies.



\input{sections/Ethics}


\bibliography{arxiv_neutral}
\bibliographystyle{arxiv_neutral}


\input{sections/Appendix}


\end{document}

%% file: math_commands.tex
\usepackage{amsmath,amsfonts,bm}

\def\eqref#1{equation~\ref{#1}}

\def\1{\bm{1}}

\def\rq{{\textnormal{q}}}

\DeclareMathAlphabet{\mathsfit}{\encodingdefault}{\sfdefault}{m}{sl}
\SetMathAlphabet{\mathsfit}{bold}{\encodingdefault}{\sfdefault}{bx}{n}



%% file: tables/pretraining-data-results.tex
\begin{table*}[t]
\centering
\renewcommand{\arraystretch}{1.2}
\setlength{\tabcolsep}{4pt}
\small
\caption{\small Variant usage across six pretraining corpora. Under the \textbf{Lexical Variants} analysis, we report corpus-level distributions for orthographic (spelling-based) and vocabulary-based AmE--Ref(BrE) differences. We also report \textbf{\textsc{DiAlign}} scores, which capture broader regional alignment from grammatical, structural, collocational, and multi-word evidence; bracketed values denote mean alignment confidence. Datasets marked with * are evaluated on sampled
subsets. RedPajama and Dolma contain mixed-domain content, including
\iconCode~code, \iconPapers~papers, \iconStackExchange~forums, and
\iconSocial~social media. All probabilities are reported as percentages.
Token statistics were computed using the LLaMA tokenizer
\citep{grattafiori2024llama3}.
\vspace{-2.15mm}
}
\resizebox{\linewidth}{!}{
\begin{tabular}{llccccccccc}
\hline
\rowcolor[HTML]{F5F5F5}
&
&
&
&
\multicolumn{4}{c}{\textbf{Lexical Variants}} &
\multicolumn{3}{c}{\textbf{DiAlign (broader variants)}} \\ \cline{5-8} \cline{9-11}

\rowcolor[HTML]{F5F5F5}
\textbf{Data Source} &
\textbf{Document Type} &
\textbf{Documents} &
\textbf{Tokens} &
\multicolumn{2}{c}{\textbf{Orthographic}} &
\multicolumn{2}{c}{\textbf{Vocabulary}} &
\multicolumn{3}{c}{\textbf{\% of}} \\ \cline{5-8} \cline{9-11}

\rowcolor[HTML]{F5F5F5}
&
&
\textit{(millions)} &
\textit{(billions)} &
\textbf{AmE} (\usflag) &
\textbf{\bRefBrE} (\ukflag) &
\textbf{AmE} (\usflag) &
\textbf{\bRefBrE} (\ukflag) &
\textbf{AmE} (\usflag) &
&
\textbf{\bRefBrE} (\ukflag) \\ \hline \hline

Book Corpus (\citeyear{10.1109/ICCV.2015.11})   & \iconBooks books        & 74    & 1.28 & 86.81 & 13.19 & 75.00 & 25.00 & 79.90 {\small \color{gray}[0.79]} &  & 20.10 {\small \color{gray}[0.76]} \\
Wikipedia (\citeyear{wikidump})                 & \iconRefs encyclopedic & 6.4   & 4.3  & 72.94 & 27.06 & 61.43 & 38.57 & 72.18 {\small \color{gray}[0.80]} &  & 27.82 {\small \color{gray}[0.78]} \\
Common Crawl (C4) (\citeyear{10.5555/3455716.3455856}) & \iconWeb web pages & 365   & 156  & 75.12 & 24.88 & 67.00 & 33.00 & 73.67 {\small \color{gray}[0.81]} &  & 26.33 {\small \color{gray}[0.81]} \\
Falcon RefinedWeb (\citeyear{penedo2023the})    & \iconWeb web pages     & 968   & 600  & 77.34 & 22.66 & 68.35 & 31.65 & 74.04 {\small \color{gray}[0.82]} &  & 25.96 {\small \color{gray}[0.75]} \\
RedPajama* (\citeyear{weber2024redpajama})      & \iconBooks \ \iconRefs \ \iconWeb \ \iconCode \ \iconPapers \ \iconStackExchange mixed & 0.93  & 1.0  & 76.03 & 23.97 & 66.05 & 33.95 & 72.38 {\small \color{gray}[0.85]} &  & 27.62 {\small \color{gray}[0.75]} \\
Dolma* (\citeyear{soldaini-etal-2024-dolma})    & \iconBooks \ \iconRefs \ \iconWeb \ \iconCode \ \iconPapers \ \iconSocial mixed & 14.28 & 10   & 77.30 & 22.70 & 67.77 & 32.23 & 75.13 {\small \color{gray}[0.81]} &  & 24.87 {\small \color{gray}[0.77]} \\ \hline

\end{tabular}
}
\vspace{-1.25mm}
\label{tab:pretraining-data-results}
\end{table*}

%% file: tables/posttraining_stage.tex
\begin{wraptable}{r}{0.512\textwidth}
\vspace{-6mm}
\centering
\small
\setlength{\tabcolsep}{1.5pt}
\renewcommand{\arraystretch}{0.95}
\caption{\small
\textbf{American English is consistently overrepresented across public
post-training data.} Results aggregate 21 datasets spanning instruction
tuning, human feedback, preference, reward modeling, and safety.
\textbf{Variant hits} counts matches to either form in the curated
AmE--Ref(BrE) pairs, and \textbf{AmE (\%)} reports the weighted AmE share.
Dataset-level results 
in Appendix Table~\ref{tab:posttraining_dataset_full}.}
\vspace{-2.5mm}

\begin{tabular}{%
p{0.58\linewidth}
>{\centering\arraybackslash}p{0.21\linewidth}
>{\centering\arraybackslash}p{0.18\linewidth}}
\toprule
\textbf{Post-training stage} &
\textbf{Hits} &
\textbf{AmE (\%)} \\
\midrule

Instruction tuning / SFT
& 16,348,769 & 75.95 \\

Instruction tuning / chat
& 1,204,637 & 84.35 \\

Instruction tuning / hum feedback
& 61,030 & 83.08 \\

Instruction tuning / task mixture
& 6,578,816 & 71.77 \\

RLHF / preference data
& 4,168,689 & 80.47 \\

RLHF / reward-model feedback
& 77,332 & 84.72 \\

RLHF / safety feedback
& 242,377 & 83.48 \\

RLHF / safety preference
& 178,469 & 86.86 \\

\midrule

\textbf{Preference / reward / safety}
& \textbf{4,666,867} & \textbf{80.94} \\

\textbf{All post-training data}
& \textbf{28,860,119} & \textbf{76.17} \\

\bottomrule
\end{tabular}
\vspace{-6.5mm}

\label{tab:posttraining_stage}
\end{wraptable}

%% file: tables/tokenizer-fertility.tex
\begin{table*}[t]
\centering
\renewcommand{\arraystretch}{1.15} 
\setlength{\tabcolsep}{3.5pt} 
\small 
\caption{\small
Fertility scores for AmE and Ref(BrE) variants across the evaluated
tokenizers, reported separately for orthographic and vocabulary contrasts.
Lower fertility indicates more compact tokenization. $\Delta_o$ and
$\Delta_v$ denote the relative fertility gap between matched AmE and
Ref(BrE) forms. \textbf{Bold} and {\ul underlined} values indicate the
lowest and second-lowest fertility, respectively. All AmE--Ref(BrE)
differences are statistically significant under the Wilcoxon signed-rank
test ($p<0.01$), except for Velvet-2B in the orthographic category, marked
with $\dagger$.
\vspace{-1.25mm}
}
\resizebox{13cm}{!}  
{
\begin{tabular}{lccccccccccc}
\hline
\rowcolor[HTML]{F5F5F5} 
\multicolumn{1}{c}{\cellcolor[HTML]{F5F5F5}} &
  \cellcolor[HTML]{F5F5F5} &
  \cellcolor[HTML]{F5F5F5} &
  \cellcolor[HTML]{F5F5F5} &
   &
  \multicolumn{3}{c}{\cellcolor[HTML]{F5F5F5}\textbf{Orthographic}} &
   &
  \multicolumn{3}{c}{\cellcolor[HTML]{F5F5F5}\textbf{Vocabulary}} \\ \cline{6-8} \cline{10-12} 
\rowcolor[HTML]{F5F5F5} 
\multicolumn{1}{c}{\multirow{-2}{*}{\cellcolor[HTML]{F5F5F5}\textbf{Tokenizers}}} &
  \multirow{-2}{*}{\cellcolor[HTML]{F5F5F5}\textbf{\begin{tabular}[c]{@{}c@{}}Developer \\ Country\end{tabular}}} &
  \multirow{-2}{*}{\cellcolor[HTML]{F5F5F5}\textbf{\begin{tabular}[c]{@{}c@{}}Model\\ Access\end{tabular}}} &
  \multirow{-2}{*}{\cellcolor[HTML]{F5F5F5}\textbf{\begin{tabular}[c]{@{}c@{}}Vocab\\ Size\end{tabular}}} &
   &
  \textbf{AmE (\usflag)} &
  \textbf{\bRefBrE (\ukflag)} &
  \textbf{$\Delta_o$} &
   &
  \textbf{AmE (\usflag)} &
  \textbf{\bRefBrE (\ukflag)} &
  \textbf{$\Delta_v$} \\ \hline
\href{https://platform.openai.com/docs/models/gpt-4}{GPT-4}             & USA (\usflag)    & \closedSource & 100K &  & 2.73 & 2.86 &  \inc{4.76} &  & 2.27 & 2.64 &  \inc{16.30} \\
\href{https://platform.openai.com/docs/models/gpt-4o}{GPT-4o}            & USA (\usflag)     & \closedSource & 200K &  & 2.65 & 2.77 &  \inc{4.53} &  & 2.21 & 2.57 &  \inc{16.29} \\
\hdashline
\href{https://huggingface.co/meta-llama/Llama-3.3-70B-Instruct}{Llama-3.3-70B}     & USA (\usflag)    & \openSource  & 128K &  & 2.72 & 2.85 &  \inc{4.78} &  & 2.27 & 2.63 &  \inc{15.86} \\
\href{https://huggingface.co/google/gemma-3-27b-it}{Gemma-3-27B}       & USA (\usflag)    & \openSource  & 262K &  & \cellcolor{violet!8}\textbf{2.40} & \cellcolor{violet!8}\textbf{2.53} &  \inc{5.42} &  & \cellcolor{violet!8}\textbf{2.02} & \cellcolor{violet!8}\textbf{2.35} &  \inc{16.34} \\ \hline
\href{https://huggingface.co/deepseek-ai/DeepSeek-V3-0324}{DeepSeek-V3}       & China (\chinaflag)  & \openSource  & 128K &  & 2.71 & 2.80 &  \inc{3.32} &  & 2.37 & 2.67 &  \cellcolor{violet!8}\inc{12.66} \\
\href{https://huggingface.co/mistralai/Mistral-Small-24B-Instruct-2501}{Mistral-Small-24B} & France (\franceflag) & \openSource  & 131K &  & 2.81 & 2.89 &  \cellcolor{blue!8}{\ul \inc{2.85}}  &  & 2.45 & 2.79 &  \inc{13.88} \\
\href{https://huggingface.co/stabilityai/stablelm-2-zephyr-1_6b}{StableLM-2-1.6B} & UK (\ukflag) & \openSource  & 100K &  & 2.73 & 2.86 &  \inc{4.76}  &  & 2.27 & 2.64 &  \inc{16.30} \\
\href{https://huggingface.co/Almawave/Velvet-2B}{Velvet-2B}        & Italy (\italyflag)  & \openSource  & 127K &  & 2.90 & 2.88 &  \cellcolor{violet!8}\dec{0.69}$^{\dagger}$ &  & 2.40 & 2.72 &  \cellcolor{blue!8}{\ul \inc{13.33}} \\ 
\href{https://huggingface.co/tiiuae/Falcon3-7B-Instruct}{Falcon3-7B}       & UAE (\uaeflag)    & \openSource  & 131K &  & \cellcolor{blue!8}{\ul 2.44} & \cellcolor{blue!8}{\ul 2.56} &  \inc{4.92} &  & \cellcolor{blue!8}{\ul 2.03} & \cellcolor{blue!8}{\ul 2.41} &  \inc{18.72} \\ \hline
\end{tabular}
}
\vspace{-2.5mm}
\label{tab:tokenizer-fertility}
\end{table*}

%% file: tables/generation-alignment-results.tex
\begin{wraptable}{r}{0.61\textwidth}
\centering
\small
\setlength{\tabcolsep}{2.55pt}
\renewcommand{\arraystretch}{1.34}
\vspace{-5.25mm}
\caption{\small
Regional preferences in LLM generation on Natural Questions
(\emph{formal}) and ELI5 (\emph{informal}). We report the percentage of
outputs classified as AmE under default English and British English
(en-GB) prompting, with mean AmE alignment scores in brackets. AmE is the
dominant default, while British-English prompting generally shifts outputs
toward Ref(BrE). \textbf{Bold} marks the lowest AmE percentage in each
column; {\ul underlined} marks the second-lowest.
}
\vspace{-2mm}
\resizebox{\linewidth}{!}{
\begin{tabular}{lccccccc}
    \hline
    \rowcolor[HTML]{F5F5F5} 
    \multicolumn{1}{c}{\cellcolor[HTML]{F5F5F5}} & \cellcolor[HTML]{F5F5F5} &  & \multicolumn{2}{c}{\cellcolor[HTML]{F5F5F5}{\cellcolor{blue!10}\textbf{Natural Questions {[}\texttt{formal}{]}}}} &  & \multicolumn{2}{c}{\cellcolor[HTML]{F5F5F5}{\cellcolor{cyan!8}\textbf{ELI5 {[}\texttt{informal}{]}}}} \\ \cline{4-5} \cline{7-8} 
    \rowcolor[HTML]{F5F5F5} 
    \multicolumn{1}{c}{\multirow{-2}{*}{\cellcolor[HTML]{F5F5F5}\textbf{LLMs}}} & \multirow{-2}{*}{\cellcolor[HTML]{F5F5F5}\textbf{\begin{tabular}[c]{@{}c@{}}Developer\\ Country\end{tabular}}} &  & \shortstack{\textbf{\begin{tabular}[c]{@{}c@{}}Default English\\ ($\text{AmE}^{\text{Default}}$)\end{tabular}}} & \shortstack{\textbf{\begin{tabular}[c]{@{}c@{}}British English\\ ($\text{AmE}^{\text{BrE}}$)\end{tabular}}} &  & \shortstack{\textbf{\begin{tabular}[c]{@{}c@{}}Default English\\ ($\text{AmE}^{\text{Default}}$)\end{tabular}}} & \shortstack{\textbf{\begin{tabular}[c]{@{}c@{}}British English\\ ($\text{AmE}^{\text{BrE}}$)\end{tabular}}} \\ \hline \hline
    
    \href{https://platform.openai.com/docs/models/gpt-4o}{GPT-4o} & USA (\usflag) &  & 79.00\% {\small \color{gray}[0.81]} & 45.33\% {\small \color{gray}[0.77]} &  & 77.00\% {\small \color{gray}[0.82]} & \cellcolor{blue!8}{\ul 34.67\%} {\small \color{gray}[0.78]} \\
    
    \href{https://cloud.google.com/vertex-ai/generative-ai/docs/models/gemini/2-0-flash}{Gemini-2.0-flash} & USA (\usflag) &  & 76.00\% {\small \color{gray}[0.83]} & 51.00\% {\small \color{gray}[0.79]} &  & 75.33\% {\small \color{gray}[0.82]} & 42.33\% {\small \color{gray}[0.80]} \\
    
    \href{https://www.anthropic.com/news/claude-3-7-sonnet}{Claude-3.7-sonnet} & USA (\usflag) &  & 75.67\% {\small \color{gray}[0.85]} & \cellcolor{blue!8}{\ul 42.33\%} {\small \color{gray}[0.80]} &  & 73.33\% {\small \color{gray}[0.86]} & 37.67\% {\small \color{gray}[0.78]} \\ \hdashline
    
    \href{https://huggingface.co/meta-llama/Llama-3.3-70B-Instruct}{Llama-3.3-70B} & USA (\usflag) &  & 74.67\% {\small \color{gray}[0.82]} & 47.33\% {\small \color{gray}[0.78]} &  & 69.00\% {\small \color{gray}[0.79]} &  \cellcolor{violet!8}\textbf{30.00\%} {\small \color{gray}[0.76]} \\
    
    \href{https://huggingface.co/google/gemma-3-27b-it}{Gemma-3-27B}  & USA (\usflag) &  &  \cellcolor{violet!8}\textbf{69.33\%} {\small \color{gray}[0.81]} & 45.67\% {\small \color{gray}[0.78]} &  & \cellcolor{blue!8}{\ul 68.33\%} {\small \color{gray}[0.83]} & 38.00\% {\small \color{gray}[0.78]} \\ \hline
    
    \href{https://huggingface.co/deepseek-ai/DeepSeek-V3-0324}{DeepSeek-V3} & China (\chinaflag) &  & 74.67\% {\small \color{gray}[0.82]} & \cellcolor{violet!8}\textbf{41.67\%} {\small \color{gray}[0.78]} &  & 73.33\% {\small \color{gray}[0.84]} & 40.47\% {\small \color{gray}[0.76]} \\
    
    \href{https://huggingface.co/mistralai/Mistral-Small-24B-Instruct-2501}{Mistral-Small-24B} & France (\franceflag) &  & 73.67\% {\small \color{gray}[0.81]} & 48.00\% {\small \color{gray}[0.75]} &  & 72.33\% {\small \color{gray}[0.82]} & 38.67\% {\small \color{gray}[0.76]} \\
    
    \href{https://huggingface.co/stabilityai/stablelm-2-zephyr-1_6b}{StableLM-2-1.6B} & UK (\ukflag) &  & 74.00\% {\small \color{gray}[0.79]} & 69.67\% {\small \color{gray}[0.78]} &  & 73.33\% {\small \color{gray}[0.77]} & 67.00\% {\small \color{gray}[0.77]} \\
    
    \href{https://huggingface.co/Almawave/Velvet-2B}{Velvet-2B}  & Italy (\italyflag) &  & \cellcolor{blue!8}{\ul 72.91\%} {\small \color{gray}[0.80]} & 69.33\% {\small \color{gray}[0.81]} &  & 71.00\% {\small \color{gray}[0.80]} & 67.67\% {\small \color{gray}[0.80]} \\
    
    \href{https://huggingface.co/tiiuae/Falcon3-7B-Instruct}{Falcon3-7B} & UAE (\uaeflag) &  & 73.67\% {\small \color{gray}[0.80]} & 66.00\% {\small \color{gray}[0.79]} &  & \cellcolor{violet!8}\textbf{66.00\%} {\small \color{gray}[0.81]} & 63.33\% {\small \color{gray}[0.77]} \\ \hline
\end{tabular}
}
\label{tab:generation-alignment-results}
\vspace{-5.0mm}
\end{wraptable}

%% file: sections/Ethics.tex
\section*{Ethics Statement}

This work studies structural linguistic bias across the LLM development
pipeline. It involves no human subjects, private user data, or sensitive
personal attributes. Our analyses use publicly available pretraining and
post-training datasets for research purposes
(Appendix~\ref{sec:pretraining-datasets}). The curated
\AmE{}--\RefBrE{} resource was compiled from publicly accessible linguistic
and web-based sources (Table~\ref{tab:lexicon-sources}) and contains no
personal or proprietary material. We plan to release it for non-commercial
research under a CC BY-NC-SA 4.0 license.\footnote{\url{https://creativecommons.org/licenses/by-nc-sa/4.0/}}

The ethical motivation is to understand where linguistic advantage becomes
visible, persists, and manifests across data exposure, representation, and
generation. Such defaults can shape which regional forms receive greater
visibility as LLMs are deployed across diverse national and institutional
contexts, with implications for linguistic homogenization and unequal
representation. British English is used solely as a controlled reference for
measuring the American English preference; neither variety is treated as
intrinsically superior, more correct, or more appropriate. Developer country
and tokenizer provenance are likewise analyzed descriptively and are not
interpreted as evidence of intent, nationality, or regional causation.

The pipeline-wide audit and \textsc{DiAlign} are intended for responsible
regional-language auditing and for identifying where targeted improvements
may be most effective. \textsc{DiAlign} excludes spans containing named
entities such as people, organizations, and locations, reducing privacy and
reputational risks from such signals. The framework can be extended beyond
AmE--Ref(BrE) when suitable regional reference distributions and linguistic
resources are available. Scope, limitations, and risks of binary regional
comparison are discussed in Appendix~\ref{sec:limitations}.

%% file: sections/Appendix.tex
\clearpage
\appendix
\onecolumn

\begin{center}
    \Large\bfseries Supplementary Material: Appendices \\[22pt]
\end{center}


\begin{center}
\begin{minipage}{0.92\linewidth}

\noindent{\large\bfseries Appendix Contents}

\vspace{7pt}

\noindent
\hyperref[sec:variant-corpus]{\textbf{A\quad Regional Variant Resource}}

\vspace{2pt}

\hspace*{1.8em}\hyperref[sec:variant-construction]{A.1\quad Construction and Filtering}\\
\hspace*{1.8em}\hyperref[sec:appendix-variant-grouping]{A.2\quad Variant Grouping and Typology}\\
\hspace*{1.8em}\hyperref[sec:variant-sources]{A.3\quad Sources and Extended Examples}

\vspace{5pt}

\noindent
\hyperref[sec:dialign-appendix]{\textbf{B\quad \textsc{DiAlign}: Implementation, Validation, and Generalization}}

\vspace{2pt}

\hspace*{1.8em}\hyperref[sec:dialign-implementation-details]{B.1\quad Implementation Details}\\
\hspace*{1.8em}\hyperref[sec:dialign-walkthrough]{B.2\quad Illustrative Walkthrough}\\
\hspace*{1.8em}\hyperref[sec:dialign-meta-eval-details]{B.3\quad Meta-evaluation Benchmark}\\
\hspace*{1.8em}\hyperref[sec:dialign-validation]{B.4\quad Validation and Ablation}\\
\hspace*{1.8em}\hyperref[sec:dialign-temporal]{B.5\quad Temporal Robustness}\\
\hspace*{1.8em}\hyperref[sec:dialign-generalization]{B.6\quad Generalizing \textsc{DiAlign}}

\vspace{5pt}

\noindent
\hyperref[app:rq1-details]{\textbf{C\quad Additional Details for RQ1: Data Exposure}}

\vspace{2pt}

\hspace*{1.8em}\hyperref[sec:pretraining-datasets]{C.1\quad Pretraining Datasets}\\
\hspace*{1.8em}\hyperref[sec:preprocessing-pipeline]{C.2\quad Pretraining Preprocessing}\\
\hspace*{1.8em}\hyperref[sec:pretraining-scale]{C.3\quad Corpus Processing and Sampling}\\
\hspace*{1.8em}\hyperref[sec:rq1-additional-results]{C.4\quad Additional Pretraining Results}\\
\hspace*{1.8em}\hyperref[app:posttraining_datasets]{C.5\quad Post-Training Dataset Audit}

\vspace{5pt}

\noindent
\hyperref[app:rq2-supplement]{\textbf{D\quad Additional Details for RQ2: Representation and Prediction Cost}}

\vspace{2pt}

\hspace*{1.8em}\hyperref[app:tokenizer-granularity]{D.1\quad Additional Tokenization Granularity}\\
\hspace*{1.8em}\hyperref[app:tokenizer_provenance]{D.2\quad Tokenizer Provenance and Lineage}\\
\hspace*{1.8em}\hyperref[app:tokenizer_lineage_metrics]{D.3\quad Lineage Diagnostics}\\
\hspace*{1.8em}\hyperref[app:tokenizer_reuse]{D.4\quad Behavioral Evidence for Tokenizer Reuse}\\
\hspace*{1.8em}\hyperref[app:rq2-counterfactuals]{D.5\quad Counterfactual Sentence Construction}\\
\hspace*{1.8em}\hyperref[app:rq2-loss]{D.6\quad Prediction-Cost Evaluation}\\
\hspace*{1.8em}\hyperref[app:rq2-model-scope]{D.7\quad Model Scope}\\
\hspace*{1.8em}\hyperref[app:rq2-semantic]{D.8\quad Semantic-Equivalence Control}\\
\hspace*{1.8em}\hyperref[app:rq2-base-post]{D.9\quad Base and Post-Trained Checkpoint Comparison}

\vspace{5pt}

\noindent
\hyperref[app:rq3-setup]{\textbf{E\quad Additional Details for RQ3: Generation}}

\vspace{2pt}

\hspace*{1.8em}\hyperref[app:rq3-data]{E.1\quad Datasets and Sampling}\\
\hspace*{1.8em}\hyperref[app:rq3-prompting]{E.2\quad Prompting and Decoding}\\
\hspace*{1.8em}\hyperref[app:rq3-scoring]{E.3\quad Scoring and Outcome Measures}\\
\hspace*{1.8em}\hyperref[sec:length-control]{E.4\quad Word-Length Adherence}\\
\hspace*{1.8em}\hyperref[app:rq3-length-robustness]{E.5\quad Response-Length Robustness}

\vspace{5pt}

\noindent
\hyperref[sec:discussion-recommendations]{\textbf{F\quad Discussion and Practical Recommendations}}

\vspace{5pt}

\noindent
\hyperref[sec:extended-related-work]{\textbf{G\quad Extended Related Work}}

\vspace{5pt}


\noindent
\hyperref[sec:brief-history]{\textbf{H\quad Brief Historical Context}}

\vspace{5pt}

\noindent
\hyperref[sec:limitations]{\textbf{I\quad Scope, Limitations, and Future Directions}}

\end{minipage}
\end{center}

\vspace{10pt}

\newpage


\section{Regional Variant Resource}
\label{sec:variant-corpus}

\subsection{Construction and Filtering}
\label{sec:variant-construction}

We construct a curated resource of \texttt{1,813} matched
AmE--Ref(BrE) variants for consistent measurement across the three research
questions [RQ1~(\cref{sec:audit-pretraining-corpus}), RQ2~(\cref{sec:regional-tokenizers}), and RQ3~(\cref{sec:evaluate-generation})]. The resource contains \texttt{1,538}
orthographic pairs and \texttt{275} vocabulary pairs.

Variant pairs were manually compiled from linguistic references and
web-based lexicons (Table~\ref{tab:lexicon-sources}). We merge entries
across sources, normalize their representation, and remove duplicates before
applying a strict matching criterion.

For analyses requiring direct word-level comparison, we retain one-to-one
mappings in which each AmE form corresponds to one Ref(BrE) form. This
criterion gives the resource a common unit of analysis across corpus
frequency counts, tokenizer segmentation, and paired prediction-cost
experiments. Multi-word and structurally different constructions remain
important for regional variation, but are handled separately by
\textsc{DiAlign}, whose local $n$-gram analysis is not restricted to the
one-to-one lexicon.

\input{tables/AmE-BrE-variations}

\subsection{Variant Grouping and Typology}
\label{sec:appendix-variant-grouping}

To characterize the inventory, we assign every pair to one of ten mutually
exclusive groups using a deterministic precedence order. These groups are
then summarized under orthographic, vocabulary, and residual categories.

\paragraph{Group Definitions.}
\begin{itemize}[leftmargin=*,noitemsep]

    \item \textbf{Group 1 (\texttt{-or} vs.\ \texttt{-our}):}
    suffix \texttt{-or} (AmE) \(\leftrightarrow\)
    suffix \texttt{-our} (Ref(BrE)).

    \item \textbf{Group 2 (\texttt{-ize} vs.\ \texttt{-ise}):}
    suffix \texttt{-ize} (AmE) \(\leftrightarrow\)
    suffix \texttt{-ise} (Ref(BrE)).

    \item \textbf{Group 3 (\texttt{-er} vs.\ \texttt{-re}):}
    suffix \texttt{-er} (AmE) \(\leftrightarrow\)
    suffix \texttt{-re} (Ref(BrE)).

    \item \textbf{Group 4 (\texttt{-og} vs.\ \texttt{-ogue}):}
    suffix \texttt{-og} (AmE) \(\leftrightarrow\)
    suffix \texttt{-ogue} (Ref(BrE)).

    \item \textbf{Group 5 (single \texttt{l} vs.\ double \texttt{ll}):}
    AmE single \texttt{l} \(\leftrightarrow\)
    Ref(BrE) double \texttt{ll}.

    \item \textbf{Group 6 (\texttt{-ense} vs.\ \texttt{-ence}):}
    suffix \texttt{-ense} (AmE) \(\leftrightarrow\)
    suffix \texttt{-ence} (Ref(BrE)).

    \item \textbf{Group 7 (\texttt{e} vs.\ \texttt{ae}):}
    the Ref(BrE) form contains \texttt{ae} where the AmE form uses
    \texttt{e}.

    \item \textbf{Group 8 (small orthographic edit):}
    equal-length forms with Levenshtein distance 1--2.

    \item \textbf{Group 9 (different lexical items):}
    forms with substantially different surface realization, identified by
    unequal length or edit distance greater than 2.

    \item \textbf{Group 10 (miscellaneous):}
    remaining pairs not captured by Groups 1--9.

\end{itemize}

\paragraph{Category Assignment.}
Groups 1--8 form the \emph{Orthographic/Spelling} category, Group 9 forms
the \emph{Vocabulary} category, and Group 10 contains residual cases.
This mutually exclusive grouping supports both aggregate comparisons and
finer-grained analyses of recurring orthographic patterns.

\input{tables/AmE-BrE-overview}

\subsection{Sources and Extended Examples}
\label{sec:variant-sources}

Table~\ref{tab:bre_ame_overview} summarizes representative
AmE--Ref(BrE) contrasts across orthography, vocabulary, grammar, and
writing conventions. Tables~\ref{tab:br-am-core} and
\ref{tab:br-am-phrases} provide a more detailed set of examples, while
Table~\ref{tab:lexicon-sources} lists the linguistic references and
web-based resources used to construct the curated variant resource.


\section{\textsc{DiAlign}: Implementation, Validation, and Generalization}
\label{sec:dialign-appendix}

\subsection{Implementation Details}
\label{sec:dialign-implementation-details}

This section records the implementation choices used for the
\textsc{DiAlign} experiments in the paper.

\paragraph{Parameterization.}
We use the following configuration:

\begin{itemize}[leftmargin=*,noitemsep]

    \item \textbf{$n$-gram range:}
    \(n\in\{2,3,4,5\}\). This range captures multi-word and structural
    evidence while remaining within the maximum order supported by
    Google Books Ngram.

    \item \textbf{Temporal range:}
    \([y_{\min},y_{\max}]=[1950,2022]\).
    The broader window provides stable frequency estimates across
    low-frequency constructions. A contemporary-window analysis is reported
    in \cref{sec:dialign-temporal}.

    \item \textbf{Smoothing:}
    \(0\). We use the normalized yearly frequencies returned by the
    reference corpus without additional smoothing.

    \item \textbf{Lexicon boost:}
    \(\beta=1.5\), applied when a retained span contains an entry from the
    curated regional variant lexicon.

\end{itemize}

\paragraph{Frequency Lookup.}
For every retained span, we query the American English corpus
(ID~17) and British English corpus (ID~6) in Google Books Ngram using the
\texttt{requests} library
~\citep{lin-etal-2012-syntactic}. The service returns normalized yearly
frequencies relative to the size of each corpus for that year. We average
these normalized values over the selected temporal range.

Queries are case-insensitive and use exponential-backoff retries for
transient failures. A persistent cache stores every queried
\((n\text{-gram},\text{corpus})\) pair, so repeated spans reuse previously
retrieved frequencies. This substantially reduces reference-side queries
when scoring large collections.

\paragraph{Filtering.}
We remove spans that are likely to reflect topic or named entities rather
than regional usage. A candidate is discarded if it contains a person,
organization, or location identified through NLTK named-entity chunking, or
if every token in the span is a stopword.\footnote{\url{https://www.nltk.org}}

The resulting evidence consists of local constructions whose relative
frequencies differ across the two reference distributions. These local
signals are then combined with the divergence weighting and lexicon boost
defined in \cref{sec:alignment-score}.

\subsection{Illustrative Walkthrough}
\label{sec:dialign-walkthrough}

\begin{figure}[t]
    \centering
    \includegraphics[width=\linewidth]{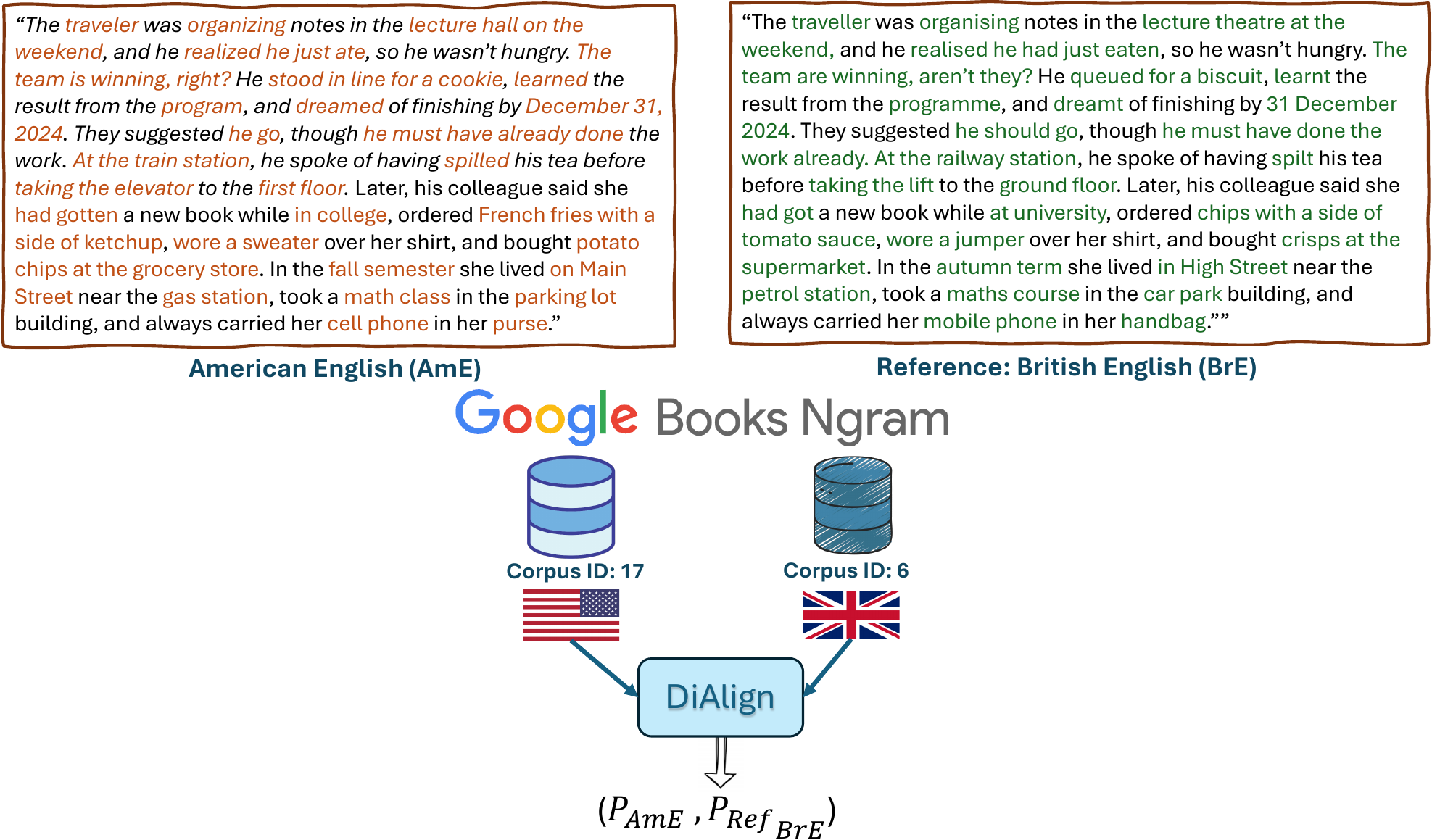}
    \caption{\small Illustrative walkthrough of \textsc{DiAlign}. Parallel AmE and Ref(BrE) passages contain matched contrasts in spelling, vocabulary, grammar, conventions, and multi-word usage. \textsc{DiAlign} decomposes each passage into overlapping 2--5-gram candidate spans, filters non-diagnostic spans, and compares the retained constructions against the American English (ID~17) and British English (ID~6) Google Books reference corpora. The resulting local evidence is aggregated into the regional alignment scores $(P_{\AmE},P_{\RefBrE})$.
    }
    \label{fig:dialign-example}
\end{figure}

Figure~\ref{fig:dialign-example} illustrates \textsc{DiAlign} using parallel
AmE and Ref(BrE) versions of the same passage. The example intentionally
contains several types of regional variation so that the span-level behavior
of the method is visible.

The passage includes examples such as:

\begin{itemize}[leftmargin=*,noitemsep]

    \item \textbf{Spelling:}
    \emph{traveler/traveller},
    \emph{organizing/organising},
    \emph{realized/realised},
    \emph{program/programme}.

    \item \textbf{Vocabulary:}
    \emph{cookie/biscuit},
    \emph{elevator/lift},
    \emph{train station/railway station}.

    \item \textbf{Verb morphology:}
    \emph{dreamed/dreamt},
    \emph{learned/learnt},
    \emph{gotten/got}.

    \item \textbf{Tense and aspect:}
    \emph{I just ate / I've just eaten}.

    \item \textbf{Collective noun agreement:}
    \emph{The team is winning / The team are winning}.

    \item \textbf{Prepositional usage:}
    \emph{on the weekend / at the weekend}.

    \item \textbf{Date conventions:}
    \emph{December 31, 2024 / 31 December 2024}.

    \item \textbf{Institutional and everyday usage:}
    \emph{in college / at university},
    \emph{cell phone / mobile phone},
    \emph{parking lot / car park}.

\end{itemize}

Because the passage is decomposed into overlapping 2--5-grams, these
signals contribute locally rather than through whole-document similarity.
The example also illustrates why \textsc{DiAlign} can capture grammatical
and collocational differences that are absent from the one-to-one variant
inventory.

\subsection{Meta-evaluation Benchmark}
\label{sec:dialign-meta-eval-details}

We evaluate \textsc{DiAlign} on a balanced set of
\texttt{1,500} short news passages with strong source provenance:
750 AmE samples and 750 Ref(BrE) samples.

\paragraph{Ref(BrE) Samples.}
We use English summaries from XL-Sum
~\citep{hasan-etal-2021-xl}, which is sourced from BBC News.\footnote{
\url{https://huggingface.co/datasets/csebuetnlp/xlsum}}
We select entries whose identifiers begin with
\texttt{uk-england-} and randomly sample 750 summaries.

\paragraph{AmE Samples.}
We use the News Category Dataset
~\citep{misra2022newscategorydataset}, built from HuffPost articles.
From the \texttt{U.S. NEWS} category, we randomly sample 750
\texttt{short\_description} fields.\footnote{
\url{https://huggingface.co/datasets/heegyu/news-category-dataset}}

\paragraph{Benchmark Statistics.}
The benchmark is therefore exactly balanced by reference variety.
AmE passages contain 34.4 words on average and Ref(BrE) passages contain
24.7 words on average.

The labels are based on editorial provenance rather than separate
sentence-level human annotation. BBC and HuffPost also differ in editorial
style and topic distribution, so we treat the benchmark as a
source-provenance validation set. Importantly, \textsc{DiAlign} does not
score passages as whole documents: it decomposes each text into overlapping
local 2--5-gram spans, removes named-entity and stopword-only spans, and
aggregates the retained contrastive evidence. This design limits reliance on document-level topic, source identity, and global stylistic differences,
while the benchmark tests whether local linguistic patterns track the
expected regional convention.

\subsection{Validation and Ablation}
\label{sec:dialign-validation}

On the balanced benchmark, \textsc{DiAlign} reaches
\textbf{93.18\% accuracy}, \textbf{90.67\% precision},
\textbf{96.25\% recall}, and \textbf{93.38 F1}
(Figure~\ref{fig:DiAlign-performance}). Correct predictions are concentrated
at larger confidence margins, while errors occur primarily near the
decision boundary.

\input{tables/dialign-ablation}

The ablation results in Table~\ref{tab:dialign-ablation} show that both
components contribute to the final estimator. Removing either divergence
weighting or lexicon boosting reduces accuracy from 93.18\% to 92.25\%;
removing both lowers it to 91.31\%. The two components therefore provide
complementary evidence rather than relying on a single lexical heuristic.

\subsection{Temporal Robustness}
\label{sec:dialign-temporal}

Google Books contains text spanning many decades, so we test whether the
alignment signal depends strongly on older usage. We rerun the complete
meta-evaluation using only frequencies from \texttt{2000--2022}, while
keeping all other scoring choices fixed.

\begin{table}[t]
\centering
\small
\setlength{\tabcolsep}{8pt}
\renewcommand{\arraystretch}{1.08}
\caption{\small Temporal robustness of \textsc{DiAlign} to the Google Books reference window. Restricting the reference distributions from 1950--2022 to 2000--2022 changes accuracy by only 0.27 points and F1 by 0.26 points, indicating that performance is stable to the temporal range used for frequency estimation.
}
\begin{tabular}{lrrrr}
\toprule
\textbf{Reference period}
& \textbf{Accuracy}
& \textbf{Precision}
& \textbf{Recall}
& \textbf{F1} \\
\midrule
1950--2022 & 93.18 & 90.67 & 96.25 & 93.38 \\
2000--2022 & 92.91 & 90.42 & 95.98 & 93.12 \\
\bottomrule
\end{tabular}
\label{tab:dialign-temporal}
\end{table}

The contemporary window produces nearly identical results:
accuracy changes from 93.18\% to 92.91\%, and F1 from 93.38 to 93.12.
The regional signal used by \textsc{DiAlign} is therefore stable across
these two reference periods.

\subsection{Generalizing \textsc{DiAlign}}
\label{sec:dialign-generalization}

\textsc{DiAlign} requires a pair of comparable reference distributions:
one for the variety being studied and one for the reference variety. Google
Books provides this pair directly for the AmE--Ref(BrE) analysis in this
paper, but the scoring procedure itself is corpus-independent.

For example, an AmE--Indian English study could construct an Indian-English
reference corpus from regional news organizations, institutional websites,
and web sources with reliable provenance. Country-code top-level domains
(ccTLDs), such as \texttt{.in} for India, can contribute one provenance
signal when combined with publisher, institutional, and source-location
metadata. Comparable U.S. sources could provide the AmE distribution.
Both collections could then be converted into normalized 2--5-gram
frequency tables and used directly by \textsc{DiAlign}.

The same strategy applies to other regional varieties. Potential provenance
signals include \texttt{.ng} for Nigeria, \texttt{.au} for Australia,
\texttt{.nz} for New Zealand, and \texttt{.sg} for Singapore. Combining
multiple signals is preferable to treating a domain suffix as a linguistic
label on its own.

For any two varieties \(A\) and \(B\), the core contrast becomes

\[
\mathrm{LR}_{A,B}(g)
=
\log_2
\left(
\frac{f_A(g)}{f_B(g)}
\right).
\]

Positive values provide evidence toward \(A\), negative values toward
\(B\), and the remaining weighting and aggregation procedure is unchanged.

The framework can also be instantiated beyond English when suitable
variety-specific corpora can be constructed. The central requirement is a
pair of sufficiently large and comparable reference distributions with
reliable provenance. Google Books is therefore the reference resource used
for the present study, rather than a requirement of the method.


\section{Additional Details for RQ1: Data Exposure}
\label{app:rq1-details}

\subsection{Pretraining Datasets}
\label{sec:pretraining-datasets}

We audit six widely used pretraining corpora representing books,
encyclopedic text, web data, and mixed-domain collections.

\begin{itemize}[leftmargin=*]

\item \textbf{BookCorpus}~\citep{10.1109/ICCV.2015.11}:
a collection of unpublished novels frequently used in language-model
pretraining. The analyzed release contains approximately 74 million
documents and 1.28 billion tokens.\footnote{
\url{https://huggingface.co/datasets/bookcorpus/bookcorpus}}

\item \textbf{Wikipedia}~\citep{wikidump}:
encyclopedic text from Wikipedia dumps. The analyzed release contains
approximately 6.4 million documents and 4.3 billion tokens.\footnote{
\url{https://huggingface.co/datasets/wikimedia/wikipedia}}

\item \textbf{Common Crawl (C4)}~\citep{10.5555/3455716.3455856}:
a cleaned and deduplicated Common Crawl collection containing approximately
365 million documents and 156 billion tokens.\footnote{
\url{https://huggingface.co/datasets/allenai/c4}}

\item \textbf{Falcon RefinedWeb}~\citep{penedo2023the}:
a Common Crawl-derived web corpus with additional filtering and
deduplication, comprising approximately 968 million documents and
600 billion tokens.\footnote{
\url{https://huggingface.co/datasets/tiiuae/falcon-refinedweb}}

\item \textbf{RedPajama}~\citep{weber2024redpajama}:
a reproduction of major LLaMA-style data sources spanning books, code,
papers, and web content. We use the sampled 1T release analyzed in this
study.\footnote{
\url{https://huggingface.co/datasets/togethercomputer/RedPajama-Data-1T-Sample}}

\item \textbf{Dolma}~\citep{soldaini-etal-2024-dolma}:
an open mixed-domain collection combining web pages, books, code, papers,
forums, and social-media data. We use the v1.6 sample release.\footnote{
\url{https://huggingface.co/datasets/allenai/dolma}}

\end{itemize}

\subsection{Pretraining Preprocessing}
\label{sec:preprocessing-pipeline}

All six corpora are processed through the same normalization pipeline before
variant counting. Text is lowercased, and HTML tags, hyperlinks, and email
addresses are removed. Hyphens and slashes are replaced with whitespace so
that joined forms do not obscure word boundaries. Non-alphabetic characters
are then replaced with whitespace and repeated whitespace is collapsed.

This normalization yields a consistent representation for matching the
\texttt{1,813} AmE--Ref(BrE) pairs across corpora. The same preprocessing is
applied to both members of every pair, so corpus comparisons use an
identical matching procedure.

\subsection{Corpus Processing and Sampling}
\label{sec:pretraining-scale}

We process the large pretraining corpora using Hugging Face streaming,
which allows examples to be analyzed without locally materializing the full
dataset.\footnote{\url{https://huggingface.co/docs/datasets/en/stream}}

Variant-frequency counts are computed over the streamed corpus material.
For the broader \textsc{DiAlign} analysis, reference lookup is the dominant
computational cost because every distinct local span requires corresponding
AmE and Ref(BrE) reference frequencies. We therefore apply
\textsc{DiAlign} to a uniform sample of \texttt{10,000} documents from each
pretraining corpus. The same sampling size and scoring procedure are used
across all six resources.

\subsection{Additional Pretraining Results}
\label{sec:rq1-additional-results}

\begin{figure*}[t]
    \centering
    \includegraphics[width=\linewidth]{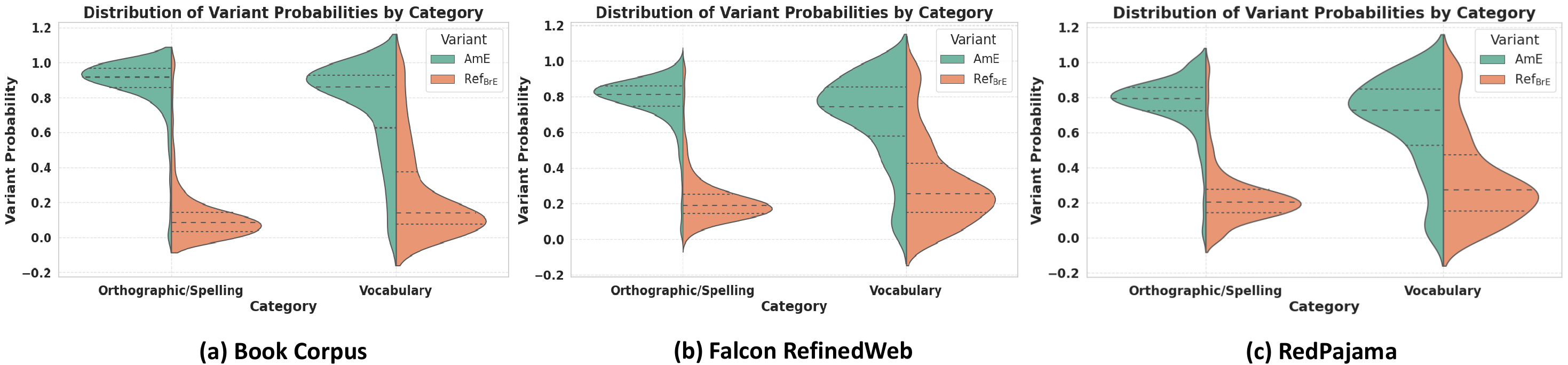}
    \vspace{-3mm}
    \caption{\small Variant-probability distributions for \textbf{(a)} BookCorpus, \textbf{(b)} Falcon RefinedWeb, and \textbf{(c)} RedPajama. Orthographic contrasts are more strongly concentrated toward AmE, while vocabulary contrasts are more dispersed but still lean toward AmE.
    }
    \label{fig:violin-others}
\end{figure*}

\begin{figure*}[t]
    \centering
    \includegraphics[width=\linewidth]{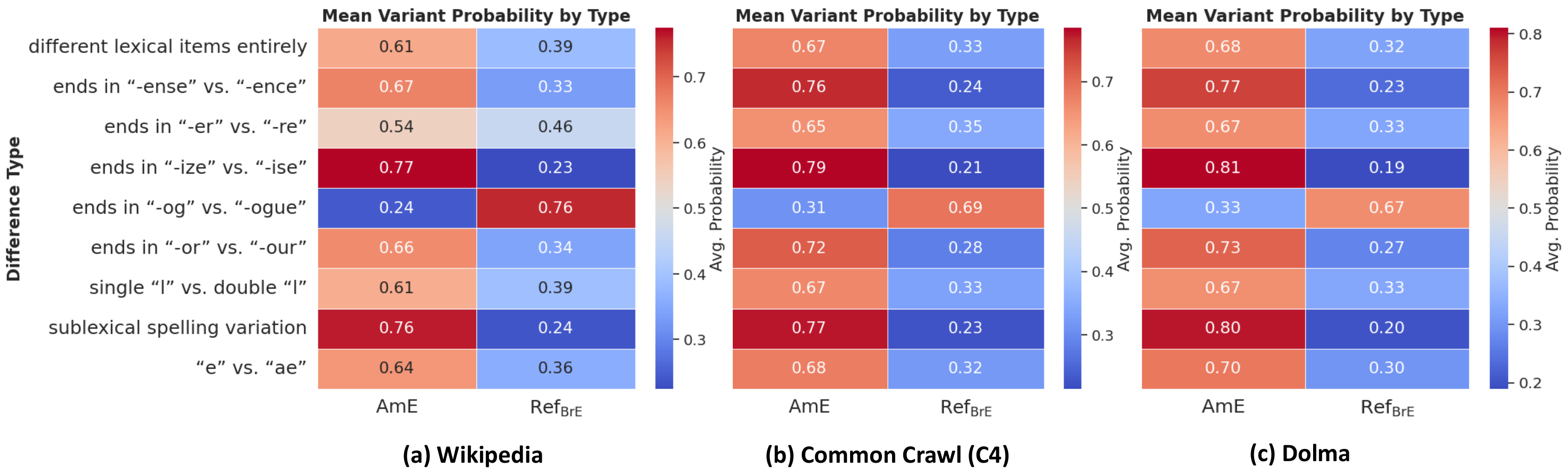}
    \vspace{-3mm}
    \caption{\small Category-level AmE and Ref(BrE) probabilities for \textbf{(a)} Wikipedia, \textbf{(b)} Common Crawl (C4), and \textbf{(c)} Dolma. Cells show the mean pair-level probability within each linguistic contrast category.
    }
    \label{fig:heatmap-main}
\end{figure*}

\begin{figure*}[t]
    \centering
    \includegraphics[width=\linewidth]{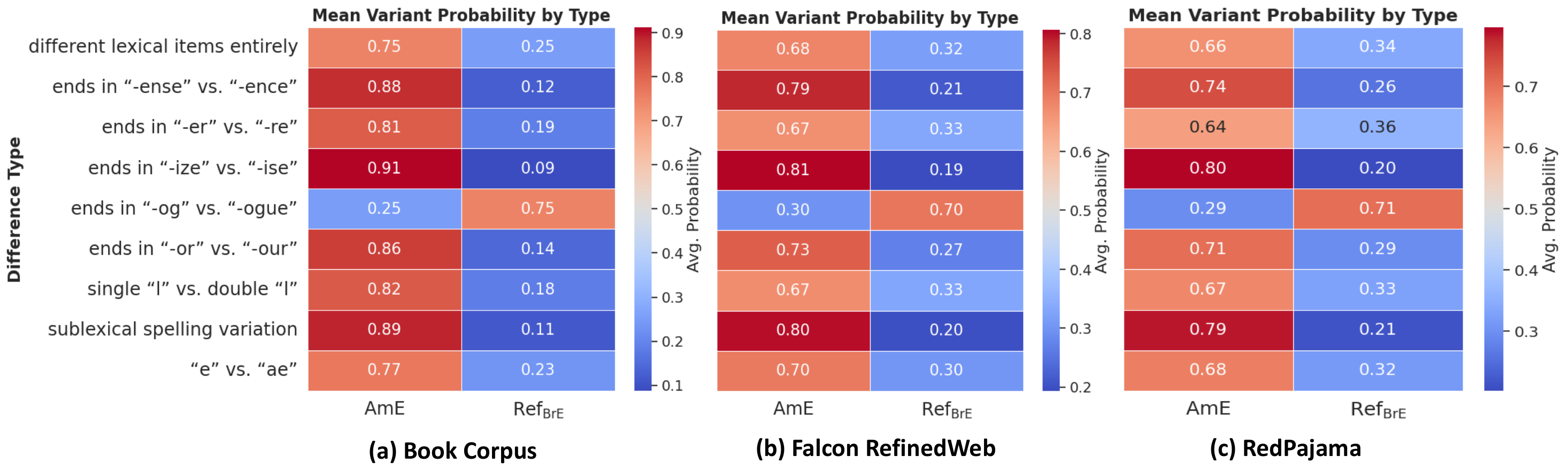}
    \vspace{-5mm}
    \caption{\small Category-level AmE and Ref(BrE) probabilities for \textbf{(a)} BookCorpus, \textbf{(b)} Falcon RefinedWeb, and \textbf{(c)} RedPajama. The same category-level pattern observed in Figure~\ref{fig:heatmap-main} is also evident across these corpora.
    }
    \label{fig:heatmap-app}
\end{figure*}

Figures~\ref{fig:violin-others}--\ref{fig:heatmap-app} provide the
distributional views omitted from the main paper.

\begin{figure}[t]
    \centering
    \includegraphics[width=0.98\linewidth]{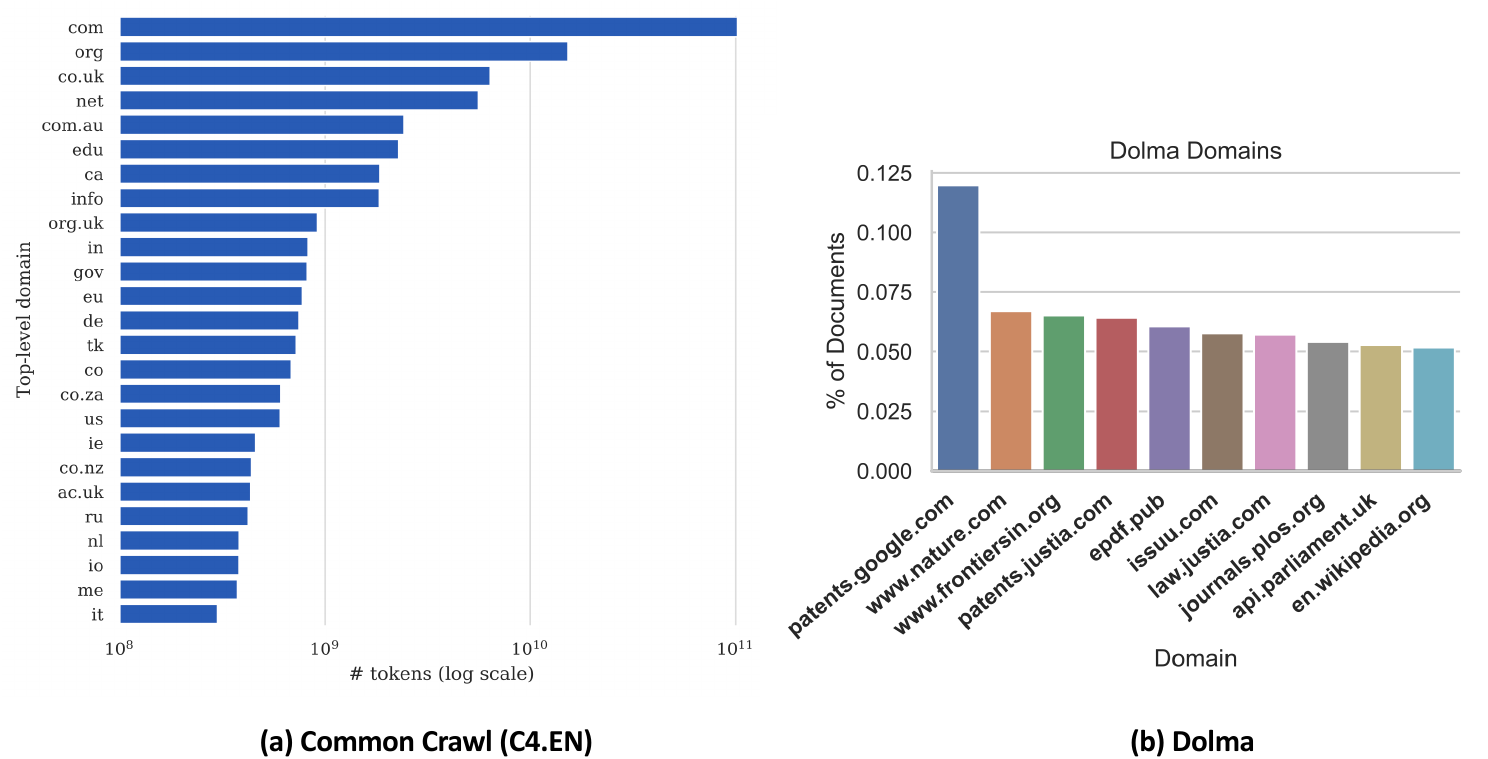}
    \caption{\small Source distributions in two pretraining corpora. \textbf{(a)} Top-level domains in C4.EN are concentrated in \texttt{.com} and \texttt{.org}, with much smaller shares from country-specific domains such as \texttt{.co.uk}. \textbf{(b)} High-frequency sources in Dolma illustrate the concentrated provenance of a curated mixed-domain corpus~\citep{10.5555/3455716.3455856,soldaini-etal-2024-dolma}.
    }
    \label{fig:domain-distribution}
\end{figure}

\subsection{Post-Training Dataset Audit}
\label{app:posttraining_datasets}

Table~\ref{tab:posttraining_dataset_full} reports the dataset-level results
underlying Table~\ref{tab:posttraining_stage} in the main paper. The audit
covers 21 public resources spanning supervised fine-tuning, chat data, human
feedback, preference learning, reward modeling, and safety alignment.

Where multiple closely related releases exist, we use one cleaned, current,
or original public release to avoid duplicate counting across dataset
families.

\input{tables/posttraining_dataset_full}

For each dataset, \textbf{Weighted AmE (\%)} reports the aggregate AmE share
across all observed variant occurrences. \textbf{Orth. AmE (\%)} and
\textbf{Vocab. AmE (\%)} instead report mean pair-level probabilities within
their respective linguistic categories, so these columns are not complements
of one another.

For OpenAI Summarize Comparisons, we remove the context-sensitive
\emph{mail/post} pair because \emph{post} frequently denotes an online post
rather than postal mail. After excluding this pair, the dataset contains
176,589 retained variant occurrences and has a weighted AmE share of
73.18\%.

Across all 21 datasets, weighted AmE prevalence ranges from 65.29\% in
Databricks Dolly 15k to 89.72\% in WizardLM Evol-Instruct v2.


\section{Additional Details for RQ2: Representation \& Prediction Cost}
\label{app:rq2-supplement}

\subsection{Additional Tokenization Granularity}
\label{app:tokenizer-granularity}

\begin{figure*}[t]
    \centering
    \includegraphics[width=\linewidth]
    {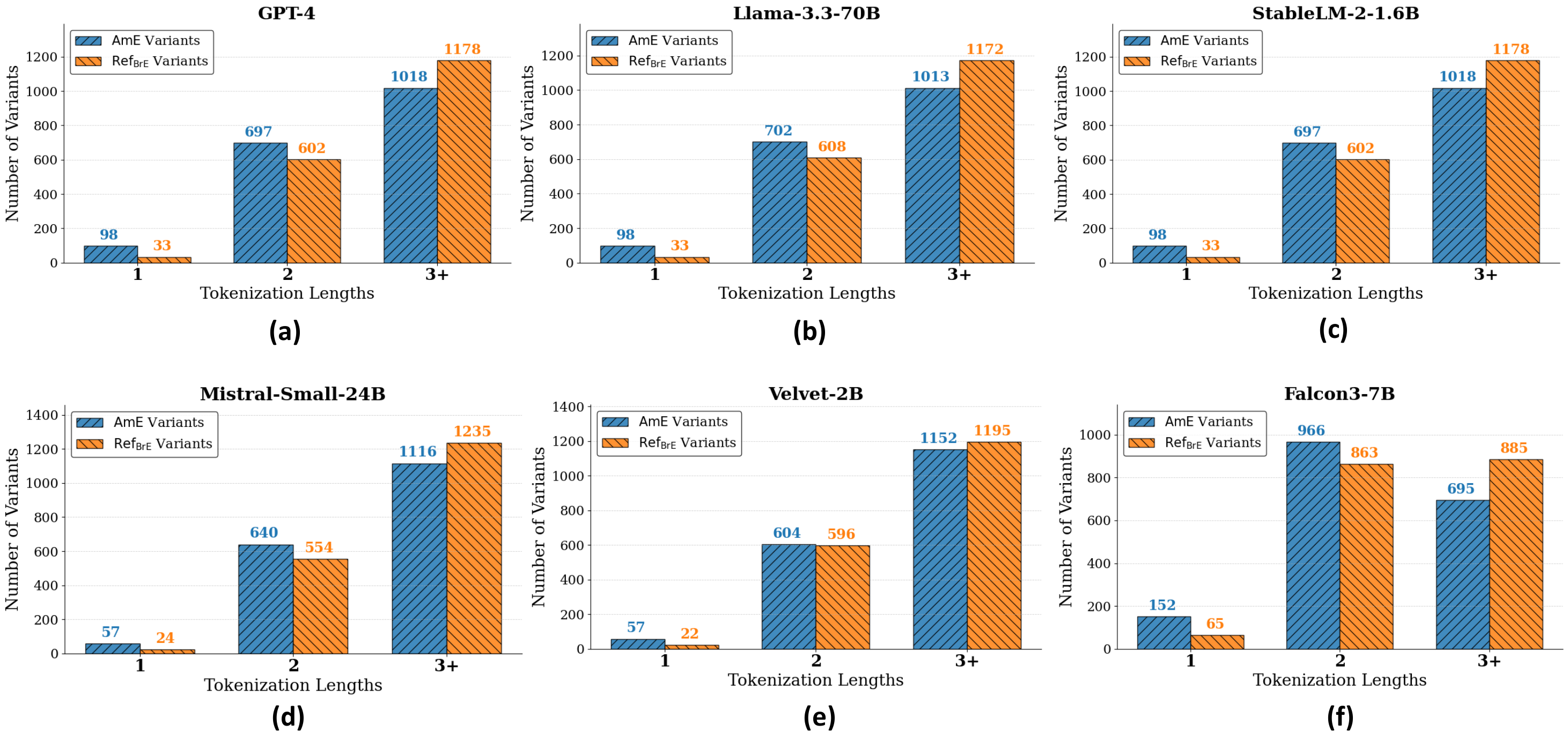}
    \vspace{-3mm}
    \caption{\small Additional tokenization-granularity results for \textbf{(a)} GPT-4, \textbf{(b)} Llama-3.3-70B, \textbf{(c)} StableLM-2-1.6B, \textbf{(d)} Mistral-Small-24B, \textbf{(e)} Velvet-2B, and \textbf{(f)} Falcon3-7B. Bars show the number of AmE and Ref(BrE) variants represented by 1, 2, or 3+ subwords.
    }
    \label{fig:tokenizer-granularity-app}
\end{figure*}

Figure~\ref{fig:tokenizer-granularity-app} extends the main granularity
analysis to the remaining tokenizer families. The same long-tail pattern is
visible across the panel: Ref(BrE) variants are more frequently split into
three or more subword units.

\subsection{Tokenizer Provenance and Lineage}
\label{app:tokenizer_provenance}

The tokenizer results are best interpreted with vocabulary provenance
separated from model developer origin. We therefore audit three related but
distinct properties: model origin, tokenizer vocabulary lineage, and
observed segmentation behavior.

Seven of the nine evaluated tokenizers use independent or source
vocabularies. Llama-3.3 and StableLM-2 instead inherit or extend
US-origin tokenizer artifacts. These tokenizers remain part of the
behavioral analysis because their segmentation behavior is directly
relevant to users and models, while their lineage provides important context
for interpreting geographic metadata.

\subsection{Lineage Diagnostics}
\label{app:tokenizer_lineage_metrics}

Vocabulary overlap alone cannot distinguish direct inheritance from ordinary
surface similarity. We therefore use three complementary diagnostics.

For tokenizer vocabularies \(V_i\) and \(V_j\), directional vocabulary
coverage is

\begin{equation}
S_{i\rightarrow j}
=
100 \times
\frac{|V_i\cap V_j|}{|V_i|}.
\label{eq:tokenizer_vocab_coverage}
\end{equation}

This quantity captures asymmetric extension, where one tokenizer may contain
nearly the complete vocabulary of another while adding new tokens.

For shared token surfaces, exact rank/id agreement is

\begin{equation}
R_{ij}
=
100 \times
\frac{
\left|
\left\{
v\in V_i\cap V_j :
\mathrm{rank}_i(v)=\mathrm{rank}_j(v)
\right\}
\right|
}{
|V_i\cap V_j|
}.
\label{eq:tokenizer_rank_agreement}
\end{equation}

Finally, for inputs \(x_1,\ldots,x_N\), token-boundary identity is

\begin{equation}
B_{ij}
=
100\times
\frac{1}{N}
\sum_{k=1}^{N}
\mathbf{1}
\left\{
\mathrm{boundaries}_i(x_k)
=
\mathrm{boundaries}_j(x_k)
\right\}.
\label{eq:tokenizer_boundary_identity}
\end{equation}

We compute boundary identity on the \texttt{1,813} AmE and
\texttt{1,813} Ref(BrE) forms, as well as validation and generated-text
inputs. The combination of vocabulary coverage, rank agreement, and actual
segmentation behavior distinguishes artifact inheritance from superficial
vocabulary overlap.

\input{tables/tokenizer_lineage}

\subsection{Behavioral Evidence for Tokenizer Reuse}
\label{app:tokenizer_reuse}

StableLM-2 provides the clearest inheritance case. Its tokenizer exhibits
\textbf{100.00\% token-boundary identity} with GPT-4 on both the AmE and
Ref(BrE) variant sets. In addition, all \texttt{100,256} shared tokens in
the GPT-4-sized base vocabulary retain identical rank/id assignments.

Llama-3.3 shows similarly strong behavioral continuity, with 99.17\%
boundary identity on AmE variants and 99.06\% on Ref(BrE) variants relative
to GPT-4.

The independently sourced DeepSeek, Mistral, Falcon, and Velvet tokenizers
show substantially lower boundary identity. GPT-4o is also informative:
despite considerable token-surface overlap with GPT-4, its rank structure
and segmentation behavior differ markedly. Surface vocabulary overlap alone
is therefore insufficient to establish lineage.

\input{tables/tokenizer_boundary_similarity}

These results show how tokenizer conventions can persist across model
families through artifact reuse. Model developer origin and tokenizer
provenance therefore describe different parts of the model-development
stack, and we report them separately throughout the paper.

\subsection{Counterfactual Sentence Construction}
\label{app:rq2-counterfactuals}

The prediction-cost analysis uses the same \texttt{1,813} matched regional
variant pairs, comprising \texttt{1,538} orthographic pairs and
\texttt{275} vocabulary pairs. Instantiating these variants in controlled
contexts produces \texttt{8,533} matched sentence-pair rows. Within each
pair, only the target AmE or Ref(BrE) span changes.

Most pairs use conservative minimal-mention templates:

\begin{quote}
\small
\texttt{The word is \{variant\}.}\\
\texttt{This sentence uses the word \{variant\}.}\\
\texttt{The document contains the term \{variant\}.}
\end{quote}

Orthographic pairs additionally use spelling-specific contexts:

\begin{quote}
\small
\texttt{The article used the spelling \{variant\}.}\\
\texttt{The preferred form in this document is \{variant\}.}
\end{quote}

These templates isolate the regional surface form while holding the
surrounding sentence context fixed. The broader controlled RQ2 resource uses
core-quality spelling pairs instantiated with either minimal-mention or
orthographic-context templates. This broader slice is also used for the
semantic-equivalence analysis in \cref{app:rq2-semantic}.

The strict confirmatory prediction-cost results reported in
Table~\ref{tab:rq2-loss-gap-appendix} use the more conservative
minimal-mention subset.

We additionally instantiate 18 manually reviewed lexical pairs in natural
contexts. For example,

\begin{quote}
\small
\texttt{She rented a small \{variant\} near the station.}
\end{quote}

instantiates \emph{apartment/flat}. These examples provide a separate
natural-context robustness subset.

\input{tables/rq2-loss-gap-appendix}

\subsection{Prediction-Cost Evaluation}
\label{app:rq2-loss}

For each checkpoint, we compute sentence negative log-likelihood and
normalize by the model-specific token count:

\[
\Delta^{\mathrm{loss}}_{m,i}
=
\frac{\mathrm{NLL}_m(x_i^{\RefBrE})}
     {|\tau_m(x_i^{\RefBrE})|}
-
\frac{\mathrm{NLL}_m(x_i^{\AmE})}
     {|\tau_m(x_i^{\AmE})|}.
\]

Positive values indicate higher prediction cost for the Ref(BrE)
counterfactual.

For the strict confirmatory analysis, we restrict to core-quality spelling
pairs instantiated with minimal-mention templates. We report results on this
clean subset and on tokenizer-controlled slices. The strictest sentence-level
control retains only pairs with equal token counts:

\[
|\tau_m(x_i^{\RefBrE})|
=
|\tau_m(x_i^{\AmE})|.
\]

This ensures that the reported per-token loss gap is not attributable to one
counterfactual being represented by more sentence tokens than the other.

We also evaluate a target-level control in which the whitespace-prefixed AmE
and Ref(BrE) forms receive identical token counts. This captures tokenizers
that encode word-in-context forms using leading-space units.

Alongside the mean gap, we report the directional share, defined as the
percentage of pairs for which
\(\Delta^{\mathrm{loss}}_{m,i}>0\).

Every evaluated checkpoint assigns higher per-token loss to the Ref(BrE)
counterfactual on the strict equal-sentence-token-count slice, and every
paired bootstrap 95\% confidence interval excludes zero. The full evaluation
contains \texttt{85,150} paired-loss records across ten checkpoints, with no
skipped models.

\subsection{Model Scope}
\label{app:rq2-model-scope}

The prediction-cost panel contains five model families for which matched
public base and post-trained checkpoints are available at comparable scale.

\input{tables/rq2-checkpoints}

DeepSeek-V3 remains part of the tokenizer and provenance analyses, but a
comparable small DeepSeek-V3 base/post-trained pair is not publicly
available. Velvet-2B is likewise retained in the tokenizer analysis, while
its model family does not provide a separately released public base
checkpoint for the matched comparison.

\subsection{Semantic-Equivalence Control}
\label{app:rq2-semantic}

The semantic-equivalence analysis uses the broader controlled spelling
slice described in \cref{app:rq2-counterfactuals}, including both
minimal-mention and orthographic-context templates. Its purpose is to verify
that changing the regional form preserves the intended sentence meaning.

Let \(h_{m,\ell}(x)\) denote the mean-pooled representation of sentence
\(x\) at layer \(\ell\). For each matched pair, we compute

\[
\mathrm{sim}_{m,\ell,i}^{\AmE,\RefBrE}
=
\cos\!\left(
h_{m,\ell}(x_i^{\AmE}),
h_{m,\ell}(x_i^{\RefBrE})
\right).
\]

Matched regional counterfactuals are compared against two controls:
a character-perturbed AmE sentence and an unrelated Ref(BrE) sentence.

\input{tables/rq2-layerwise-semantic-equivalence}

At the final layer, matched counterfactuals reach a mean cosine similarity of
0.989, compared with 0.928 for character perturbations and 0.814 for
unrelated controls. The matched pairs remain closer at embedding, middle,
and final representation levels.

The semantic evaluation contains \texttt{108,000} similarity records across
the ten checkpoints, with no skipped models.

\subsection{Base and Post-Trained Checkpoint Comparison}
\label{app:rq2-base-post}

Table~\ref{tab:rq2-loss-gap-appendix} provides matched comparisons between
base and post-trained checkpoints for all five model families on the strict
minimal-mention, equal-sentence-token-count slice. In every family, the
Ref(BrE) prediction-cost gap is larger in the post-trained checkpoint.

This is a consistent checkpoint-level pattern across Llama, StableLM,
Ministral, Gemma, and Falcon3. Because each pair differs in its
post-training procedure and data, the comparison characterizes the observed
checkpoints rather than isolating a single post-training intervention.


\section{Additional Details for RQ3: Generation}
\label{app:rq3-setup}

\subsection{Datasets and Sampling}
\label{app:rq3-data}

RQ3 evaluates whether generated responses align more strongly with AmE or
Ref(BrE) under neutral English prompting and explicit British-English
steering.

We use two open-domain QA datasets with different communicative registers:

\begin{itemize}[leftmargin=*]

  \item \textbf{Natural Questions (NQ)}
  ~\citep{kwiatkowski-etal-2019-natural}:
  real search questions paired with Wikipedia-derived answers, providing a
  comparatively formal and encyclopedic setting.\footnote{
  \url{https://huggingface.co/datasets/sentence-transformers/natural-questions}}

  \item \textbf{ELI5}
  ~\citep{fan-etal-2019-eli5}:
  community question answering from Reddit, providing a more conversational
  and explanatory register.\footnote{
  \url{https://huggingface.co/datasets/sentence-transformers/eli5}}

\end{itemize}

\begin{wrapfigure}{r}{0.58\columnwidth}
    \centering
    \vspace{-8.5mm}
    \includegraphics[
        width=\linewidth,
        keepaspectratio
    ]{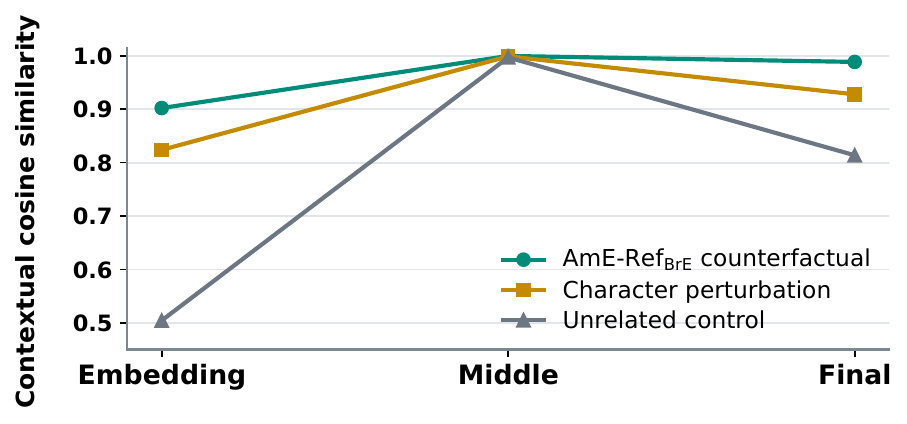}
    \vspace{-3mm}
    \caption{\small
    \textbf{Layerwise semantic-equivalence control.}
    Matched \AmE{}--\RefBrE{} counterfactuals remain closer than
    character-perturbed or unrelated controls across embedding,
    middle, and final representations, with the clearest separation
    at the final layer.
    }
    \label{fig:rq2-layerwise-semantic-equivalence}
    \vspace{-13mm}
\end{wrapfigure}

We discard questions shorter than five words and examples whose reference
answer contains fewer than 30 words. Questions containing any of the
\texttt{3,626} individual forms in the AmE--Ref(BrE) inventory are also
removed so the prompt itself does not supply an explicit regional lexical
cue.

From the remaining pool, we uniformly sample 300 NQ questions and 300 ELI5
questions.

\subsection{Prompting and Decoding}
\label{app:rq3-prompting}

Each question is presented under two language conditions:
\texttt{English} and \texttt{British English (en-GB)}. The requested output
length is 50 words, providing a common target across models and conditions.

\begin{figure}[t]
\centering
\begin{tcolorbox}[
  enhanced,
  breakable,
  title={Prompt template for RQ3: Regional Preference in Generation},
  colback=gray!2,
  colframe=black!75,
  fonttitle=\bfseries,
  boxrule=0.6pt,
  arc=2pt,
  left=2mm,
  right=2mm,
  top=1mm,
  bottom=1mm,
  width=0.95\linewidth
]
\ttfamily
Answer the following question in \{language\}. Write a single, coherent
paragraph in plain text, using descriptive and open-ended language.
Avoid bullet points, lists, or formatting. Your response must be exactly
\{WORD\_LIMIT\} words long---no more, no fewer. Count your words carefully.\\

Question: \{question\}
\end{tcolorbox}
\caption{\small
Prompt used under the two language conditions.
We set \texttt{WORD\_LIMIT}$=50$ and
\texttt{\{language\}}~$\in$
\{\texttt{English}, \texttt{British English (en-GB)}\}.
}
\label{fig:rq3-prompt}
\end{figure}

Decoding parameters are fixed across both conditions:

\[
\texttt{temperature}=0.0,\qquad
\texttt{top\_p}=1.0,\qquad
\texttt{max\_tokens}=512.
\]

Greedy decoding removes sampling variance, while the common decoding
configuration ensures that the two language conditions differ only in the
regional instruction.

\subsection{Scoring and Outcome Measures}
\label{app:rq3-scoring}

Each generated paragraph is scored using the same \textsc{DiAlign}
procedure described in \cref{sec:alignment-score}. For each response, we
record \(P_{\AmE}\), \(P_{\RefBrE}\), and the resulting majority alignment.

A response with no surviving directional evidence is treated as a
zero-signal case and omitted from the aggregate classification rate. Such
cases are rare and correspond to passages for which the retained local spans
provide no measurable contrast between the two reference distributions.

For each model, dataset, and prompt condition, we report:

\begin{enumerate}
    \item the percentage of responses classified as AmE, and
    \item the mean AmE alignment score \(P_{\AmE}\).
\end{enumerate}

\subsection{Word-Length Adherence}
\label{sec:length-control}

\input{tables/length-control-app}

The 50-word instruction is a common target rather than a post-processing
constraint, so adherence varies by model. GPT-4o, Gemini, Claude, and Gemma
remain relatively close to the requested length, while several smaller
open-weight models exhibit wider distributions.

The resulting variation provides an additional opportunity to test whether
response length itself explains the measured AmE preference.

\subsection{Response-Length Robustness}
\label{app:rq3-length-robustness}

\textsc{DiAlign} aggregates overlapping 2--5-gram evidence. For an input of
\(N\geq5\) tokens, the number of candidate spans before filtering is

\[
|\mathcal{G}(x)|=4N-10.
\]

This gives short responses a comparatively dense set of overlapping local
constructions while the fixed-length instruction keeps the intended
generation scale comparable across models.

We additionally examine the outputs by realized word length:

\begin{table}[t]
\centering
\small
\setlength{\tabcolsep}{7pt}
\renewcommand{\arraystretch}{1.08}
\caption{\small
Realized response length and AmE alignment across length bins.
Default-English outputs remain AmE-majority throughout, while
British-English prompting produces the lowest AmE alignment among shorter, more instruction-adherent responses.
}
\begin{tabular}{llrr}
\toprule
\textbf{Condition}
& \textbf{Length bin}
& \textbf{Avg. words}
& \textbf{DiAlign AmE (\%)} \\
\midrule

Default & 1--50   & 44.08  & 67.64 \\
Default & 51--75  & 55.82  & 67.33 \\
Default & 76--100 & 86.70  & 64.31 \\
Default & 101--150 & 116.03 & 62.54 \\
Default & 151+    & 225.60 & 61.01 \\
\midrule
British-English steer & 1--50   & 43.62  & 47.12 \\
British-English steer & 51--75  & 55.99  & 48.44 \\
British-English steer & 76--100 & 86.43  & 58.80 \\
British-English steer & 101--150 & 116.26 & 57.79 \\
British-English steer & 151+    & 192.20 & 60.94 \\
\bottomrule
\end{tabular}

\label{tab:rq3-length-robustness}
\end{table}

The aggregate association between response length and AmE alignment is very
small:
\(r=0.017\) across all outputs,
\(r=-0.043\) under default English prompting, and
\(r=0.057\) under British-English steering.

Response length therefore does not explain the aggregate AmE preference.
The length-bin analysis also shows that explicit British-English prompting
has its strongest effect among shorter responses, while longer steered
generations move somewhat closer to the AmE distribution.


\section{Discussion and Practical Recommendations}
\label{sec:discussion-recommendations}

The AmE advantage appears consistently across data exposure, representation,
and generation. RQ1 shows AmE overrepresentation in both pretraining and
post-training resources, RQ2 identifies corresponding asymmetries in
tokenization and model prediction cost, and RQ3 shows that AmE remains the
dominant generation default under neutral English prompting. These findings
suggest that regional imbalance should be addressed across the LLM
development pipeline rather than only through output-level correction.

\paragraph{Audit and Rebalance Training Data.}
RQ1 (\sref{sec:audit-pretraining-corpus}) shows that regional skew is already
present in both pretraining and post-training data. Dataset construction can
therefore track regional English variation alongside conventional measures
of domain, quality, and language composition. For web-scale resources,
provenance signals such as publisher location, institutional metadata,
source URLs, and country-code top-level domains can support regional
sampling strategies. For example, \texttt{.uk}, \texttt{.in},
\texttt{.ng}, \texttt{.au}, and \texttt{.sg} can provide useful regional
signals when combined with source and publisher metadata rather than being
treated as linguistic labels on their own.

The same audit should extend beyond pretraining. Instruction-tuning,
preference, reward-model, and safety datasets can introduce additional
regional skew after pretraining. Variant-frequency counts provide a
lightweight way to monitor explicit spelling and vocabulary differences,
while \textsc{DiAlign} can complement them with broader grammatical,
structural, collocational, and multi-word evidence~(\sref{sec:alignment-score}). These diagnostics can be applied before
training to compare candidate mixtures and adjust sampling when a strong
unintended imbalance is present.

\paragraph{Include Regional Coverage in Tokenizer Design.}
RQ2 (\sref{sec:regional-tokenizers}) shows that Ref(BrE) forms are generally
segmented less compactly than matched AmE forms. Tokenizer evaluation can
therefore include regional coverage alongside aggregate compression
metrics. Fertility and granularity provide simple diagnostics for
identifying forms that are repeatedly split into longer subword sequences.

When such gaps are large, vocabulary extension or tokenizer retraining can
increase coverage of underrepresented variants
~\citep{tejaswi-etal-2024-exploring}. For example, forms that consistently
fall into the 3+ subword category while their matched AmE counterparts are
encoded more compactly are natural candidates for closer inspection.
Tokenizer provenance should also be considered before reuse, since our
lineage analysis shows that segmentation behavior can persist across model
families through inherited tokenizer artifacts.

\paragraph{Evaluate Prediction Cost Beyond Tokenization.}
More balanced segmentation does not necessarily imply equal model
preference. Our controlled likelihood analysis shows that Ref(BrE)
counterfactuals receive higher prediction cost even when sentence token
count is identical. A practical regional evaluation can therefore report
both tokenization differences and \(\Delta^{\mathrm{loss}}\).

For example, a model may tokenize \emph{color} and \emph{colour} using the
same number of sentence tokens while still assigning the Ref(BrE)
counterfactual higher per-token negative log-likelihood. Evaluating both
quantities helps distinguish tokenizer-level representation from
preferences encoded in the model itself. Repeating the same evaluation for
base and post-trained checkpoints can also reveal whether the regional gap
becomes larger or smaller across model development.

\paragraph{Treat Regional Prompting as Steering.}
RQ3 shows that explicit British-English prompting generally moves outputs
toward Ref(BrE), but the size of the shift varies across models and
registers. Regional prompting is therefore useful as an interface control,
but it should not replace upstream data and representation audits.

Evaluation should also go beyond simple spelling substitution. A response
may use \emph{colour} instead of \emph{color} while still following AmE
preferences in vocabulary, grammar, conventions, or multi-word usage.
Regional settings should therefore be evaluated across several types of
variation rather than judged only by a small set of visible lexical changes.

\paragraph{Use \textsc{DiAlign} for Regional Auditing.}
\textsc{DiAlign} can support these checks during both data construction and
generation evaluation. During corpus building, it can estimate the regional
alignment of candidate training data or synthetic additions. During
evaluation, it can compare neutral-English and region-specific generations
using the same local evidence.

The framework is also adaptable beyond the present AmE--Ref(BrE)
comparison. For example, an AmE--Indian English analysis could construct a
regional reference corpus using Indian news, institutional websites, and
reliable provenance signals such as \texttt{.in}, combined with publisher
and source-location metadata. Similar strategies could support studies of
Nigerian, Australian, New Zealand, or Singapore English
(Appendix~\ref{sec:dialign-generalization}).

These recommendations map directly to the stages examined in this paper:
audit regional exposure in training data, measure asymmetry in tokenization
and model prediction cost, and evaluate the realized generation default.
Study limitations and directions for extending the analysis are discussed
in Appendix~\ref{sec:limitations}.


\section{Extended Related Work}
\label{sec:extended-related-work}

\paragraph{Training data audits and curation.}
The composition of training data has substantial effects on model behavior
and generalization. The \emph{Pretrainer's Guide} isolates the influence of
data age, domain coverage, quality, and toxicity
~\citep{longpre-etal-2024-pretrainers}, while corpus audits have exposed
duplication, benchmark contamination, low-quality content, and gaps in
provenance. \emph{WIMBD}, for example, audits C4 and RedPajama for
contamination and undesirable content~\citep{elazar2024whats}, and
\emph{Data Portraits} develops membership-testing tools for tracing training
data~\citep{marone2023data}. Broader analyses also document provenance
imbalances across text, speech, and video resources, including heavy reliance
on web-crawled and Western-centric sources~\citep{longpre2025bridging}.

A parallel line of work studies how training corpora can be constructed and
selected more effectively. This includes practical recipes for
trillion-token datasets~\citep{parmar-etal-2024-data}, LLM-assisted quality
selection through \emph{QuRating}~\citep{10.5555/3692070.3694241},
register-aware sampling~\citep{myntti2025register}, domain-based organization
of web text~\citep{wettig2025organize}, and reuse of filtered web content~\citep{nguyen2025recycling}. Our analysis adds within-English variation to
this auditing perspective by measuring AmE representation across pretraining
and post-training resources and following the same directional asymmetry into
later stages of the pipeline.

\paragraph{Tokenization and representational asymmetry.}
Tokenization determines how linguistic forms are mapped into the discrete
units consumed by language models. Prior work documents large disparities
in subword segmentation across languages and scripts
~\citep{nayeem2025STRR}, with consequences for sequence length, effective
context capacity, latency, and monetary cost
~\citep{petrov2023language,alqahtani-etal-2026-stop}. Tokenization length
and corpus frequency can also correlate with demographic properties of
personal names, disproportionately fragmenting underrepresented forms
~\citep{an-rudinger-2023-nichelle}. Related robustness studies show that
lexical alternations, such as brand and generic drug names, can produce
substantial variation in model behavior
~\citep{gallifant-etal-2024-language}.

Other work examines the mechanisms and mitigation of such disparities.
In machine translation, training distributions and subword segmentation can
jointly disadvantage female and non-stereotypical gender inflections, while
targeted token-embedding adaptation can reduce these gaps
~\citep{iluz-etal-2023-exploring}. Controlled interventions further show
that changing a word from a multi-token representation to a single token can
substantially alter its probability
~\citep{lesci-etal-2025-causal}. Parity-Aware Byte-Pair Encoding similarly
trades a small amount of compression for more balanced token counts across
languages~\citep{foroutan2025parity}. Our analysis examines the same
representational layer within English through tokenization efficiency,
tokenizer provenance, and paired prediction cost.

\paragraph{English variation in NLP and LLMs.}
Research on language variation has documented systematic performance
differences across regional and social varieties of English, particularly
African American English (AAE) and South Asian Englishes (SAsE)
~\citep{demszky-etal-2021-learning,holt-etal-2024-perceptions,
10.1145/3712060}. For AAE, disparities have been reported in part-of-speech
tagging~\citep{jorgensen-etal-2016-learning}, language identification
~\citep{blodgett-etal-2016-demographic}, sentiment analysis
~\citep{kiritchenko-mohammad-2018-examining}, dependency parsing
~\citep{blodgett-etal-2018-twitter}, hate-speech detection
~\citep{sap-etal-2019-risk}, and natural-language understanding
~\citep{ziems-etal-2022-value}. Recent studies also show that LLMs can
reproduce negative stereotypes associated with AAE
~\citep{Hofmann2024,fleisig-etal-2024-linguistic}.

User-centered studies report that SAsE speakers encounter lower reliability
for their linguistic usage~\citep{holt-etal-2024-perceptions}, while
\emph{Multi-VALUE} evaluates models across many English varieties and
linguistic features, revealing broad robustness gaps
~\citep{ziems-etal-2023-multi}. Even differences between standardized
English conventions affect model behavior: retrieval performance decreases
when queries and documents use different spelling conventions
~\citep{10.1145/3539618.3592030}, and language models respond differently to
observed and novel spelling variants~\citep{nielsen-etal-2023-spelling}.

Our study places the American-English default within this literature on
linguistic variation, but follows it across the development pipeline rather
than a single downstream task. We examine its prevalence in pretraining and
post-training data, its expression in tokenization and model prediction cost, and its realization in generations. This pipeline-level analysis
provides a common empirical view of how the same AmE advantage recurs across
data exposure, representation, and generation.


\section{Brief Historical Context}
\label{sec:brief-history}

The divergence between American English (AmE) and British English (BrE)
reflects both deliberate standardization and broader historical change.
Samuel Johnson's 1755 dictionary helped codify spelling conventions that
became influential in British literary and educational practice. In the
United States, Noah Webster's 1828 dictionary promoted distinct and often
simplified spellings as part of an emerging American linguistic identity
~\citep{baker2017american}. These processes established many of the
orthographic contrasts still visible today.

\begin{figure*}[htbp]
    \centering
    \includegraphics[width=0.96\linewidth]
    {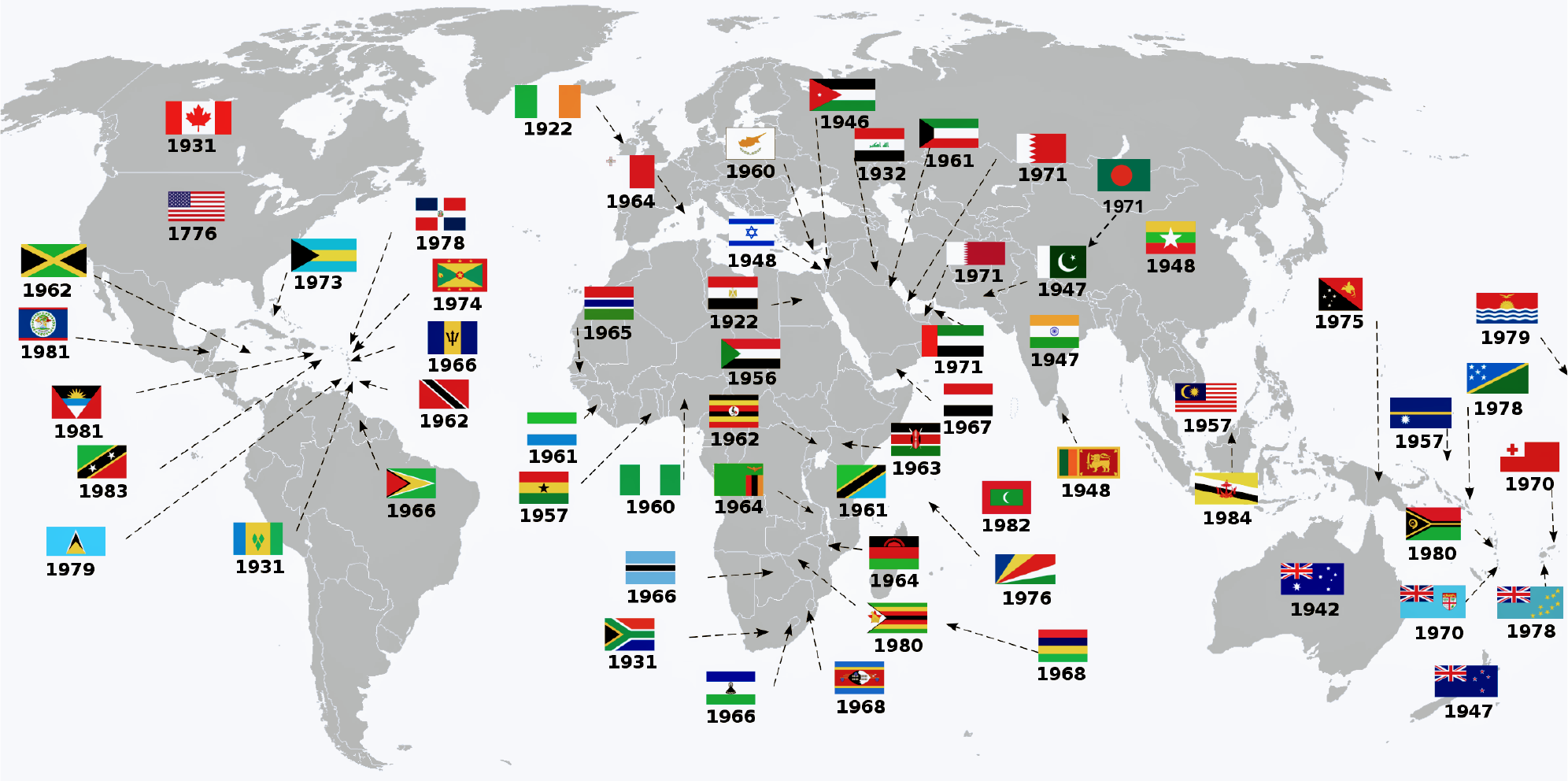}
    \caption{\small Timeline of independence across countries formerly under British colonization. The map highlights the wave of decolonization in the mid-twentieth century, when nations in Africa, Asia, the Caribbean, and the Pacific gained sovereignty. This geopolitical shift marked the decline of direct colonial governance but reinforced the institutional legacy of \emph{British English (BrE)} in education, government, journalism, and law across many of these regions~\citep{0b15e89a04f846adb87cf3cb75ed6a2b, 9781405198431}.
    }
    \label{fig:british-colonization}
\end{figure*}

The two standards subsequently acquired international reach through distinct
geopolitical trajectories. British colonial expansion embedded BrE
conventions in administration, education, journalism, and law across Africa,
Asia, the Caribbean, and the Pacific. These conventions often persisted
after political independence and remain institutionally influential across
much of the Commonwealth and in English-using contexts such as India,
Nigeria, Singapore, Australia, and New Zealand (Figure~\ref{fig:british-colonization})~\citep{calabrese2015variation,0b15e89a04f846adb87cf3cb75ed6a2b,
9781405198431}. This institutional footprint also persists through
UK-linked language and educational institutions, including the British
Council and international English assessment through IELTS.

American English expanded especially rapidly during the twentieth century
through U.S. economic influence, mass media and global entertainment,
technological development, and global commerce~\citep{crystal2003english,
10.1371/journal.pone.0197741,nordquist2024englishlanguage}. Hollywood,
popular culture, digital platforms, and technology ecosystems further
strengthened its international visibility, particularly in contemporary
online communication.

This history makes the AmE--BrE contrast useful for controlled analysis.
Both are extensively documented standard varieties with systematic
differences across spelling, vocabulary, grammar, conventions, and
multi-word usage. BrE also provides a meaningful reference for institutional
norms inherited by several other English varieties, while AmE has become
especially prominent in the digital ecosystem from which modern LLMs are
trained. Google Books provides separate reference corpora for the two
varieties, allowing \textsc{DiAlign} to measure regional alignment beyond
the curated one-to-one variant inventory.


\section{Scope, Limitations, and Future Directions}
\label{sec:limitations}

Our experiments use controlled comparisons designed to isolate the
American-English advantage across multiple stages of the LLM pipeline. This
design supports high-precision, cross-stage evidence and defines clear
directions for broader generalization.

\paragraph{Reference Variety and Broader English Variation.}
We use Ref(BrE) as a controlled, globally meaningful comparator because it
provides systematic contrasts with AmE across spelling, vocabulary, grammar,
and usage that can be carried consistently from corpus statistics through
tokenization, prediction cost, and generation. This prioritizes direct
cross-stage comparability while providing a template for broader regional
analysis. The same pipeline can be extended to Indian, Nigerian, Singapore,
Australian, New Zealand, and other English varieties when suitable reference
resources are available. Appendix~\ref{sec:dialign-generalization} describes how \textsc{DiAlign} can be instantiated with alternative corpora.

\paragraph{Lexicon Precision and Coverage.}
The \texttt{1,813}-pair resource is designed for high-precision one-to-one
comparison, which is especially important for tokenizer and counterfactual
likelihood experiments where both regional forms must be directly comparable.
We complement this resource with \textsc{DiAlign}, which captures
grammatical, structural, collocational, stylistic, and multi-word evidence
beyond the fixed lexicon. RQ1 uses type-level corpus counts for curated
variants, with context-sensitive cases removed when they materially affect
interpretation. Contextual modeling of larger web-scale variant inventories
provides a direct extension.

\paragraph{Reference Corpora and Meta-evaluation.}
Google Books provides separate American and British English distributions,
making it a consistent contrastive reference for \textsc{DiAlign}.
Normalized regional frequencies allow direct comparison of the same local
constructions across the two reference distributions. \textsc{DiAlign}
decomposes each input into overlapping 2--5-gram spans and aggregates retained
contrastive evidence at the local level. The resulting signal generalizes
across domains: \textsc{DiAlign} reaches 93.18\% accuracy on the external
BBC--HuffPost news benchmark, shows the same directional AmE pattern across
encyclopedic, web, and mixed-domain pretraining corpora, and extends to
model-generated open-domain responses in RQ3. Separately, the
\texttt{2000--2022} robustness analysis yields nearly identical performance
to the \texttt{1950--2022} setting, demonstrating temporal stability of the
regional contrast. A matched-domain benchmark with independently annotated
variety labels would further broaden evaluation across source conditions.

\paragraph{Generation Setting.}
RQ3 uses two contrasting open-domain registers under a fixed response-length
protocol to isolate regional preference while controlling generation length.
Natural Questions provides a more formal, encyclopedic setting, while ELI5
provides a more conversational one, and the observed preference is stable to
realized response length (\cref{app:rq3-length-robustness}). Long-form
dialogue, specialized professional domains, mixed-variety inputs, and
creative generation provide natural extensions for testing the default under
more complex discourse and user contexts.

\paragraph{Tokenizer and Checkpoint Coverage.}
Our tokenizer panel spans multiple model families, vocabulary sizes, and
development contexts, while the provenance audit separates independent
tokenizers from inherited infrastructure. Prediction-cost experiments focus
on five model families with matched public base and post-trained checkpoints,
prioritizing within-family comparability over nominal model count. Larger
closed models and families without comparable public checkpoints remain
represented where artifact access permits, including tokenizer and generation
experiments.

\paragraph{Attribution Across Pipeline Stages.}
The study traces where a recurring AmE asymmetry becomes visible across data
exposure, tokenization, prediction cost, and generation. The consistency
across these stages establishes a structural pattern. Causal decomposition
requires controlled interventions such as retraining with rebalanced data,
tokenizer replacement, or matched post-training interventions, providing a
natural next step from structural tracing to causal attribution.


\input{tables/lexicon-sources-desc}

\input{tables/AmE-BrE-differences-1}
\input{tables/AmE-BrE-differences-2}

%% file: tables/AmE-BrE-variations.tex
\begin{table*}[!htbp]
\centering
\renewcommand{\arraystretch}{1.3} 
\setlength{\tabcolsep}{2.5pt} 
\small 
\caption{\small
Distribution of the \texttt{1,813} matched AmE (\usflag) and Ref(BrE)
(\ukflag) variant pairs across linguistic categories in the curated
resource. We report each category's share of the full inventory with
representative examples, grouped into orthographic (\emph{spelling}) and
vocabulary-based contrasts.
}
\resizebox{14cm}{!}  
{
\begin{tabular}{cccclc}
\hline
\rowcolor[HTML]{F5F5F5} 
\textbf{Category} &
  \textbf{\begin{tabular}[c]{@{}c@{}}Difference Type\end{tabular}} &
  \textbf{\% of Pairs} &
  \multicolumn{3}{c}{\cellcolor[HTML]{F5F5F5}\textbf{Examples}} \\ \hline \hline
                    & ends in ``-or'' (AmE) \textbf{vs.} ``-our'' (BrE)  & 2.26\%  & color (\usflag) \textbf{vs.} colour (\ukflag) &  & labor (\usflag) \textbf{vs.} labour (\ukflag) \\
                    & ends in ``-ize'' (AmE) \textbf{vs.} ``-ise'' (BrE)  & 11.58\% & organize (\usflag) \textbf{vs.} organise (\ukflag) &  & realize (\usflag) \textbf{vs.} realise (\ukflag) \\ 
                    & ends in ``-er'' (AmE) \textbf{vs.} ``-re'' (BrE)            & 1.65\%  & center (\usflag) \textbf{vs.} centre (\ukflag) &  & liter (\usflag) \textbf{vs.} litre (\ukflag) \\
                    & ends in ``-og'' (AmE) \textbf{vs.} ``-ogue'' (BrE)          & 0.55\%  & dialog (\usflag) \textbf{vs.} dialogue (\ukflag) &  & catalog (\usflag) \textbf{vs.} catalogue (\ukflag) \\
                    & ends in ``-ense'' (AmE) \textbf{vs.} ``-ence'' (BrE)          & 0.22\%  & defense (\usflag) \textbf{vs.} defence (\ukflag) &  & pretense  (\usflag) \textbf{vs.} pretence  (\ukflag) \\
                    & ``e'' (AmE) \textbf{vs.} ``ae'' (BrE)          & 4.03\%  & esthetic (\usflag) \textbf{vs.} aesthetic (\ukflag) &  & pediatric (\usflag) \textbf{vs.} paediatric (\ukflag) \\
                    & words with single ``l'' \textbf{vs.} double ``l''   & 8.88\%  & traveler (\usflag) \textbf{vs.} traveller (\ukflag) &  & enroll (\usflag) \textbf{vs.} enrol (\ukflag) \\
\multirow{-8}{*}{\textbf{\begin{tabular}[c]{@{}c@{}}Orthographic/\\ Spelling\end{tabular}}} &
  sublexical spelling variation &
  49.75\% &
  jewelry  (\usflag) \textbf{vs.} jewellery (\ukflag) &
   &
  program (\usflag) \textbf{vs.} programme (\ukflag)  \\ \hline
\textbf{Vocabulary} & different lexical items entirely & 21.07\% & elevator (\usflag) \textbf{vs.} lift (\ukflag) &  & flashlight (\usflag) \textbf{vs.} torch (\ukflag) \\ \hline
\end{tabular}
}
\label{tab:AmE-BrE-variations}
\end{table*}

%% file: tables/AmE-BrE-overview.tex
\begin{table*}[t]
\centering
\renewcommand{\arraystretch}{1.4}
\setlength{\tabcolsep}{4pt}
\footnotesize
\caption{\small
Representative distinctions between American English (AmE) (\usflag) and
the British English reference, Ref(BrE) (\ukflag), across orthography,
vocabulary, grammar, and writing conventions. Examples reflect common
majority-preference usage and are illustrative rather than exhaustive.
Additional examples are provided in
Tables~\ref{tab:br-am-core} and~\ref{tab:br-am-phrases} of the Appendix.
\vspace{-1mm}
}
\resizebox{0.99\textwidth}{!}{
\begin{tabular}{p{2.2cm} p{6.2cm} p{6.2cm}}
\toprule
\textbf{Level}
& \textbf{American English (AmE) (\usflag)}
& \textbf{Reference: British English (BrE) (\ukflag)} \\
\midrule

\textbf{Orthography}
& \emph{color, center, organize, traveling, favorite, check, jewelry, program, catalog, defense}
& \emph{colour, centre, organise, travelling, favourite, cheque, jewellery, programme, catalogue, defence} \\

\textbf{Vocabulary}
& \emph{cell phone, crosswalk, downtown, train station, parking lot, ZIP code, vacation, apartment, elevator, line, sidewalk, flashlight, soda, popsicle}
& \emph{mobile phone, zebra crossing, city centre, railway station, car park, postcode, holiday, flat, lift, queue, pavement, torch, fizzy drink, ice lolly} \\

\textbf{Grammar/usage}
& \emph{I just ate}; \emph{on the weekend}; \emph{Do you have a pen?}; \emph{The team is winning}; \emph{different than}
& \emph{I've just eaten}; \emph{at the weekend}; \emph{Have you got a pen?}; \emph{The team are winning}; \emph{different from} \\

\textbf{Conventions}
& double quotes; punctuation inside quotes; \emph{December 31, 2024}; \emph{first floor}; \emph{11:15 PM}; \emph{$68^\circ$F}
& single quotes; punctuation outside quotes; \emph{31 December 2024}; \emph{ground floor}; \emph{11.15 pm}; \emph{$20^\circ$C} \\

\bottomrule
\end{tabular}
}
\vspace{-1.5mm}
\label{tab:bre_ame_overview}
\end{table*}

%% file: tables/dialign-ablation.tex
\begin{table*}[!htbp]
\centering
\small
\caption{\small Ablation study of \textsc{DiAlign} on the balanced AmE--Ref(BrE) meta-evaluation benchmark. We report classification performance (Accuracy, Precision, Recall, and F1) and mean confidence for AmE and Ref(BrE) predictions. Removing either divergence weighting (DW) or the boosting factor (BF) reduces performance, with the largest drop when both components are removed.
}
\resizebox{14cm}{!}  
{
\begin{tabular}{lccccccc}
\toprule
\textbf{Ablations} & \textbf{Accuracy} & \textbf{Precision} & \textbf{Recall} & \textbf{F1 Score} & \textbf{Avg. Conf. (AmE)} & \textbf{Avg. Conf. (BrE)} \\
\midrule
\textsc{DiAlign} (final)         & 93.18 & 90.67 & 96.25 & 93.38 & 0.84 & 0.91 \\
\midrule
-- w/o Divergence Weight (DW)    & 92.25 & 89.29 & 95.98 & 92.52 & 0.77 & 0.85 \\
-- w/o Boosting Factor (BF)      & 92.25 & 89.29 & 95.98 & 92.52 & 0.82 & 0.89 \\
-- w/o Both (DW + BF)            & 91.31 & 88.13 & 95.45 & 91.65 & 0.75 & 0.83 \\
\bottomrule
\end{tabular}
}
\label{tab:dialign-ablation}
\end{table*}

%% file: tables/posttraining_dataset_full.tex
\begin{table*}[t]
\centering
\small
\caption{\small \textbf{Dataset-level post-training audit.}
\textbf{Hits} counts occurrences matching either member of the curated AmE--Ref(BrE) variant pairs. \textbf{Weighted AmE (\%)} reports the aggregate AmE share across observed hits, while \textbf{Orth. AmE (\%)} and \textbf{Vocab. AmE (\%)} report mean pair-level AmE shares within orthographic and vocabulary categories, respectively. \textbf{HF} links identify the exact Hugging Face dataset releases used in the audit.}
\label{tab:posttraining_dataset_full}

\setlength{\tabcolsep}{3.5pt}
\renewcommand{\arraystretch}{1.03}

\resizebox{\linewidth}{!}{%
\begin{tabular}{llrrcccc}
\toprule
\textbf{Dataset} &
\textbf{Stage} &
\textbf{Rows} &
\textbf{Hits} &
\makecell{\textbf{Weighted}\\\textbf{AmE (\%)}} &
\makecell{\textbf{Orth.}\\\textbf{AmE (\%)}} &
\makecell{\textbf{Vocab.}\\\textbf{AmE (\%)}} &
\textbf{HF} \\
\midrule

Alpaca Cleaned~\citep{alpaca}
& SFT & 51,760 & 74,991 & 85.35 & 94.84 & 80.05
& \href{https://huggingface.co/datasets/yahma/alpaca-cleaned}{HF} \\

Databricks Dolly 15k~\citep{DatabricksBlog2023DollyV2}
& SFT & 15,011 & 19,997 & 65.29 & 78.09 & 67.65
& \href{https://huggingface.co/datasets/databricks/databricks-dolly-15k}{HF} \\

OpenAssistant Conversations v2~\citep{pf2023openassistant}
& SFT & 61,278 & 61,030 & 83.08 & 92.56 & 76.85
& \href{https://huggingface.co/datasets/OpenAssistant/oasst2}{HF} \\

UltraChat 200k~\citep{ding-etal-2023-enhancing}
& SFT & 207,865 & 2,442,647 & 84.23 & 90.32 & 75.68
& \href{https://huggingface.co/datasets/HuggingFaceH4/ultrachat_200k}{HF} \\

OpenHermes 2.5~\citep{OpenHermes}
& SFT & 1,001,551 & 2,517,105 & 81.81 & 85.78 & 74.01
& \href{https://huggingface.co/datasets/teknium/OpenHermes-2.5}{HF} \\

WizardLM Evol-Instruct v2~\citep{xu2024wizardlm}
& SFT & 143,000 & 634,919 & 89.72 & 95.20 & 80.33
& \href{https://huggingface.co/datasets/WizardLMTeam/WizardLM_evol_instruct_V2_196k}{HF} \\

No Robots~\citep{no-robots}
& SFT & 9,500 & 20,951 & 76.51 & 90.18 & 77.66
& \href{https://huggingface.co/datasets/HuggingFaceH4/no_robots}{HF} \\

LIMA~\citep{NEURIPS2023_ac662d74}
& SFT & 1,030 & 5,348 & 78.78 & 89.05 & 74.62
& \href{https://huggingface.co/datasets/GAIR/lima}{HF} \\

FLAN Collection~\citep{pmlr-v202-longpre23a}
& SFT & 1,386,050 & 2,099,656 & 69.78 & 76.20 & 69.85
& \href{https://huggingface.co/datasets/Muennighoff/flan}{HF} \\

Super-NaturalInstructions~\citep{wang-etal-2022-super}
& SFT & 3,082,094 & 4,479,160 & 72.71 & 80.41 & 73.01
& \href{https://huggingface.co/datasets/Muennighoff/natural-instructions}{HF} \\

ShareGPT Vicuna~\citep{vicuna2023}
& SFT & 120,675 & 1,204,637 & 84.35 & 91.40 & 79.69
& \href{https://huggingface.co/datasets/Aeala/ShareGPT_Vicuna_unfiltered}{HF} \\

OpenOrca~\citep{mukherjee2023}
& SFT & 4,233,923 & 10,632,811 & 71.79 & 80.65 & 72.21
& \href{https://huggingface.co/datasets/Open-Orca/OpenOrca}{HF} \\

\addlinespace
\midrule
\addlinespace

Anthropic HH-RLHF~\citep{bai2022training}
& RLHF & 160,800 & 488,280 & 78.87 & 93.85 & 80.91
& \href{https://huggingface.co/datasets/Anthropic/hh-rlhf}{HF} \\

OpenAI WebGPT Comparisons~\citep{nakano2022webgpt}
& RLHF & 19,578 & 55,193 & 79.50 & 91.58 & 79.97
& \href{https://huggingface.co/datasets/openai/webgpt_comparisons}{HF} \\

Stanford Human Preferences~\citep{pmlr-v162-ethayarajh22a}
& RLHF & 348,718 & 1,013,154 & 72.40 & 84.36 & 73.05
& \href{https://huggingface.co/datasets/stanfordnlp/SHP}{HF} \\

UltraFeedback Binarized~\citep{pmlr-v235-cui24f}
& RLHF & 61,135 & 390,601 & 82.69 & 90.46 & 77.77
& \href{https://huggingface.co/datasets/HuggingFaceH4/ultrafeedback_binarized}{HF} \\

OpenAI Summarize Comparisons~\citep{NEURIPS2020-1f89885d}
& RLHF & 92,534 & 176,589 & 73.18 & 85.31 & 74.52
& \href{https://huggingface.co/datasets/CarperAI/openai_summarize_comparisons}{HF} \\

HelpSteer2~\citep{wang2024helpsteer}
& RLHF & 20,324 & 77,332 & 84.72 & 92.87 & 75.87
& \href{https://huggingface.co/datasets/nvidia/HelpSteer2}{HF} \\

Nectar~\citep{zhu2024starlingb}
& RLHF & 182,954 & 2,044,872 & 85.09 & 93.15 & 79.39
& \href{https://huggingface.co/datasets/berkeley-nest/Nectar}{HF} \\

PKU-SafeRLHF~\citep{ji-etal-2025-pku}
& RLHF & 73,907 & 178,469 & 86.86 & 90.76 & 77.96
& \href{https://huggingface.co/datasets/PKU-Alignment/PKU-SafeRLHF}{HF} \\

BeaverTails~\citep{ji2023beavertails}
& RLHF & 300,567 & 242,377 & 83.48 & 92.52 & 79.63
& \href{https://huggingface.co/datasets/PKU-Alignment/BeaverTails}{HF} \\

\addlinespace
\midrule
\addlinespace

\textbf{Overall}
& --
& \textbf{11,574,254}
& \textbf{28,860,119}
& \textbf{76.17}
& \textbf{88.55}
& \textbf{76.22}
& -- \\

\bottomrule
\end{tabular}
}
\end{table*}

%% file: tables/tokenizer_lineage.tex
\begin{table*}[t]
\centering
\small
\caption{\small
\textbf{Tokenizer provenance audit.}
Model developer country and tokenizer provenance are reported separately. \textbf{Independent} refers specifically to tokenizer-vocabulary provenance, not to model weights, architecture, or training data. Derived tokenizers are retained in the behavioral analysis but are not counted as independent country-level evidence.
}
\label{tab:tokenizer_lineage}

\setlength{\tabcolsep}{1.5pt}
\renewcommand{\arraystretch}{1.25}

\resizebox{\linewidth}{!}{%
\begin{tabular}{lclcl}
\toprule
\textbf{Tokenizer} &
\textbf{Developer Country } &
\textbf{Tokenizer lineage} &
\textbf{Independent} &
\textbf{Interpretation} \\
\midrule

GPT-4
& USA (\usflag)
& OpenAI \texttt{cl100k\_base}
& Yes
& Independent US baseline \\

GPT-4o
& USA (\usflag)
& OpenAI \texttt{o200k\_base}
& Yes
& Independent US baseline \\

Llama-3.3-70B
& USA (\usflag)
& Extended from US-origin tiktoken-style vocabulary
& No
& Derived; excluded from independent country-level evidence \\

Gemma-3-27B
& USA (\usflag)
& Gemma/Gemini-family vocabulary
& Yes
& Independent US tokenizer \\

DeepSeek-V3
& China (\chinaflag)
& DeepSeek BPE
& Yes
& Independent non-US tokenizer \\

Mistral-Small-24B
& France (\franceflag)
& Mistral Tekken family
& Yes
& Independent non-US tokenizer \\

StableLM-2-1.6B
& UK (\ukflag)
& Arcade100k/cl100k-like tiktoken BPE
& No
& Derived; not independent UK evidence \\

Velvet-2B
& Italy (\italyflag)
& Velvet-family vocabulary
& Yes
& Independent non-US tokenizer \\

Falcon3-7B
& UAE (\uaeflag)
& Falcon3 BPE
& Yes
& Independent non-US tokenizer \\

\bottomrule
\end{tabular}
}
\end{table*}

%% file: tables/tokenizer_boundary_similarity.tex
\begin{table*}[t]
\centering
\small
\caption{\textbf{Tokenizer behavioral similarity with GPT-4.}
Values report the percentage of strings for which each tokenizer produces the same token-boundary sequence as GPT-4. StableLM-2 and Llama-3.3 show near-perfect agreement, in contrast to the substantially lower agreement observed for independently sourced tokenizer families.}
\label{tab:tokenizer_boundary_similarity}

\setlength{\tabcolsep}{5pt}
\renewcommand{\arraystretch}{1.05}

\begin{tabular}{lccccc}
\toprule
\textbf{Tokenizer compared with GPT-4} &
\textbf{AmE} &
\textbf{BrE} &
\textbf{Validation} &
\textbf{Generated} &
\textbf{All} \\
\midrule

StableLM-2-1.6B
& \textbf{100.00}
& \textbf{100.00}
& 80.80
& 81.17
& 91.87 \\

Llama-3.3-70B
& 99.17
& 99.06
& 99.87
& 99.83
& \textbf{99.43} \\

GPT-4o
& 74.85
& 77.05
& 40.27
& 29.00
& 58.58 \\

DeepSeek-V3
& 67.02
& 70.22
& 47.80
& 25.08
& 55.42 \\

Mistral-Small-24B
& 65.47
& 66.91
& 26.47
& 10.00
& 46.11 \\

Falcon3-7B
& 58.58
& 58.91
& 11.93
& 5.17
& 37.48 \\

Velvet-2B
& 20.63
& 35.58
& 20.20
& 9.08
& 22.62 \\

\bottomrule
\end{tabular}
\end{table*}

%% file: tables/rq2-loss-gap-appendix.tex
\begin{table}[t]
\centering
\small
\setlength{\tabcolsep}{5.2pt}
\renewcommand{\arraystretch}{1.08}
\caption{\textbf{Prediction-cost asymmetry on the tokenizer-controlled RQ2 slice.} Rows contain orthographic minimal-mention counterfactuals for which the \AmE{} and \RefBrE{} sentences receive identical token counts under the evaluated model. \textbf{Gap/token} denotes the \RefBrE{} minus \AmE{} difference in per-token negative log-likelihood, with positive values indicating higher prediction cost for \RefBrE{}. Confidence intervals are paired-bootstrap 95\% intervals.
}
\begin{tabular}{llrrrr}
\toprule
\textbf{Family} &
\textbf{Stage} &
\textbf{Rows} &
\makecell{\textbf{Gap /}\\\textbf{token}} &
\textbf{95\% CI} &
\makecell{\textbf{\RefBrE{}}\\\textbf{higher}} \\
\midrule

Llama
& Base
& 1,008
& 0.189
& [0.185, 0.193]
& 100.0\% \\

Llama
& Post
& 1,008
& 0.264
& [0.262, 0.267]
& 100.0\% \\

StableLM
& Base
& 2,350
& 0.210
& [0.204, 0.214]
& 100.0\% \\

StableLM
& Post
& 2,350
& 0.400
& [0.389, 0.411]
& 100.0\% \\

Ministral
& Base
& 1,068
& 0.213
& [0.208, 0.218]
& 100.0\% \\

Ministral
& Post
& 1,068
& 0.234
& [0.229, 0.239]
& 100.0\% \\

Gemma
& Base
& 1,448
& 0.226
& [0.222, 0.230]
& 100.0\% \\

Gemma
& Post
& 1,448
& 0.340
& [0.329, 0.351]
& 100.0\% \\

Falcon3
& Base
& 2,350
& 0.303
& [0.295, 0.310]
& 100.0\% \\

Falcon3
& Post
& 2,350
& 0.420
& [0.408, 0.430]
& 100.0\% \\

\bottomrule
\end{tabular}
\label{tab:rq2-loss-gap-appendix}
\end{table}

%% file: tables/rq2-checkpoints.tex
\begin{table}[t]
\centering
\small
\setlength{\tabcolsep}{6.8pt}
\renewcommand{\arraystretch}{1.08}
\caption{
Matched base and post-trained checkpoints used for the RQ2 prediction-cost
analysis, spanning five model families and ten checkpoints.
}
\begin{tabular}{lll}
\toprule
\textbf{Family}
& \textbf{Stage}
& \textbf{Hugging Face checkpoint} \\
\midrule

Gemma
& Base
& \href{https://huggingface.co/google/gemma-3-1b-pt}
{\texttt{google/gemma-3-1b-pt}} \\

Gemma
& Post-trained
& \href{https://huggingface.co/google/gemma-3-1b-it}
{\texttt{google/gemma-3-1b-it}} \\

Ministral
& Base
& \href{https://huggingface.co/mistralai/Ministral-3-3B-Base-2512}
{\texttt{mistralai/Ministral-3-3B-Base-2512}} \\

Ministral
& Post-trained
& \href{https://huggingface.co/mistralai/Ministral-3-3B-Instruct-2512}
{\texttt{mistralai/Ministral-3-3B-Instruct-2512}} \\

Llama
& Base
& \href{https://huggingface.co/meta-llama/Llama-3.1-8B}
{\texttt{meta-llama/Llama-3.1-8B}} \\

Llama
& Post-trained
& \href{https://huggingface.co/meta-llama/Llama-3.1-8B-Instruct}
{\texttt{meta-llama/Llama-3.1-8B-Instruct}} \\

StableLM
& Base
& \href{https://huggingface.co/stabilityai/stablelm-2-1_6b}
{\texttt{stabilityai/stablelm-2-1\_6b}} \\

StableLM
& Post-trained
& \href{https://huggingface.co/stabilityai/stablelm-2-zephyr-1_6b}
{\texttt{stabilityai/stablelm-2-zephyr-1\_6b}} \\

Falcon3
& Base
& \href{https://huggingface.co/tiiuae/Falcon3-7B-Base}
{\texttt{tiiuae/Falcon3-7B-Base}} \\

Falcon3
& Post-trained
& \href{https://huggingface.co/tiiuae/Falcon3-7B-Instruct}
{\texttt{tiiuae/Falcon3-7B-Instruct}} \\

\bottomrule
\end{tabular}

\label{tab:rq2-checkpoints}
\end{table}

%% file: tables/rq2-layerwise-semantic-equivalence.tex
\begin{table}[t]
\centering
\small
\setlength{\tabcolsep}{6pt}
\renewcommand{\arraystretch}{1.08}
\caption{\small
\textbf{Layerwise semantic-equivalence control.}
Values report model-level mean cosine similarities averaged across ten
evaluated checkpoints. Matched \AmE{}--\RefBrE{} counterfactuals are more
similar than character-perturbed and unrelated controls across layers, with
the clearest separation at the final layer.
}
\begin{tabular}{llrr}
\toprule
\textbf{Layer} &
\textbf{Comparison} &
\textbf{Models} &
\makecell{\textbf{Mean}\\\textbf{cosine}} \\
\midrule

Embedding
& \AmE{}--\RefBrE{} counterfactual
& 10
& 0.902 \\

Embedding
& Character perturbation
& 10
& 0.824 \\

Embedding
& Unrelated control
& 10
& 0.504 \\

\addlinespace

Middle
& \AmE{}--\RefBrE{} counterfactual
& 10
& 1.000 \\

Middle
& Character perturbation
& 10
& 0.999 \\

Middle
& Unrelated control
& 10
& 0.997 \\

\addlinespace

Final
& \AmE{}--\RefBrE{} counterfactual
& 10
& 0.989 \\

Final
& Character perturbation
& 10
& 0.928 \\

Final
& Unrelated control
& 10
& 0.814 \\

\bottomrule
\end{tabular}
\label{tab:rq2-layerwise-semantic-equivalence}
\end{table}

%% file: tables/length-control-app.tex
\begin{table*}[t]
\vspace{2mm}
\centering
\renewcommand{\arraystretch}{1.35} 
\setlength{\tabcolsep}{4pt} 
\small 
\caption{\small
Word-length adherence across models in RQ3
(Section~\ref{sec:evaluate-generation}). Each model was instructed to
produce exactly 50 words per response. We report mean length, standard deviation, and range for Natural Questions (\emph{formal}) and ELI5 (\emph{informal}) under default English and British English (en-GB) prompting.
}
\resizebox{14cm}{!}  
{
\begin{tabular}{lccccccccccccc}
\hline
\rowcolor[HTML]{F5F5F5} 
\multicolumn{1}{c}{\cellcolor[HTML]{F5F5F5}} & \cellcolor[HTML]{F5F5F5} & \textbf{} & \multicolumn{5}{c}{\cellcolor[HTML]{F5F5F5} {\cellcolor{blue!10}\textbf{Natural Questions (NQ) {[}formal{]}}}} & \textbf{} & \multicolumn{5}{c}{\cellcolor[HTML]{F5F5F5} {\cellcolor{cyan!8}\textbf{ELI5 {[}informal{]}}}} \\ \cline{4-8} \cline{10-14} 
\rowcolor[HTML]{F5F5F5} 
\multicolumn{1}{c}{\cellcolor[HTML]{F5F5F5}} & \cellcolor[HTML]{F5F5F5} & \textbf{} & \multicolumn{2}{c}{\cellcolor[HTML]{F5F5F5}\textbf{\begin{tabular}[c]{@{}c@{}}Default English\\ ($\%\text{AmE}^{\text{Default}}$)\end{tabular}}} &  & \multicolumn{2}{c}{\cellcolor[HTML]{F5F5F5}\textbf{\begin{tabular}[c]{@{}c@{}}British English\\ ($\%\text{AmE}^{\text{BrE}}$)\end{tabular}}} & \textbf{} & \multicolumn{2}{c}{\cellcolor[HTML]{F5F5F5}\textbf{\begin{tabular}[c]{@{}c@{}}Default English\\ ($\%\text{AmE}^{\text{Default}}$)\end{tabular}}} &  & \multicolumn{2}{c}{\cellcolor[HTML]{F5F5F5}\textbf{\begin{tabular}[c]{@{}c@{}}British English\\ ($\%\text{AmE}^{\text{BrE}}$)\end{tabular}}} \\ \cline{4-5} \cline{7-8} \cline{10-11} \cline{13-14} 
\rowcolor[HTML]{F5F5F5} 
\multicolumn{1}{c}{\multirow{-3}{*}{\cellcolor[HTML]{F5F5F5}\textbf{LLMs}}} & \multirow{-3}{*}{\cellcolor[HTML]{F5F5F5}\textbf{\begin{tabular}[c]{@{}c@{}}Model \\ Access\end{tabular}}} & \textbf{} & \textbf{\begin{tabular}[c]{@{}c@{}}\#Words\\ Avg. {[}SD{]}\end{tabular}} & \textbf{\begin{tabular}[c]{@{}c@{}}Range\\ {[}min-max{]}\end{tabular}} &  & \textbf{\begin{tabular}[c]{@{}c@{}}\#Words\\ Avg. {[}SD{]}\end{tabular}} & \textbf{\begin{tabular}[c]{@{}c@{}}Range\\ {[}min-max{]}\end{tabular}} & \textbf{} & \textbf{\begin{tabular}[c]{@{}c@{}}\#Words\\ Avg. {[}SD{]}\end{tabular}} & \textbf{\begin{tabular}[c]{@{}c@{}}Range\\ {[}min-max{]}\end{tabular}} &  & \textbf{\begin{tabular}[c]{@{}c@{}}\#Words\\ Avg. {[}SD{]}\end{tabular}} & \textbf{\begin{tabular}[c]{@{}c@{}}Range\\ {[}min-max{]}\end{tabular}} \\ \hline \hline

\href{https://platform.openai.com/docs/models/gpt-4o}{GPT-4o} & \closedSource &  & 50.19 {\small \color{gray}[1.75]} & [46–57] &  & 50.32 {\small \color{gray}[1.61]} & [45–56] &  & 50.41 {\small \color{gray}[1.60]} & [47–55] &  & 50.35 {\small \color{gray}[1.53]} & [46–55] 

\\

\href{https://cloud.google.com/vertex-ai/generative-ai/docs/models/gemini/2-0-flash}{Gemini-2.0-flash} & \closedSource &  & 52.74 {\small \color{gray}[2.71]} & [46–60] &  & 51.92 {\small \color{gray}[2.89]} & [45–61] &  & 53.10 {\small \color{gray}[2.62]} & [47–61] &  & 52.82 {\small \color{gray}[2.95]} & [44–60] 

\\

\href{https://www.anthropic.com/news/claude-3-7-sonnet}{Claude-3.7-sonnet} & \closedSource &  & 45.25 {\small \color{gray}[2.29]} & [39–52] &  & 45.49 {\small \color{gray}[2.35]} & [39–57] &  & 47.70 {\small \color{gray}[2.45]} & [42–56] &  & 47.38 {\small \color{gray}[2.32]} & [41–54] 

\\ \hdashline

\href{https://huggingface.co/meta-llama/Llama-3.3-70B-Instruct}{Llama-3.3-70B} & \openSource &  & 41.68 {\small \color{gray}[7.12]} & [16–50] &  & 41.47 {\small \color{gray}[7.44]} & [18–50] &  & 41.33 {\small \color{gray}[7.83]} & [21–50] &  & 40.10 {\small \color{gray}[8.48]} & [18–51] 

\\

\href{https://huggingface.co/google/gemma-3-27b-it}{Gemma-3-27B} & \openSource &  & 49.03 {\small \color{gray}[2.82]} & [42–58] &  & 49.19 {\small \color{gray}[2.64]} & [44–61] &  & 49.18 {\small \color{gray}[2.84]} & [43–60] &  & 49.90 {\small \color{gray}[2.74]} & [44–58] 

\\

\href{https://huggingface.co/deepseek-ai/DeepSeek-V3-0324}{DeepSeek-V3} & \openSource &  & 53.03 {\small \color{gray}[4.13]} & [48–102] &  & 53.40 {\small \color{gray}[5.10]} & [47–103] &  & 53.60 {\small \color{gray}[4.66]} & [48–98] &  & 53.33 {\small \color{gray}[4.97]} & [4–96]

\\

\href{https://huggingface.co/mistralai/Mistral-Small-24B-Instruct-2501}{Mistral-Small-24B} & \openSource &  & 51.44 {\small \color{gray}[18.80]} & [13–318] &  & 50.18 {\small \color{gray}[10.97]} & [20–88] &  & 57.26 {\small \color{gray}[11.68]} & [23–120] &  & 56.78 {\small \color{gray}[10.43]} & [34–101]

\\

\href{https://huggingface.co/stabilityai/stablelm-2-zephyr-1_6b}{StableLM-2-1.6B} & \openSource &  & 82.94 {\small \color{gray}[39.90]} & [8–364] &  & 74.33 {\small \color{gray}[40.15]} & [8–351] &  & 96.82 {\small \color{gray}[24.23]} & [23–199] &  & 92.64 {\small \color{gray}[24.76]} & [41–199] 

\\

\href{https://huggingface.co/Almawave/Velvet-2B}{Velvet-2B}  & \openSource &  & 47.07 {\small \color{gray}[30.94]} & [8–406] &  & 44.56 {\small \color{gray}[23.03]} & [8–135] &  & 65.48 {\small \color{gray}[18.55]} & [22–122] &  & 63.10 {\small \color{gray}[17.25]} & [31–120] 

\\

\href{https://huggingface.co/tiiuae/Falcon3-7B-Instruct}{Falcon3-7B} & \openSource &  & 44.01 {\small \color{gray}[10.87]} & [18–83] &  & 41.72 {\small \color{gray}[10.45]} & [19–83] &  & 49.41 {\small \color{gray}[9.47]} & [25–80] &  & 48.14 {\small \color{gray}[9.39]} & [23–91] \\ \hline
\end{tabular}
}
\label{tab:length-control-app}
\end{table*}

%% file: tables/lexicon-sources-desc.tex
\clearpage
\begin{table}[h!]
\centering
\scriptsize
\renewcommand{\arraystretch}{2.0} 

\caption{
Key linguistic and web-based sources used to construct the
AmE--Ref(BrE) variant lexicon. Variant pairs were manually compiled across multiple sources, merged, and deduplicated to form a unified reference set for consistent analysis.
}
\begin{tabular}{p{1.9cm} p{3.5cm} p{7cm}}
\toprule
\textbf{Source} & \textbf{Title (\emph{linked})} & \textbf{Description} \\ \hline
\midrule

\textbf{Wikipedia} & \href{https://en.wikipedia.org/wiki/American_and_British_English_spelling_differences}{American and British English spelling differences} & A widely cited reference outlining systematic orthographic differences between American and British English. The page provides examples of variant spellings (e.g., \emph{color} vs. \emph{colour}), historical background, and explanations of regional conventions. It served as one of the authentic linguistic resources for curating consistent one-to-one variant pairs in our lexicon. \\ \hline

\textbf{ThoughtCo.} & \href{https://www.thoughtco.com/american-english-to-british-english-4010264}{American English to British English Vocabulary} & A curated reference list of American and British English vocabulary differences, created by experienced educators and subject experts. Provides reliable lexical contrasts in an accessible format, supporting the construction of our AmE--BrE lexicon. \\ \hline

\textbf{Research Article} & \href{https://journals.plos.org/plosone/article?id=10.1371/journal.pone.0197741}{Mapping the Americanization of English in Space and Time} & An empirical study tracing how American English variants spread globally across regions and over time. Offers quantitative evidence of AmE–BrE lexical contrasts, providing authoritative grounding for the curated variant pairs in our unified lexicon. \\ \hline

\textbf{IELTS} & \href{https://ielts.idp.com/canada/prepare/article-british-vs-american-english}{British vs. American English in the IELTS Test: Key Differences} &  
An official IELTS guide highlighting key vocabulary, spelling, and grammar differences between AmE and BrE. The resource systematically documents contrasts across domains such as food, school, homes, and grammar, making it a practical reference for understanding standardized English variations. \\ \hline

\textbf{Grammarly} & \href{https://www.grammarly.com/blog/product/how-to-switch-dialects/}{How to Select Your English Dialect} & A practical guide from Grammarly explaining how to switch between English dialects in writing tools, highlighting spelling, vocabulary, and usage variations (AmE vs BrE). Because it enumerates common dialectal choices in real writing, it serves as a useful supplementary resource for identifying variant pairs. \\ \hline

\textbf{SpellZone}  & \href{https://www.spellzone.com/blog/sixty_american_english_words_and_their_british_english_counterparts.htm}{Sixty American English Words and their British English Counterparts} & SpellZone provides a practical reference list of 60 common AmE–BrE word pairs, illustrating clear lexical contrasts in spelling and vocabulary. The resource highlights straightforward one-to-one mappings useful for systematic dialectal analysis. \\ \hline

\textbf{IELTS} & \href{https://ieltsidpindia.com/blog/british-vs-american-english}{Differences between British vs.\ American English} & A guidance article from IELTS that outlines vocabulary, spelling, and grammatical contrasts between AmE and BrE, emphasizing how learners must maintain internal consistency between the dialects. This resource helps validate by showing differences accepted in international testing and educational settings. \\ \hline

\textbf{SpellZone}  & \href{https://www.spellzone.com/pages/british-american.cfm}{Differences between British and American English spelling} & It provides an overview of common orthographic contrasts (e.g.\ “colour/color”, “centre/center”, “-re” vs “-er”) between BrE and AmE. This resource was used as a web-based lexicon support to validate our curated variant pairs. \\ \hline

\textbf{British Council} & \href{https://www.britishcouncilfoundation.id/en/english/articles/british-and-american-english}{Differences between British and American English} & An educational article by the British Council outlining vocabulary, grammar, and spelling distinctions between British and American English. It supports validation of variant pairs and highlights pedagogically recognized dialectal contrasts. \\ \hline

\textbf{IELTS Liz} & \href{https://ieltsliz.com/uk-us-spelling-main-differences/}{UK US Spelling Main Differences} & A practical guide by IELTS Liz summarizing the core orthographic differences between British and American spelling. This resource helps cross-check variant consistency and supports the curated lexicon’s alignment with real exam-related usage. \\ \hline

\textbf{Word Finder} & \href{https://wordfinder.yourdictionary.com/blog/british-vs-american-english-words-more-than-an-occasional-u/}{British vs. American English Words} & A comparative list of British and American English words, highlighting more than just spelling shifts ("-u") and covering vocabulary contrasts in everyday usage. It offers additional variant candidates and informs our lexicon selection process. \\ \hline

\textbf{Language Gallery} & \href{https://www.thelanguagegallery.com/blog/british-vs-american-spelling-what-s-the-difference}{British VS American Spelling: What’s the Difference?} & A language-education blog article detailing common orthographic differences between British and American English (e.g., “realise/realize,” “theatre/theater”). It served as a supplementary web-based lexicon to inform our manual variant curation. \\

\bottomrule
\end{tabular}
\label{tab:lexicon-sources}
\end{table}

%% file: tables/AmE-BrE-differences-1.tex
\begin{table}[t]
\centering
\footnotesize
\renewcommand{\arraystretch}{1.15}
\caption{\small
Representative AmE--Ref(BrE) distinctions across orthography, grammar, and formatting conventions. Entries reflect common majority-preference usage and are illustrative rather than exhaustive.
}
\label{tab:br-am-core}
\begin{tabular}{p{3.8cm} p{4.4cm} p{4.4cm}}
\toprule
Category & British English (BrE) & American English (AmE) \\
\midrule
o vs.\ ou & \textit{colour, honour, behaviour} & \textit{color, honor, behavior} \\
-re vs.\ -er endings & \textit{centre, fibre, theatre} & \textit{center, fiber, theater} \\
-ise vs.\ -ize endings & \textit{recognise, authorise} & \textit{recognize, authorize} \\
-yse vs.\ -yze endings & \textit{analyse, paralyse, catalyse} & \textit{analyze, paralyze, catalyze} \\
Single vs.\ double l (inflection) & \textit{travelled, counselled} & \textit{traveled, counseled} \\
-ll + -ly suffix & \textit{skilfully, wilfully} & \textit{skillfully, willfully} \\
Composite vowels & \textit{anaesthetic, diarrhoea, paediatric, oestrogen} & \textit{anesthetic, diarrhea, pediatric, estrogen} \\
Final silent -e/-ue & \textit{catalogue, analogue, axe} & \textit{catalog, analog, ax} \\
Silent -e before suffix & \textit{ageing, likeable} & \textit{aging, likable} \\
-ce vs.\ -se (noun/verb) & \textit{licence (n), practise (v)} & \textit{license (n/v), practice} \\
-ce vs.\ -se nouns & \textit{defence, offence} & \textit{defense, offense} \\
Programme vs.\ program & \textit{TV programme, postgraduate programme} & \textit{TV program, graduate program} \\
Orthographic pairs  & \textit{grey, cheque, manoeuvre, tyre, storey} & \textit{gray, check, maneuver, tire, story (floor)} \\
Directional suffix -ward(s) & \textit{towards, forwards, upwards} & \textit{toward, forward, upward} \\
Sceptic/k alternation & \textit{sceptic, sceptical} & \textit{skeptic, skeptical} \\
Judgement spelling & \textit{judgement} & \textit{judgment} \\
Maths/Math & \textit{maths} & \textit{math} \\
Season name & \textit{autumn} & \textit{fall} \\
\midrule
Present perfect vs.\ past & \textit{I've just eaten.} & \textit{I just ate.} \\
Mandative subjunctive & \textit{They suggested he \emph{should} apply.} & \textit{They suggested he apply.} \\
shall vs.\ will & \textit{I shall go tomorrow.} & \textit{I will go tomorrow.} \\
Irregular verb morphology & \textit{learnt, dreamt, spoilt} & \textit{learned, dreamed, spoiled} \\
Collective noun agreement & \textit{The team are winning.} & \textit{The team is winning.} \\
Possession verb & \textit{I’ve got a car.} & \textit{I have a car.} \\
Got vs.\ gotten & \textit{He’s got very tired.} & \textit{He’s gotten very tired.} \\
Prepositional usage & \textit{at the weekend}, \textit{in a team} & \textit{on the weekend}, \textit{on a team} \\
Tag questions & \textit{You’re ready, aren’t you?} & \textit{You’re ready, right?}\\
Subjunctive usage & \textit{They suggested he should go.} & \textit{They suggested he go.}\\
Auxiliary ellipsis & \textit{He must have done.} & \textit{He must have.} \\
Numerals (“and”) & \textit{one hundred and twenty} & \textit{one hundred twenty} \\
Restrictive relative marker & \textit{the report \emph{which} was submitted} & \textit{the report \emph{that} was submitted} \\
Possession questions & \textit{Have you got a pen?} & \textit{Do you have a pen?} \\
Necessity negative & \textit{You needn’t attend.} & \textit{You don’t need to attend.} \\
Difference construction & \textit{different from / different to} & \textit{different from / different than} \\
\midrule
Quotation marks & Prefers single quotes `\lq...\rq' & Prefers double quotes ``...'' \\
Commas/periods in quotes & Outside the closing quotes & Inside the closing quotes \\
Abbreviations with periods & \textit{Mr, Dr} & \textit{Mr., Dr.} \\
Oxford/serial comma & Rare & Common \\
\midrule
Date format (written) & \textit{31 December 2024} & \textit{December 31, 2024} \\
Date punctuation (written) & \textit{19 September 1973}  & \textit{September 19, 1973}  \\
Date format (numeric) & \textit{31/12/2024 (DD/MM/YYYY)} & \textit{12/31/2024 (MM/DD/YYYY)} \\
Legal/institutional terms & \textit{Ministry of Defence} & \textit{Department of Defense} \\
Institutional article usage & \textit{in hospital; at university} & \textit{in \emph{the} hospital; at \emph{the} university} \\
Floor numbering & \textit{ground floor, first floor (one up)} & \textit{first floor, second floor (one up)} \\
Time notation & \textit{11.15 pm; 23.15 common} & \textit{11:15 PM; 24-hour less common} \\
\bottomrule
\end{tabular}
\end{table}

%% file: tables/AmE-BrE-differences-2.tex
\begin{table}[t]
\centering
\footnotesize
\renewcommand{\arraystretch}{1.15}
\caption{\small Representative AmE--Ref(BrE) preferences across everyday domains, including transport, household, and food. Entries reflect common majority-preference usage and are illustrative rather than exhaustive.
}
\label{tab:br-am-phrases}
\begin{tabular}{p{3.8cm} p{4.4cm} p{4.4cm}}
\toprule
Category & British English (BrE) & American English (AmE) \\
\midrule
Preposition before days & \textit{She resigned on Thursday.} & \textit{She resigned Thursday.} \\
Street naming & \textit{in the High Street} & \textit{on Main Street} \\
Transitivity (protest) & \textit{protest against discrimination} & \textit{protest discrimination} \\
Ditransitives (write) & \textit{write to me} & \textit{write me} \\
Meeting collocation & \textit{meet the team} & \textit{meet with the team} \\
\hline
\textit{Transport \& wayfinding} & & \\
Pedestrian crossing & \textit{zebra crossing} & \textit{crosswalk} \\
Junction type & \textit{roundabout} & \textit{traffic circle / rotary} \\
Road maintenance & \textit{roadworks} & \textit{road work} \\
Parking payment & \textit{pay and display} & \textit{metered parking} \\
Perimeter road & \textit{ring road} & \textit{beltway} \\
Vehicle hire & \textit{hire car / car hire} & \textit{rental car / car rental} \\
Estate car vs.\ wagon & \textit{estate car} & \textit{station wagon} \\
\hline
\textit{Household \& services} & & \\
Postal addressing & \textit{postcode} & \textit{ZIP code} \\
Carry-on baggage & \textit{hand luggage} & \textit{carry-on} \\
Washing liquid & \textit{washing-up liquid} & \textit{dish soap} \\
Waste container & \textit{dustbin} & \textit{trash can / garbage can} \\
Clothes washer & \textit{washing machine} & \textit{washer} \\
Cash dispenser & \textit{cashpoint} & \textit{ATM} \\
Public convenience & \textit{public toilet} & \textit{restroom} \\
Mobile device & \textit{mobile phone} & \textit{cell phone} \\
\hline
\textit{Food \& drink} & & \\
Confection & \textit{candyfloss} & \textit{cotton candy} \\
Frozen treat & \textit{ice lolly} & \textit{popsicle} \\
Leafy green & \textit{rocket} & \textit{arugula} \\
Soft drink & \textit{fizzy drink} & \textit{soda} \\
Allium term & \textit{spring onion} & \textit{green onion / scallion} \\
Cake term & \textit{fairy cake} & \textit{cupcake} \\
\hline
\textit{Places \& urban terms} & & \\
City core & \textit{city centre} & \textit{downtown} \\
Real estate profession & \textit{estate agent} & \textit{realtor / real estate agent} \\
Holiday lodging & \textit{holiday let} & \textit{vacation rental} \\
Queueing term & \textit{post office queue} & \textit{post office line} \\
Queuing expression       & \textit{join the queue}    & \textit{wait in line} \\
Public transport info & \textit{railway timetable} & \textit{train schedule} \\
\hline
\textit{Education \& work} & & \\
Practical training & \textit{work placement} & \textit{internship} \\
Assessment term & \textit{marking scheme} & \textit{grading rubric} \\
Student level & \textit{first-year student} & \textit{freshman} \\
Residence & \textit{halls of residence} & \textit{dorm / residence hall} \\ 
\hline
\textit{Single-word vocabulary} & & \\
flat / apartment & \textit{flat} & \textit{apartment} \\
lorry / truck & \textit{lorry} & \textit{truck} \\
pavement / sidewalk & \textit{pavement} & \textit{sidewalk} \\
wardrobe / closet & \textit{wardrobe} & \textit{closet} \\
lift / elevator & \textit{lift} & \textit{elevator} \\
petrol / gas & \textit{petrol} & \textit{gas} \\
railway / railroad & \textit{railway} & \textit{railroad} \\
holiday / vacation & \textit{holiday} & \textit{vacation} \\
\bottomrule
\end{tabular}
\end{table}